\documentclass{article} 
\usepackage{iclr2021_conference,times}

\usepackage{amsmath,amsfonts,bm}

\def\eqref#1{equation~\ref{#1}}

\def\1{\bm{1}}

\newcommand{\test}{\mathcal{D_{\mathrm{test}}}}

\DeclareMathAlphabet{\mathsfit}{\encodingdefault}{\sfdefault}{m}{sl}
\SetMathAlphabet{\mathsfit}{bold}{\encodingdefault}{\sfdefault}{bx}{n}

\usepackage{hyperref}
\usepackage{url}
\usepackage{minitoc}
\usepackage[toc,page,header]{appendix}
\iclrfinalcopy

\title{Recurrent Independent Mechanisms}

\author{Anirudh Goyal\textsuperscript{1}, Alex Lamb\textsuperscript{1},
Jordan Hoffmann\textsuperscript{1, 2, *},
Shagun Sodhani\textsuperscript{1, *},
Sergey Levine\textsuperscript{4}\\
\textbf{Yoshua Bengio\textsuperscript{1, **},
Bernhard Sch\"olkopf \textsuperscript{3, **}}
}

\usepackage[utf8]{inputenc} 
\usepackage[T1]{fontenc}    
\usepackage{hyperref}       
\usepackage{url}            
\usepackage{booktabs}       
\usepackage{amsfonts}       
\usepackage{nicefrac}       
\usepackage{microtype}      
\usepackage[utf8]{inputenc} 
\usepackage[T1]{fontenc}    
\usepackage{hyperref}       
\usepackage{url}            
\usepackage{amsfonts}       
\usepackage{nicefrac}       
\usepackage{microtype}      
\usepackage{subfig}

\usepackage{booktabs,multirow} 
\usepackage{amsmath}
\usepackage{makecell}
\usepackage[export]{adjustbox}
\usepackage{bbm}

\usepackage{pgffor}
\usepackage{amssymb}
\usepackage{booktabs}
\usepackage{comment}
\usepackage{tikz}
\usepackage{asypictureB}
\usepackage{bm}
\usepackage{caption}
\usepackage{wrapfig}
\usetikzlibrary{backgrounds}
\usetikzlibrary{calc}
\usetikzlibrary{fit}
\usetikzlibrary{positioning}
\usepackage{enumitem}
\usepackage[acronym,smallcaps,nowarn]{glossaries}
\usepackage{soul}
\usepackage{hyperref}
\usepackage{graphicx} 

\usepackage[font={footnotesize}]{caption}

\usepackage{todonotes}
\usepackage{algorithm}
\usepackage[noend]{algpseudocode}
\usepackage{float,pbox}

\usepackage{pgfplots}
\usepackage{setspace}
\definecolor{mygreen}{HTML}{167dde}
\definecolor{myred}{HTML}{f22835}
\colorlet{greenfill}{blue!50!white}
\colorlet{purplefill}{red!50!blue!30!white}

\colorlet{redfill}{red!50!white}
\colorlet{moreredfill}{myred!40!white}
\usetikzlibrary{positioning}
\usetikzlibrary{backgrounds}
\usetikzlibrary{calc}
\usetikzlibrary{fit}

\usepackage{pifont}
\newcommand{\cmark}{\ding{51}}%
\newcommand{\xmark}{\ding{55}}%

\begin{document}
\doparttoc 
\faketableofcontents 

\part{} 

\maketitle
\vspace{-5mm}
\begin{abstract}
We explore the hypothesis that learning modular structures which reflect the dynamics of the environment can lead to better generalization and robustness to changes that only affect a few of the underlying causes. We propose Recurrent Independent Mechanisms (RIMs), a new recurrent architecture in which multiple groups of recurrent cells operate with nearly independent transition dynamics, communicate only sparingly through the bottleneck of attention, and compete with each other so they are updated only at time steps where they are most relevant.  We show that this leads to specialization amongst the RIMs, which in turn allows for remarkably improved generalization on tasks where some factors of variation differ systematically between training and evaluation.

\end{abstract}


\vspace{-2mm}
\section{Independent Mechanisms}
\vspace{-2mm}
\let\thefootnote\relax\footnotetext{\textsuperscript{1} Mila, University of Montreal,\textsuperscript{2}
Harvard University, \textsuperscript{3}
MPI for Intelligent Systems, T\"ubingen, \textsuperscript{4} University of California, Berkeley,
\textsuperscript{**} Equal advising, \textsuperscript{*} Equal Contribution.  :\texttt{anirudhgoyal9119@gmail.com}}

Physical processes in the world often have a modular structure which human cognition appears to exploit, with complexity emerging through combinations of simpler subsystems.    Machine learning seeks to uncover and use regularities in the physical world. Although these regularities manifest themselves as statistical dependencies, they are ultimately due to dynamic processes governed by causal physical phenomena. These processes are mostly evolving independently and only interact sparsely. For instance, we can model the motion of two balls as separate independent mechanisms even though they are both gravitationally coupled to Earth as well as (weakly) to each other. Only occasionally will they strongly interact via collisions. 


The notion of independent or autonomous mechanisms has been influential in the field of causal inference. A complex generative model, temporal or not, can be thought of as the composition of \emph{independent} mechanisms or ``causal'' modules. In the causality community, this is often considered a prerequisite for being able to perform localized interventions upon variables determined by such models \citep{Pearl2009}. It has been argued that the individual modules tend to remain robust or invariant even as other modules change, e.g., in the case of distribution shift \citep{SchJanPetSgoetal12,PetJanSch17}. This independence is not between the random variables being
processed but between the description or parametrization of the mechanisms: learning about one should not tell us anything about another, and adapting one should not require also adapting another. One may hypothesize that if a brain is able to solve multiple problems beyond a single i.i.d.\ (independent and identically distributed) task, they may exploit the existence of this kind of structure by learning independent mechanisms that can flexibly be reused, composed and re-purposed. 



In the dynamic setting, we think of an overall system being assayed as composed of a number of fairly independent subsystems that evolve over time, responding to forces and interventions. An agent needs not devote equal attention to all subsystems at all times: only those aspects that significantly interact need to be considered jointly when deciding or planning \citep{bengio2017consciousness}. 
Such sparse interactions can reduce the difficulty of learning since few interactions need to be considered at a time, reducing unnecessary interference when a subsystem is adapted.
%
%
Models learned this way may better capture the compositional generative (or causal) structure of the world, and thus better generalize across tasks where a (small) subset of mechanisms change while most of them remain invariant \citep{simon1991architecture, PetJanSch17}.The central question motivating our work is how a gradient-based deep learning approach can discover a representation of high-level variables which favour forming independent but sparsely interacting recurrent mechanisms in order to benefit from the modularity and independent mechanisms assumption. 



\begin{figure}
    \vspace{-18mm}
    \centering
    \subfloat{{\includegraphics[width=0.5\linewidth]{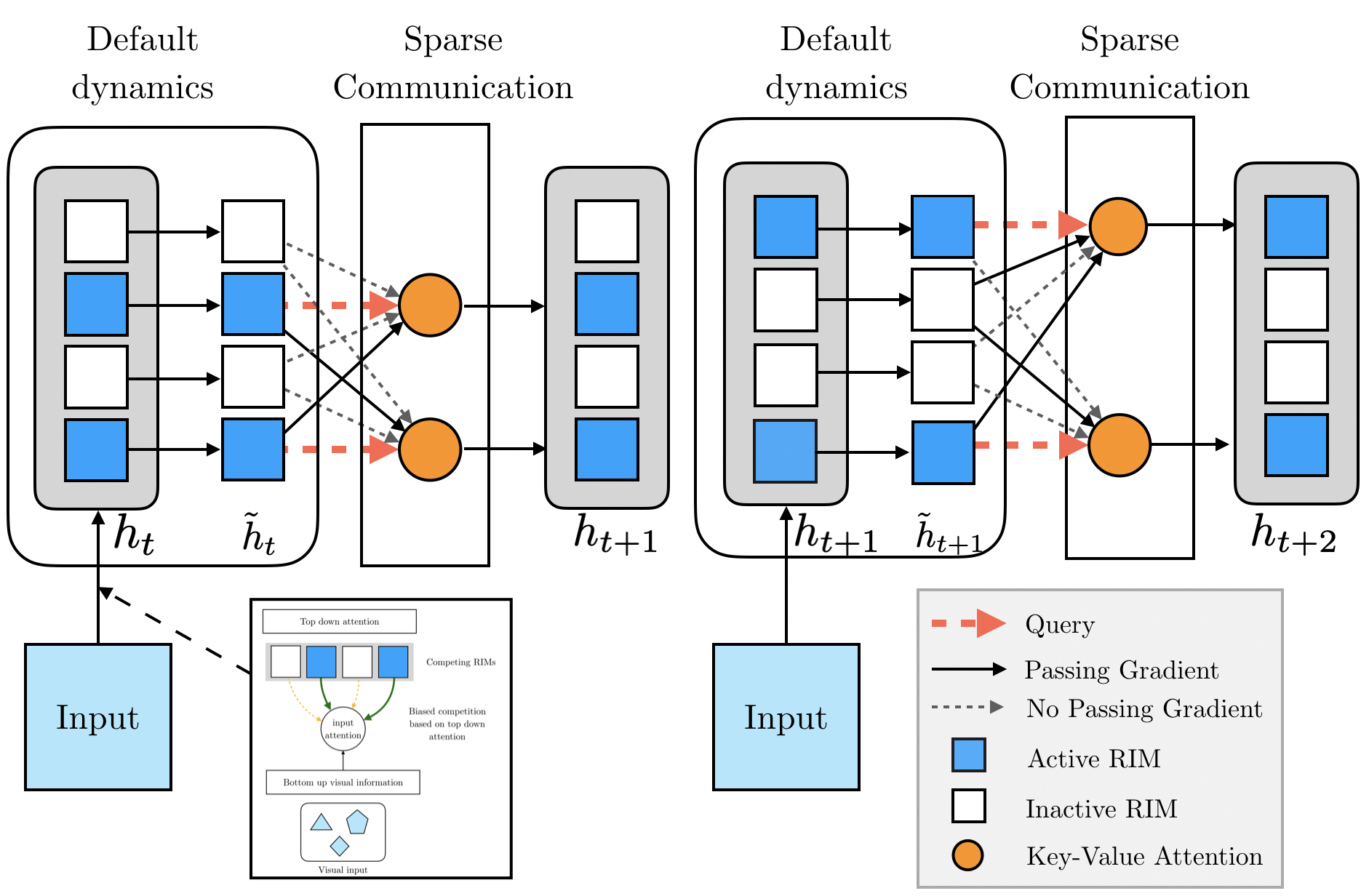}}}
    \qquad
    \subfloat{{\includegraphics[width=0.3\linewidth]{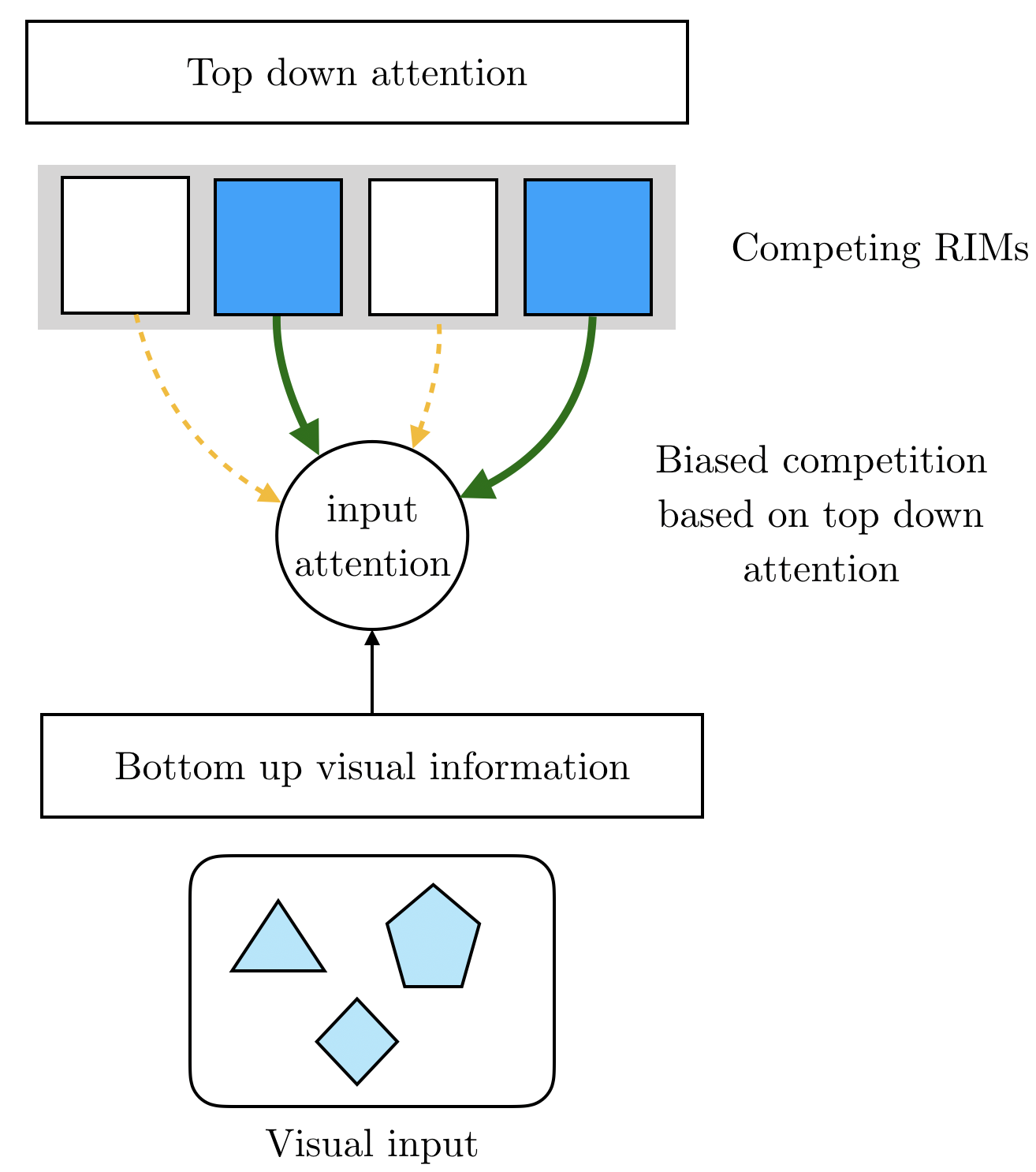}}}%
    \caption{\textbf{Illustration of Recurrent Independent Mechanisms (RIMs)}. A single step under the proposed model occurs in  four stages (left figure shows two steps). In the first stage, individual RIMs produce a query which is used to read from the current input.  In the second stage, an attention based competition mechanism is used to select which RIMs to activate (right figure) based on encoded visual input (blue RIMs are active, based on an attention score, white RIMs remain inactive).  In the third stage, individual activated RIMs follow their own default transition dynamics while non-activated RIMs remain unchanged.  In the fourth stage, the RIMs sparsely communicate information between themselves, also using key-value attention. }
    \label{fig:rims_diagram}%
    \vspace{-3mm}
\end{figure}

\paragraph{Why do Models Succeed or Fail in Capturing Independent Mechanisms?}
While universal approximation theorems apply in the limit of large i.i.d.\ data sets, we are interested in the question of whether models can learn independent mechanisms from finite data in possibly changing environments, and how to implement suitable inductive biases.   As the simplest case, we can consider training an RNN consisting of $k$ completely independent mechanisms which operate on distinct time steps.  How difficult would it be for an RNN (whether vanilla or LSTM or GRU) to correctly model that the true distribution has completely independent processes?  For the hidden states to truly compartmentalize these different processes, a fraction $\frac{k-1}{k}$ of the connections would need to be set to exactly zero weight.  This fraction approaches $100\%$ as $k$ approaches infinity.  When sample complexity or out-of-distribution generalization matter, we argue that having an inductive bias which favors this form of modularity and dynamic recombination could be greatly advantageous, compared to static fully connected monolithic architectures.


\paragraph{Assumptions on the joint distribution of high level variables.} The central question motivating our work is how a gradient-based deep learning approach can learn a representation of high level variables which favour learning independent but sparsely interacting recurrent mechanisms in order to benefit from such  modularity assumption.  The assumption about the joint distribution between the high-level variables  is different from the assumption commonly found in many papers on disentangling factors of variation~\citep{higgins2016beta, burgess2018understanding, chen2018isolating}, where the high level variables are assumed to be marginally independent of each other. We believe that these variables, (often named with words in language), have highly structured dependencies supporting independent mechanisms assumption.



\section{RIMs with Sparse Interactions}
\label{sec:rimsmodel}


Our approach to modelling a dynamical system of interest divides the overall model into $k$ small subsystems (or modules), each of which is recurrent in order to be able to capture the dynamics in the observed sequences. We refer to these subsystems as \textit{Recurrent Independent Mechanisms (RIMs)}, where each RIM has distinct functions that are learned automatically from data\footnote{Note that we are overloading the term \textit{mechanism},
using it both for the mechanisms that make up the world's dynamics as well as for the computational modules that we learn to model those mechanisms. The distinction should be clear from context.}. 
We refer to RIM $k$ at time step $t$ as having vector-valued state $h_{t,k}$, where $t=1,\dots, T$. Each RIM has parameters $\theta_k$, which are shared across all time steps.

At a high level (see Fig.~\ref{fig:rims_diagram}), we want each RIM to have its own independent dynamics operating by default, and occasionally to interact with other relevant RIMs and selected elements of the encoded input.  
The total number of parameters can be kept small since RIMs can specialize on simple sub-problems, and operate
on few key/value variables at a time
selected using an attention mechanism,
as suggested by the inductive bias from~\citet{bengio2017consciousness}. This specialization and modularization not only has computational and statistical advantages \citep{baum1989size}, but also prevents individual RIMs from dominating the computation and thus facilitates factorizing the computation into easy to recombine but simpler elements. We expect this to lead to more robust systems than training one big homogeneous system \citep{schmidhuber2018one}.
Moreover, modularity and the independent mechanisms hypothesis~\citep{PetJanSch17,bengio2019meta} also has the desirable implication that a RIM should maintain its own independent functionality even as other RIMs are changed. A more detailed account of the desiderata for the model is given in Appendix~\ref{sec:desiderata}.

\subsection{Key-Value Attention to Process Sets of Named Interchangeable Variables}
\label{sec:kvattention}



Each RIM should be activated and updated when the input is relevant to it. We thus utilize competition to allocate representational and computational resources, using an attention mechanism which selects and then
activates only a subset of the RIMs for each time step. As argued by \cite{ParKilRojSch18}, this tends to produce independence among learned mechanisms, provided the training data has been generated by a set of independent physical mechanisms. 
In contrast to \cite{ParKilRojSch18}, we use an {\em attention mechanism} for this purpose.  The introduction of content-based soft-attention mechanisms~\citep{bahdanau2014neural} has opened the door to neural networks  which operate on {\em sets of typed interchangeable objects}.
This idea has been remarkably successful and widely applied to most recent Transformer-style multi-head dot product self attention  models \citep{vaswani2017attention, santoro2018relational}, achieving new state-of-the-art results in many tasks. Soft-attention uses the product of a \emph{query} (or \emph{read key}) represented as a matrix $Q$ of dimensionality $N_r \times d$, with $d$ the dimension of each key, with a set of $N_o$ objects each associated with a \emph{key} (or \emph{write-key}) as a row in matrix $K^T$ ($N_o \times d$), and after normalization with a softmax yields outputs in the convex hull of the \emph{values} (or \emph{write-values}) $V_i$ (row $i$ of matrix $V$). 
The result is 
$$\mathrm{Attention}(Q,K,V)=\mathrm{softmax} \left (\frac{QK^T}{\sqrt{d}} \right )V,$$
where the softmax is applied to each row of its argument matrix, yielding a set of convex weights. As a result, one obtains a convex combination of the values in the rows of $V$. If the attention is focused on one element for a particular row (i.e., the softmax is saturated), this simply selects one of the objects and copies its value to row $j$ of the result.  Note that the $d$ dimensions in the key can be split into \emph{heads} which then have their own attention matrix and write values computed separately.  

When the inputs and outputs of each RIM are a set of objects or entities (each associated with a key and value vector), the RIM processing becomes a generic object-processing machine which can operate on subsymbolic ``variables'' in a sense analogous to variables in a programming language: as interchangeable arguments of functions, albeit with a distributed representation both for they name or type and for their value. Because each object has a key embedding (which one can understand both as a name and as a type), the same RIM processing can be applied to any variable which fits an expected "distributed type" (specified by a query vector). Each attention head then corresponds to a typed argument of the function computed by the RIM. When the key of an object matches the query of head $k$, it can be used as the $k$-th input vector argument
for the RIM. Whereas in regular neural networks (without attention) neurons operate on fixed variables (the neurons which are feeding them from the previous layer), the key-value attention mechanisms make it possible to select on the fly which variable instance (i.e. which entity or object) is going to be used as input for each of the arguments of the RIM dynamics, with a different set of query embeddings for each RIM head. These inputs can come from the external input or from the output of other RIMs. So, if the individual RIMs can represent these \textit{functions with typed arguments}, then they can \emph{bind} to whatever input is currently available and best suited according to its attention score: the ``input attention'' mechanism would look at the candidate input object's key and evaluate if its ``type'' matches with what this RIM expects (specified with the corresponding query). 

\subsection{Selective Activation of RIMs as a form of Top-Down Modulation}
\label{sec:selectiveactivationrims}



The proposed model learns to dynamically select those RIMs for which the current input is relevant.  
RIMs are triggered as a result of interaction between the current state of the RIM and input information coming from the environment.  At each step, we select the top-$k_A$ (out of $k_{T}$) RIMs in terms of their attention score for the real input. Intuitively, the RIMs must compete on each step to read from the input, and only the RIMs that win this competition will be able to read from the input and have their state updated.   In our use of key-value attention, the queries come from the RIMs, while the keys and values come from the current input.  This differs from the mechanics of \citet{vaswani2017attention,santoro2018relational}, with the modification that the parameters of the attention mechanism itself are separate for each RIM rather than produced on the input side as in Transformers. The input attention for a particular RIM is described as follows.

%



The input $x_t$ at time $t$ is seen as a set of elements, structured as rows of a matrix. We first concatenate a row full of zeros, to obtain
\begin{align}
    X &= \emptyset \oplus x_t.
\end{align}

As before, linear transformations are used to construct  keys ($K = XW^e$, one per input element and for the null element),  values ($V = XW^v$, again one per element), and  queries ($Q = h_{t} W_k^q$, one per RIM attention head). $W^v$ is a simple matrix mapping from an input element to the corresponding value vector for the weighted attention and $W^e$ is similarly a weight matrix which maps the input to the keys.  $W_k^q$ is a per-RIM weight matrix which maps from the RIM's hidden state to its queries.  $\oplus$ refers to the row-level concatenation operator. 
The attention thus is
\begin{align}
    A^{(in)}_{k} &= \text{softmax} \left( \frac{h_{t} W_k^q(X W^e)^T}{\sqrt{d_e}} \right) X W^v , \text{ where } \theta^{(in)}_k =(W_k^q, W^e, W^v). \label{eq:non_spat_atten}
\end{align}

Based on the softmax values in (\ref{eq:non_spat_atten}), we select the top $k_A$ RIMs (out of the total $K$ RIMs) to be activated for each step, which have the least attention on the null input (and thus put the highest attention on the input), and we call this set $\mathcal{S}_t$.  Since the queries depend on the state of the RIMs, this enables individual RIMs to attend only to the part of the input that is relevant for that particular RIM, thus enabling selective attention based on a \textit{top-down attention} process (see. Fig \ref{fig:rims_diagram}). In practice, we use multiheaded attention, and multi-headed attention doesn't change the essential computation, but when we do use it for input-attention we compute RIM activation by averaging the attention scores over the heads.

\textit{For spatially structure input:} All datasets we considered are temporal, yet there is a distinction between whether the input on each time step is highly structured (such as a video) or not (such as language modeling, where each step has a word or character).  In the former case, we can get further improvements by making the activation of RIMs not just sparse across time but also sparse across the (spatial) structure.  The input $x_t$ at time $t$ can be seen as an output of the encoder parameterized by a neural network (for ex. CNN in case of visual observations) i.e., $X = CNN(x_t)$. As before, linear transformations are used to construct position-specific input  keys ($K = XW^e$), position-specific values ($V = XW^v$), and RIM specific queries ($Q = h_t W_k^q$, one per RIM attention head). 
The attention thus is

\vspace{-1mm}
\begin{align}
    A^{(in)}_{k} &= \text{softmax} \left( \frac{h_t W_k^q(X W^e)^T}{\sqrt{d_e}} \right) X W^v , \text{ where } & \theta^{(in)}_k =(W_k^q, W^e, W^v). \label{eq:spatial_atten}
\end{align}

In order for different RIMs to specialize on different spatial regions, we can use position-specific competition among the different RIMs. The contents of the attended positions are combined yielding a RIM-specific input. As before based on the softmax values in (\ref{eq:spatial_atten}), we can select the top $k_A$ RIMs (out of the total $k_{T}$ RIMs) to be activated for each spatial position, which have the highest attention on that spatial position.




%
%

\subsection{Independent RIM Dynamics}
Now, consider the default transition dynamics which we apply for each RIM independently and during which no information passes between RIMs.  We use $\tilde{h}$ for the hidden state after the independent dynamics are applied.  The hidden states of RIMs which are not activated (we refer to the activated set as $\mathcal{S}_t$) remain unchanged, acting like untouched memory elements, i.e., 
$    h_{t+1,k} = h_{t,k} \quad \forall k \notin \mathcal{S}_t.$
Note that the gradient still flows through a RIM on a step where it is not activated.  For the RIMs that are activated, we run a per-RIM independent transition dynamics.  The form of this is somewhat flexible, but we opted to use either a GRU \citep{chung2015recurrent} or an LSTM \citep{hochreiter1997long}.  We generically refer to these independent transition dynamics as $D_k$, and we emphasize that each RIM has its own separate parameters.  Aside from being RIM-specific, the internal operation of the LSTM and GRU remain unchanged, and active RIMs are updated by 
\begin{equation*}
    \tilde{h}_{t,k} = D_k(h_{t,k}) = LSTM(h_{t,k}, A^{(in)}_{k}; \theta^{(D)}_k) \quad \forall k \in \mathcal{S}_t 
\end{equation*}
as a function of the attention mechanism $A^{(in)}_{k}$ applied on the current input, described in the previous sub-section. 

\subsection{Communication between RIMs}
\label{sec:communicationrims}

Although the RIMs operate independently by default, the attention mechanism allows sharing of information among the RIMs.  Specifically, we allow the activated RIMs to read from all other RIMs (activated or not).  The intuition behind this is that non-activated RIMs are not related to the current input, so their value needs not change.  However they may still store contextual information relevant for activated RIMs later on.  For this communication between RIMs, we use a residual connection as in \cite{santoro2018relational} to prevent vanishing or exploding gradients over long sequences.  
Using parameters $\theta^{(c)}_k =(\tilde{W}_k^q, \tilde{W}_k^e, \tilde{W}_k^v)$, we employ
\begin{align}
    Q_{t,k} = \tilde{W}^{q}_k \tilde{h}_{t,k}, \forall k \in \mathcal{S}_t \quad\quad 
    K_{t,k} = \tilde{W}^{e}_k \tilde{h}_{t,k}, \forall k \quad\quad 
    V_{t,k} = \tilde{W}^{v}_k \tilde{h}_{t,k}, \forall k  
    \nonumber
\end{align}
\begin{align*}
    h_{t+1,k} &= \text{softmax} \left( \frac{Q_{t,k} (K_{t,:})^T}{\sqrt{d_e}} \right) V_{t,:} + \tilde{h}_{t,k}  \forall k \in \mathcal{S}_t .
\end{align*}

We note that we can also consider sparsity in the communication attention such that a particular RIM only attends to sparse sub-set of other RIMs, and this sparsity is orthogonal to the kind used in input attention. In order to make the communication attention sparse, we can still use the same top-$k$ attention. 

\textit{Number of Parameters.}  RIMs can be used as a drop-in replacement for an LSTM/GRU layer. There is a subtlety that must be considered for successful integration. If the total size of the hidden state is kept the same, integrating RIMs drastically reduces the total number of recurrent parameters in the model (because of having a block-sparse structure). RIMs also adds new parameters to the model through the addition of the attention mechanisms although these are rather in small number.


\textit{Multiple Heads:} Analogously to \citet{vaswani2017attention, santoro2018relational}, we use multiple heads both for communication between RIMs as well as input attention (as in Sec \ref{sec:selectiveactivationrims}) by producing different sets of queries, keys, and values to compute a linear transformation for each head (different heads have different parameters), and then applying the attention operator for each head separately in order to select conditioning inputs for the RIMs.

\vspace{-3mm}

\section{Related Work\label{sec:related_work}}


\textbf{Neural Turing Machine (NTM) and Relational Memory Core (RMC):}  the NTM \citep{graves2014neural} updates independent memory cells using an attention mechanism to perform targeted read and write operations.  RIMs share a key idea with NTMs: that input information should only impact a sparse subset of the memory by default, while keeping most of the memory unaltered. RMC \citep{santoro2018relational}  uses a multi-head attention mechanism to share information between multiple memory elements. We encourage the RIMs to remain separate as much as possible, whereas \cite{santoro2018relational} allow information between elements to flow on each step in an unsconstrained way. Instead, each RIM has its own default dynamics, while in RMC, all the processes interact with each other.

\textbf{Separate Recurrent Models:} EntNet \citep{henaff2016tracking} and IndRNN \citep{li2018independently} can be viewed as a set of separate recurrent models. In IndRNN, each recurrent unit has completely independent dynamics, whereas EntNet uses an independent gate for writing to each memory slot. RIMs use  different recurrent models (with separate parameters), but we allow the RIMs to communicate with each other sparingly using an attention mechanism.

\textbf{Modularity and Neural Networks}: A network can be composed of several modules, each meant to perform a distinct function, and hence can be seen as a combination of experts \citep{jacobs1991adaptive, bottou1991framework,ronco1996modular, reed2015neural, andreas2016neural,ParKilRojSch18,rosenbaum2017routing, fernando2017pathnet, shazeer2017outrageously, kirsch2018modular, rosenbaum2019routing} routing information through a gated activation of modules. These works generally assume that only a single expert is active at a particular time step. In the proposed method, multiple RIMs can be active, interact and share information.



\textbf{Computation on demand:}  There are various  architectures \citep{el1996hierarchical, koutnik2014clockwork, chung2016hierarchical, neil2016phased, jernite2016variable, krueger2016zoneout} where  parts of the RNN's hidden state are kept dormant at times. The major differences to our architecture are that (a) we modularize the dynamics of recurrent cells (using RIMs), and (b) we also control the inputs of each module (using transformer style attention), while many previous gating methods did not control the inputs of each module, but only whether they should be executed or not. 



\vspace{-1mm}

\section{Experiments}

\vspace{-1mm}

The main goal of our experiments is to show that the use of RIMs improves generalization across changing environments and/or in modular tasks, and to explore how it does so.  
Our goal is not to outperform highly optimized baselines; rather, we want to show the versatility of our approach by applying it to a range of diverse tasks, focusing on tasks that involve a changing environment.
We organize our results by the capabilities they illustrate: we address generalization based on temporal patterns, based on objects, and finally consider settings where both of these occur together. 

\subsection{RIMs improve generalization by specializing over temporal patterns}
\vspace{-1mm}

We first show that when RIMs are presented with sequences containing distinct and generally independent temporal patterns, they are able to specialize so that different RIMs are activated on different patterns.  RIMs generalize well when we modify a subset of the patterns (especially those unrelated to the class label) while most recurrent models fail to generalize well to these variations.


\label{sec:copying}
\begin{figure}[h]
    \centering
    \includegraphics[width=0.79\linewidth,trim=8.0cm 0.0cm 4.0cm 0.0cm,clip]{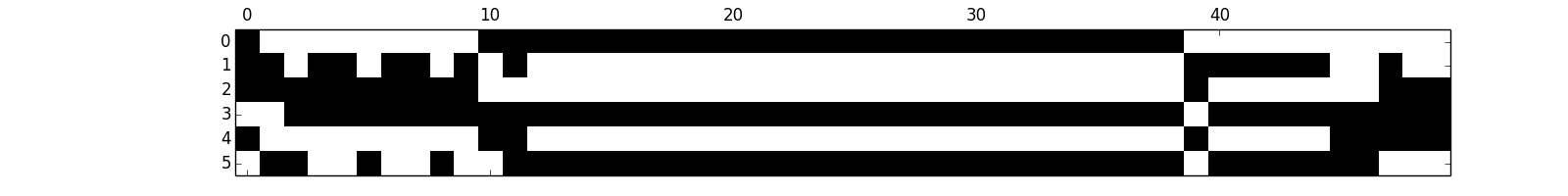}
    \caption{\textbf{Visualizing Activation Patterns}.  
    For the copying task,  one can see that the RIM activation pattern is distinct during the dormant part of the sequence in the middle (activated RIMs black, non-activated white).  X-axis=time, Y-axis=RIMs activation bit.}
    \label{fig:copying_pattern}
\end{figure}

\paragraph{Copying Task:}
First we turn our attention to the task of receiving a short sequence of characters, then receiving blank inputs for a large number of steps, and then being asked to reproduce the original sequence.  We can think of this as consisting of two temporal patterns which are independent: one where the sequence is received and another ``dormant'' pattern where no input is provided. As an example of out-of-distribution generalization, we find that using RIMs, we can extend the length of this dormant phase from 50 during training to 200 during testing and retain perfect performance (Table~\ref{tb:copy_mnist}), whereas baseline methods including LSTM, NTM, and RMC substantially degrade.  In addition, we find that this result is robust to the number of RIMs used as well as to the number of RIMs activated per-step. Our ablation results (Appendix~\ref{table:copying_ablation}) show that all major components of the RIMs model are necessary to achieve this  generalization. This is evidence that RIMs can specialize over distinct patterns in the data and improve generalization to settings where these patterns change.  

\begin{table*}[t]
\footnotesize
\resizebox{0.36\columnwidth}{!}{
\begin{minipage}[t]{0.4\textwidth}
  \begin{tabular}{p{1.3ex}rrr|r|r}
  \toprule
  \multicolumn{4}{l|}{\textbf{Copying}} &
  \multicolumn{1}{c|}{\textbf{Train(50)}} &
  \multicolumn{1}{c}{\textbf{Test(200)}} \\
  & \parbox{2em}{$k_\textrm{T}$} & $k_\textrm{A}$ & $h_\textrm{size}$ & 
  CE &
  CE \\ 
  \midrule
  \multirow{4}{*}{\rotatebox{0}{\parbox{2.5em}{\textbf{RIMs}}}}
  & 6 & 4 & 600 & \textbf{0.00} & \textbf{0.00} \\
  & 6 & 3 & 600 & \textbf{0.00} & \textbf{0.00} \\
  & 6 & 2 & 600 & \textbf{0.00} & \textbf{0.00} \\
  & 5 & 2 & 500 & \textbf{0.00} & \textbf{0.00} \\
  \midrule
  \multirow{2}{*}{\rotatebox{0}{\parbox{2.5em}{\textbf{LSTM}}}}
  & - & - & 300 &  0.00 & 4.32 \\
  & - & - & 600 &  0.00 & 3.56 \\
  \midrule
  \multirow{1}{*}{\rotatebox{0}{\parbox{2.5em}{\textbf{NTM}}}}
  & - & - & - &  0.00 & 2.54 \\
  \midrule
  \multirow{1}{*}{\rotatebox{0}{\parbox{2.5em}{\textbf{RMC}}}}
  & - & - & - & 0.00 & 0.13 \\
  \midrule
  \multirow{1}{*}{\rotatebox{0}{\parbox{2.5em}{\textbf{Transformers}}}}
  & - & - & - & 0.00 & 0.54 \\
  \bottomrule
  \end{tabular}
\end{minipage}}
\hspace{3.3em}
\resizebox{0.36\columnwidth}{!}{
\begin{minipage}[t]{0.4\textwidth}
  \begin{tabular}{p{1.3ex}rrr|r|r|r}
  \toprule
  \multicolumn{4}{l|}{\textbf{Sequential MNIST}} &
  \multicolumn{1}{c|}{\textbf{16 x 16}} &
  \multicolumn{1}{c}{\textbf{19 x 19}} &
  \multicolumn{1}{c}{\textbf{24 x 24}} \\
  & \parbox{2em}{$k_\textrm{T}$} & $k_\textrm{A}$ & $h_\textrm{size}$ & 
  Accuracy &
  Accuracy &
  Accuracy \\ 
  \midrule
  \multirow{4}{*}{\rotatebox{0}{\parbox{2.5em}{\textbf{RIMs}}}}
  & 6 & 6 & 600 & 85.5 & 56.2 & 30.9\\
  & 6 & 5 & 600 & 88.3 & 43.1 & 22.1 \\
  & 6 & 4 & 600 & \textbf{90.0} & \textbf{73.4} & \textbf{38.1}  \\
  \midrule
  \multirow{2}{*}{\rotatebox{0}{\parbox{2.5em}{\textbf{LSTM}}}}
  & - & - & 300 &  86.8 & 42.3 & 25.2 \\
  & - & - & 600 &  84.5 & 52.2 & 21.9\\
  \midrule
  \multirow{1}{*}{\rotatebox{0}{\parbox{2.5em}{\textbf{EntNet}}}}
  & - & - & - &  89.2 & 52.4 & 23.5 \\
  \midrule
  \multirow{1}{*}{\rotatebox{0}{\parbox{2.5em}{\textbf{RMC}}}}
  & - & - & - &  89.58 & 54.23 & 27.75 \\
  \midrule
  \multirow{1}{*}{\rotatebox{0}{\parbox{2.5em}{\textbf{DNC}}}}
  & - & - & - &  87.2 & 44.1 & 19.8 \\
  \multirow{1}{*}{\rotatebox{0}{\parbox{2.5em}{\textbf{Transformers}}}}
  & - & - & - &  \textbf{91.2} & 51.6 & 22.9 \\
  \bottomrule
  \end{tabular}
\end{minipage}}
\caption{Performance on the copying task (left) and Sequential MNIST resolution generalization (right).  While all of the methods are able to learn to copy for the length seen during training, the RIMs model generalizes to sequences longer than those seen during training whereas the LSTM, RMC, and NTM degrade much more.  On sequential MNIST, both the proposed and the Baseline models were trained on 14x14 resolution but evaluated at different resolutions (averaged over 3 trials). 
}
\label{tb:copy_mnist}
\vspace{-4mm}
\end{table*}




\paragraph{Sequential MNIST Resolution Task:}
RIMs are motivated by the hypothesis that generalization performance can benefit from modules which only activate on relevant parts of the sequence.  For further evidence that RIMs can achieve this out-of-distribution, we consider the task of classifying MNIST digits as sequences of pixels \citep{krueger2016zoneout} and assay generalization to images of resolutions different from those seen during training.  Our intuition is that the RIMs model should have distinct subsets of the RIMs activated for pixels with the digit and empty pixels.  RIMs should generalize better to higher resolutions by keeping RIMs dormant which store pixel information over empty regions of the image.

\textbf{Results:} Table \ref{tb:copy_mnist} shows the result of the proposed model on the Sequential MNIST Resolution Task.  If the train and test sequence lengths agree, both models achieve comparable test set performance.  However,  RIMs model is relatively robust to changing the sequence length (by changing the image resolution), whereas the LSTM performance degraded more severely.  This can be seen as a  more involved analogue of the copying task, as MNIST digits contain large empty regions.  It is essential that the model be able to store information and pass gradients through these regions.  The RIMs outperform strong baselines such as Transformers, EntNet, RMC, and (DNC) \citep{graves2016hybrid}.

\subsection{RIMs learn to specialize over objects and generalize between them}

\begin{figure*}[htb!]
    \vspace{-5mm}
    \centering
    \includegraphics[width=0.75\linewidth]{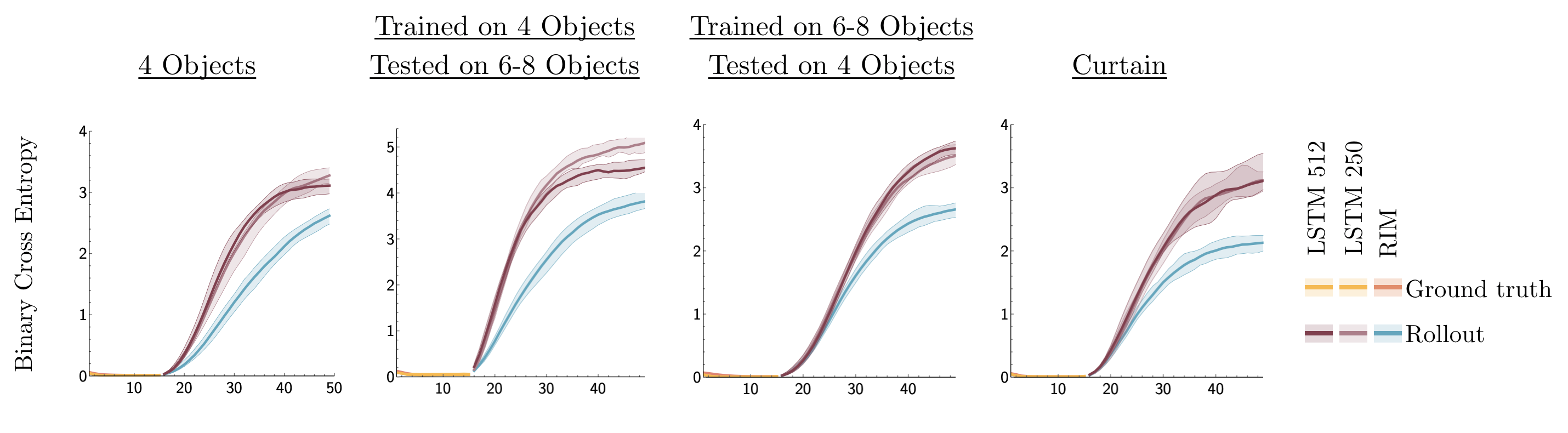}
    \includegraphics[width=0.24\textwidth,trim={0 0 17cm 0},clip]{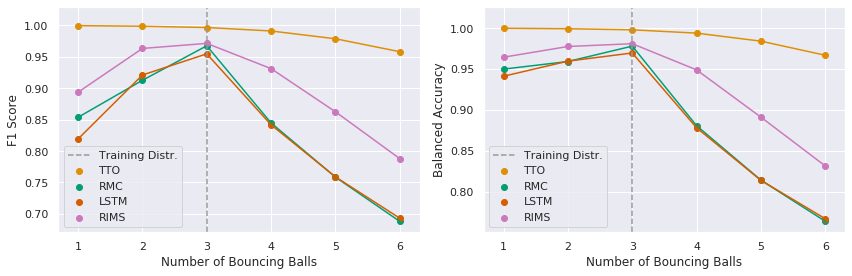}
    \vspace{-0.5cm}
    \caption{\textbf{Handling Novel Out-of-Distribution Variations}. We study the performance of RIMs compared to an LSTM baseline (4 left plots). The first 15 frames of ground truth (yellow,orange) are fed in and then the system is rolled out for the next 35 time steps (blue,purple). During the rollout phase, RIMs perform better than the LSTMs in accurately predicting the dynamics of the balls as reflected by the lower Cross Entropy (CE) [blue for RIMs, purple for LSTMs].  Notice the substantially better out-of-distribution generalization of RIMs when testing on a number of objects different from the one seen during training. (2nd to 4th plot). \textbf{ We also show (right plot) improved out-of-distribution generalization (F1 score) as compared to LSTM and RMC \citep{santoro2018relational} on another partial observation video prediction task.} X-axis = number of balls. For these experiments, the RIMs and baselines get an input image at each time step (see  Appendix \ref{sec:bballs_appendix}, figure.~\ref{fig:bb_ood} for magnified image as well as more details).  Here, TTO refers to the time travelling oracle upper bound baseline, that does not model the dynamics, and has access to true dynamics.}
    \label{fig:err}
\vspace{-4mm}

\end{figure*}

We have shown that RIMs can specialize over temporal patterns.  We now turn our attention to assaying whether RIMs can specialize to objects, and show improved generalization to cases where we add or remove objects at test time.  
\vspace{-3mm}
\paragraph{Bouncing Balls Environment:}
We consider a synthetic ``bouncing balls'' task in which multiple balls (of different masses and sizes) move using basic Newtonian physics \citep{van2018relational}.  What makes this task particularly suited to RIMs is that the balls move independently most of the time, except when they collide.  During training, we predict the next frame at each time step using teacher forcing \citep{williams1989learning}.  We can then use this model to generate multi-step rollouts.  As a preliminary experiment, we train on sequences of length 51 (the previous standard), using a binary cross entropy loss when predicting the next frame.  We consider LSTMs  as baselines. We then produce rollouts, finding that RIMs are better able to predict future motion (Figure~\ref{fig:err}).  





We take this further by evaluating RIMs on environments where the test setup is different from the training setup.  First we consider training with 4 balls and evaluating on an environment with 6-8 balls.  Second, we consider training with 6-8 balls and evaluating with just 4 balls. Robustness in these settings requires a degree of invariance w.r.t.\ the number of balls.


In addition, we consider a task where we train on 4 balls and then evaluate on sequences where part the visual space is occluded by a ``curtain.''  This allows us to assess the ability of balls to be tracked (or remembered) through the occluding region.  
Our experimental results on these generalization tasks (Figure~\ref{fig:err}) show that RIMs substantially improve over an LSTM baseline.  We found that increasing the capacity of the LSTM from 256 to 512 units did not substantially change the performance gap, suggesting that the improvement from RIMs is not primarily related to capacity.

\vspace{-2mm}

\paragraph{Environment with Novel Distractors:}
We next consider an object-picking reinforcement learning task from BabyAI \citep{chevalier2018babyai} in which an agent must retrieve a specific object in the presence of distractors.  We use a partially observed formulation of the task, where the agent only sees a small number of squares ahead of it. These tasks are difficult to solve \citep{chevalier2018babyai} with standard RL algorithms, due to (1) the partial observability of the environment and (2) the sparsity of the reward, given that the agent receives a reward only after reaching the goal.   During evaluation, we introduce new distractors to the environment which were not observed during training. 

\begin{wrapfigure}{r}{0.6\textwidth}
    \centering
    \vspace{-6mm}
    \includegraphics[width=0.45\linewidth,trim=0.5cm 0.0cm 1.0cm 1.5cm,clip]{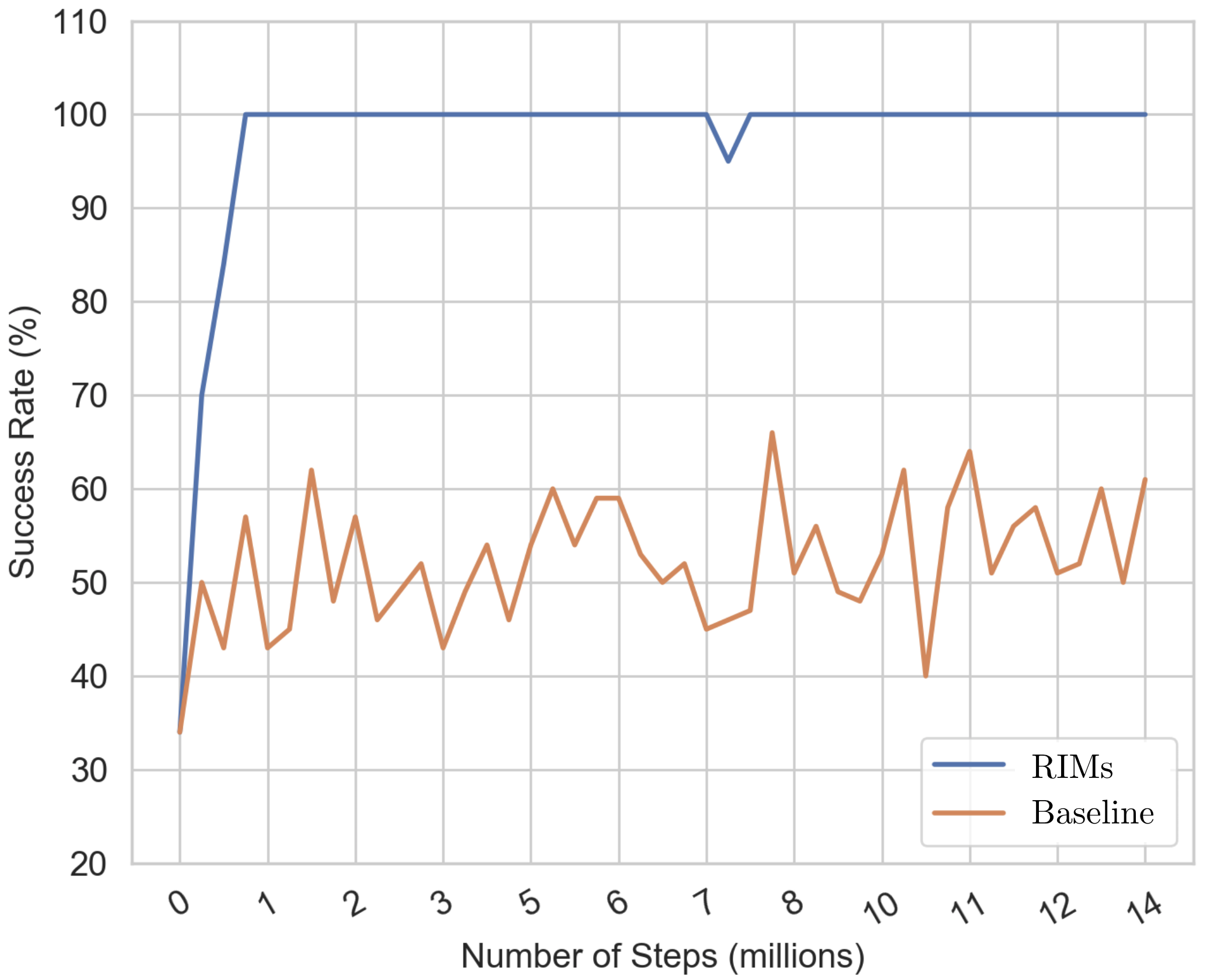}
    \includegraphics[width=0.45\linewidth,trim=0.5cm 0.0cm 1.0cm 1.5cm,clip]{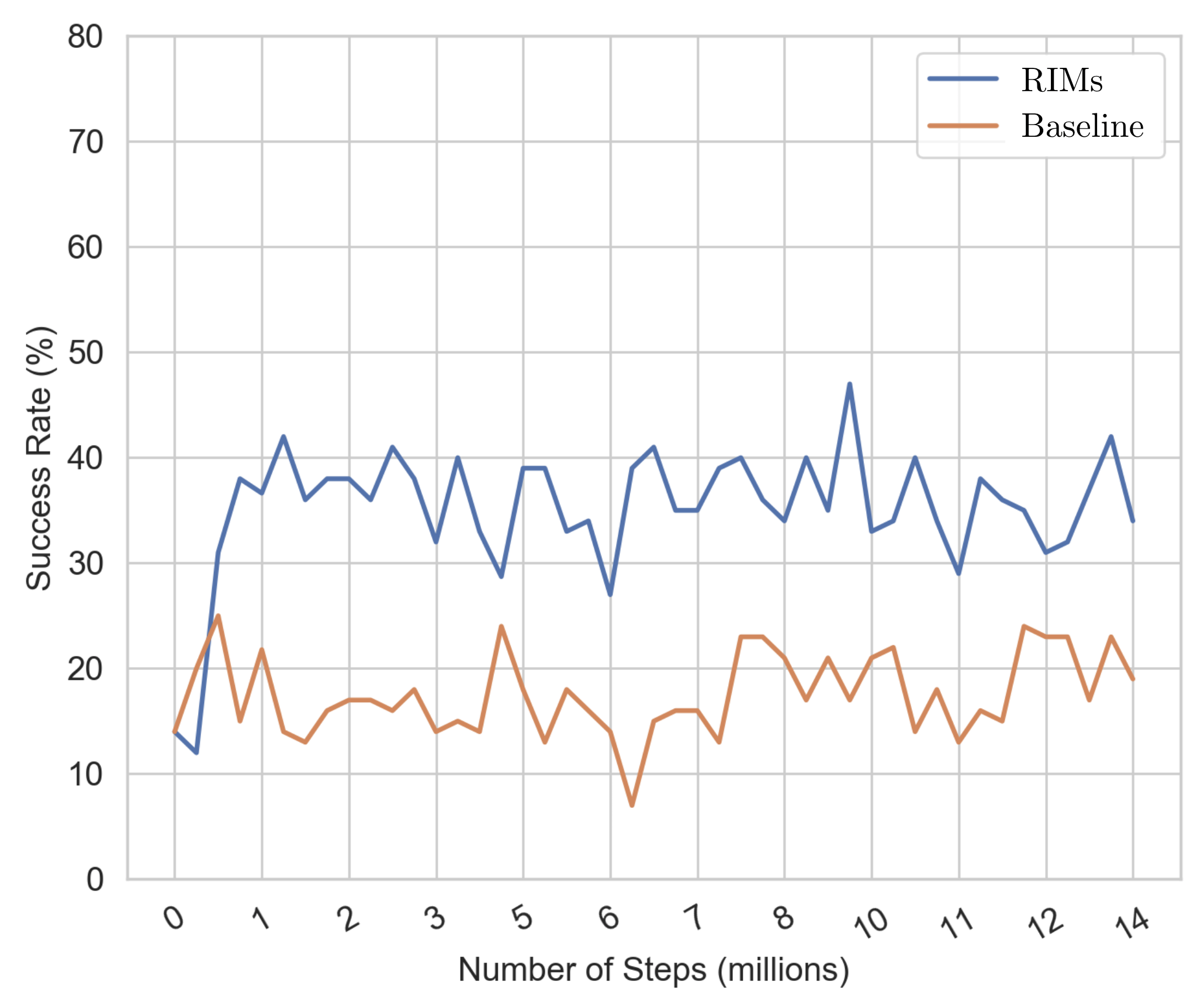}
    \caption{\textbf{Robustness to Novel Distractors:}. Left: performance of the proposed method (blue) compared to an LSTM baseline (red) in solving the object picking task in the presence of distractors.  Right: performance of proposed method and the baseline when novel distractors are added.}
    \label{fig:rl_exp}
    \vspace{-5mm}
\end{wrapfigure}


Figure~\ref{fig:rl_exp} shows that RIMs outperform LSTMs on this task (details in appendix).  When evaluating with known distractors, the RIM model achieves perfect performance while the LSTM struggles.   When evaluating in an environment with novel unseen distractors the RIM doesn't achieve perfect performance but strongly outperforms the LSTM.  An LSTM with a single memory flow may struggle to keep the distracting elements separate from elements which are necessary for the task, while the RIMs model uses attention to control which RIMs receive information at each step as well as what information they receive (as a function of their hidden state). This "top-down" attention results in a diminished representation of the distractor, not only enhancing the target visual information, but also suppressing irrelevant information.

\vspace{-3mm}
\subsection{RIMs improve generalization in complex environments}

\begin{figure*}[htb!]
    \centering
    \vspace{-3mm}
    \includegraphics[width=0.8\linewidth,trim=2.8cm 0.0cm 3.5cm 1.0cm,clip]{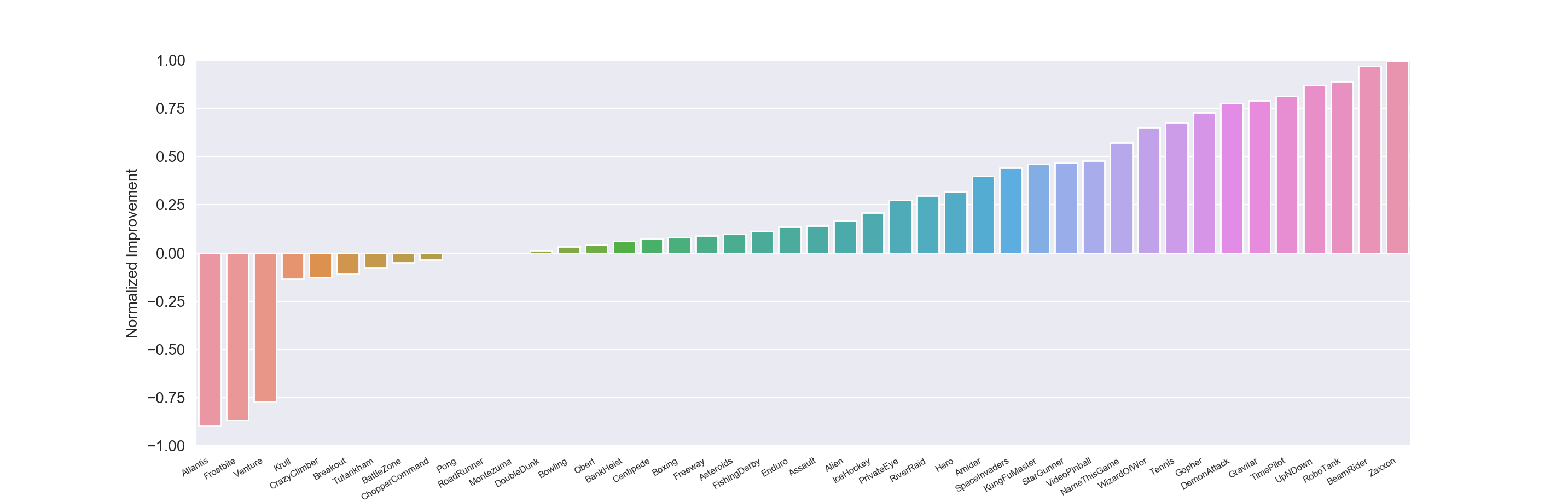}
    \caption{RIMs-PPO relative score improvement over LSTM-PPO baseline \citep{schulman2017proximal} across all Atari games averaged over 3 trials per game.  In both cases, PPO was used with the exact same settings, and the only change is the choice of recurrent architecture.    More detailed experiments with learning curves as well as comparisons with external baselines are in Appendix~\ref{appendix:detailedexp}.  }
    \label{fig:atari_rims4}
    \vspace{-3mm}
\end{figure*}

We have investigated how RIMs use specialization to improve generalization to changing important factors of variation in the data.  While these improvements have often been striking, it raises a question: what factors of variation should be changed between training and evaluation?  One setting where factors of variation change naturally is in reinforcement learning, as the data received from an environment changes as the agent learns and improves.  We conjecture that when applied to reinforcement learning, an agent using RIMs may be able to learn faster as its specialization leads to improved generalization to previously unseen aspects of the environment.  To investigate this we use an RL agent trained using Proximal Policy Optimization (PPO) \citep{schulman2017proximal} with a recurrent network producing the policy.  We employ an LSTM as a baseline, and compare results to the RIMs architecture.  This was a simple drop-in replacement and did not require changing any of the hyperparameters for PPO.  We experiment on the whole suite of Atari games and find that simply replacing the LSTM with RIMs greatly improves performance (Figure~\ref{fig:atari_rims4}).  

There is also an intriguing connection between the selective activation in RIMs and the concept of affordances from cognitive psychology \citep{gibson1977theory, cisek2010neural}.  To perform well in environments with a dynamic combination of risks and opportunities, an agent should be ready to adapt immediately, executing actions which are at least partially prepared.  This suggests agents should process sensory information in a contextual manner, building representations of potential actions that the environment currently affords.  For instance, in Demon Attack, one of the games where RIMs exhibit strong performance gains, the agent must quickly choose between targeting distant aliens to maximize points and avoiding fire from close-by aliens to avoid destruction (indeed both types of aliens are always present, but which is relevant depends on the player's position).  We hypothesize that in cases like this, selective activation of RIMs allows the agent to rapidly adapt its information processing to the types of actions relevant to the current context.  

\vspace{-1mm}
\subsection{Discussion and Ablations}

\textbf{Sparse Activation is necessary, but works for a wide range of hyperparameters: } On the copying task, we tried a wide variety of sparsity levels for different numbers of RIMs, and found that using a sparsity level between 30\% to 70\% performed optimally, suggesting that the sparsity hyperparameter is fairly flexible (refer to Table~\ref{tb:copy_ablation}, ~\ref{tb:adding_ablation} in appendix).  On Atari we found that using $k_A=5$ slightly improved over results compared with $k_A=4$, but both had similar performance across the vast majority of games.


\textbf{Input-attention is necessary: } We study the scenario where we remove the input attention process (i.e the top-down competition between different RIMs) but still allow the RIMs ot communicate with attention.  We found that this degraded results substantially on Atari but still outperformed the LSTM baseline. See  (Figure~\ref{fig:no_attention_rims_comm}) in appendix for more details.

\textbf{Communication between RIMs improves performance:} For copying and sequential MNIST, we performed an ablation where we remove the communication between RIMs and varied the number of RIMs and the number of activated RIMs (Refer to Table~\ref{tb:copy_ablation} in appendix.).  We found that the communication between RIMs is essential for good performance.

\label{sec:main_ablation}

\vspace{-3mm}
\section{Conclusion}

Many systems of interest comprise multiple dynamical processes that operate relatively independently and only occasionally have meaningful interactions.  Despite this, most machine learning models employ the opposite inductive bias, i.e., that all processes interact.  This can lead to poor generalization  and lack of robustness to changing task distributions.  We have proposed a new architecture, Recurrent Independent Mechanisms (RIMs), in which we learn multiple recurrent modules that are independent by default, but interact sparingly. For the purposes of this paper, we note that the notion of RIMs is not limited to the particular architecture employed here. The latter is used as a vehicle to assay and validate our overall hypothesis (cf.\ Appendix~\ref{sec:desiderata}), but better architectures for the RIMs model can likely be found.

\section*{Acknowledgements}
The authors acknowledge the important role played by their colleagues at Mila throughout the duration of this work.  The authors  would like to thank Mike Mozer for brainstorming session{s}. AG  would like to thank Matthew Botvinick, Charles Blundell, Greg Wayne for their useful feedback, which has improved the work tremendously. The authors are grateful to Nasim Rahaman for letting us  use video crop results. The authors  would also like to thank Rosemary Nan Ke, Stefan Bauer, Jonathan Binas, Min Lin, Disha Srivastava, Ali Farshchian,  Sarthak Mittal, Owen Xu, for useful discussions. The authors would also like to thank Shahab Bakhtiari, Kris Sankaran, Felix E. Leeb for proof reading the paper. The authors are grateful to NSERC, CIFAR, Google, Samsung, Nuance, IBM, Canada Research Chairs, Canada Graduate Scholarship Program, Nvidia for funding, and Compute Canada for computing resources.  We are very grateful to  Google for giving Google Cloud credits used in this project. This project was also known by  the name ``Blocks'' internally at Mila.





\bibliography{main}

\begin{thebibliography}{74}
\providecommand{\natexlab}[1]{#1}
\providecommand{\url}[1]{\texttt{#1}}
\expandafter\ifx\csname urlstyle\endcsname\relax
  \providecommand{\doi}[1]{doi: #1}\else
  \providecommand{\doi}{doi: \begingroup \urlstyle{rm}\Url}\fi

\bibitem[Andreas et~al.(2016)Andreas, Rohrbach, Darrell, and
  Klein]{andreas2016neural}
Jacob Andreas, Marcus Rohrbach, Trevor Darrell, and Dan Klein.
\newblock Neural module networks.
\newblock In \emph{Proceedings of the IEEE Conference on Computer Vision and
  Pattern Recognition}, pp.\  39--48, 2016.

\bibitem[Bahdanau et~al.(2014)Bahdanau, Cho, and Bengio]{bahdanau2014neural}
Dzmitry Bahdanau, Kyunghyun Cho, and Yoshua Bengio.
\newblock Neural machine translation by jointly learning to align and
  translate.
\newblock \emph{arXiv preprint arXiv:1409.0473}, 2014.

\bibitem[Battaglia et~al.(2018)Battaglia, Hamrick, Bapst, Sanchez-Gonzalez,
  Zambaldi, Malinowski, Tacchetti, Raposo, Santoro, Faulkner,
  et~al.]{battaglia2018relational}
Peter~W Battaglia, Jessica~B Hamrick, Victor Bapst, Alvaro Sanchez-Gonzalez,
  Vinicius Zambaldi, Mateusz Malinowski, Andrea Tacchetti, David Raposo, Adam
  Santoro, Ryan Faulkner, et~al.
\newblock Relational inductive biases, deep learning, and graph networks.
\newblock \emph{arXiv preprint arXiv:1806.01261}, 2018.

\bibitem[Baum \& Haussler(1989)Baum and Haussler]{baum1989size}
Eric~B Baum and David Haussler.
\newblock What size net gives valid generalization?
\newblock In \emph{Advances in neural information processing systems}, pp.\
  81--90, 1989.

\bibitem[Bengio(2017)]{bengio2017consciousness}
Yoshua Bengio.
\newblock The consciousness prior.
\newblock \emph{arXiv preprint arXiv:1709.08568}, 2017.

\bibitem[Bengio et~al.(2019)Bengio, Deleu, Rahaman, Ke, Lachapelle, Bilaniuk,
  Goyal, and Pal]{bengio2019meta}
Yoshua Bengio, Tristan Deleu, Nasim Rahaman, Rosemary Ke, S{\'e}bastien
  Lachapelle, Olexa Bilaniuk, Anirudh Goyal, and Christopher Pal.
\newblock A meta-transfer objective for learning to disentangle causal
  mechanisms.
\newblock \emph{arXiv:1901.10912}, 2019.

\bibitem[Bottou \& Gallinari(1991)Bottou and Gallinari]{bottou1991framework}
L{\'e}on Bottou and Patrick Gallinari.
\newblock A framework for the cooperation of learning algorithms.
\newblock In \emph{Advances in neural information processing systems}, pp.\
  781--788, 1991.

\bibitem[Botvinick \& Braver(2015)Botvinick and
  Braver]{botvinick2015motivation}
Matthew Botvinick and Todd Braver.
\newblock Motivation and cognitive control: from behavior to neural mechanism.
\newblock \emph{Annual review of psychology}, 66, 2015.

\bibitem[Bronstein et~al.(2017)Bronstein, Bruna, LeCun, Szlam, and
  Vandergheynst]{bronstein2017geometric}
Michael~M Bronstein, Joan Bruna, Yann LeCun, Arthur Szlam, and Pierre
  Vandergheynst.
\newblock Geometric deep learning: going beyond euclidean data.
\newblock \emph{IEEE Signal Processing Magazine}, 34\penalty0 (4):\penalty0
  18--42, 2017.

\bibitem[Burgess et~al.(2018)Burgess, Higgins, Pal, Matthey, Watters,
  Desjardins, and Lerchner]{burgess2018understanding}
Christopher~P Burgess, Irina Higgins, Arka Pal, Loic Matthey, Nick Watters,
  Guillaume Desjardins, and Alexander Lerchner.
\newblock Understanding disentangling in $\beta$-vae.
\newblock \emph{arXiv preprint arXiv:1804.03599}, 2018.

\bibitem[Chen et~al.(2018)Chen, Li, Grosse, and Duvenaud]{chen2018isolating}
Ricky~TQ Chen, Xuechen Li, Roger~B Grosse, and David~K Duvenaud.
\newblock Isolating sources of disentanglement in variational autoencoders.
\newblock In \emph{Advances in Neural Information Processing Systems}, pp.\
  2610--2620, 2018.

\bibitem[Chevalier-Boisvert \& Willems(2018)Chevalier-Boisvert and
  Willems]{gym_minigrid}
Maxime Chevalier-Boisvert and Lucas Willems.
\newblock Minimalistic gridworld environment for openai gym.
\newblock \url{https://github.com/maximecb/gym-minigrid}, 2018.

\bibitem[Chevalier-Boisvert et~al.(2018)Chevalier-Boisvert, Bahdanau, Lahlou,
  Willems, Saharia, Nguyen, and Bengio]{chevalier2018babyai}
Maxime Chevalier-Boisvert, Dzmitry Bahdanau, Salem Lahlou, Lucas Willems,
  Chitwan Saharia, Thien~Huu Nguyen, and Yoshua Bengio.
\newblock Babyai: First steps towards grounded language learning with a human
  in the loop.
\newblock \emph{arXiv preprint arXiv:1810.08272}, 2018.

\bibitem[Chung et~al.(2015)Chung, Kastner, Dinh, Goel, Courville, and
  Bengio]{chung2015recurrent}
Junyoung Chung, Kyle Kastner, Laurent Dinh, Kratarth Goel, Aaron~C Courville,
  and Yoshua Bengio.
\newblock A recurrent latent variable model for sequential data.
\newblock In \emph{Advances in neural information processing systems}, pp.\
  2980--2988, 2015.

\bibitem[Chung et~al.(2016)Chung, Ahn, and Bengio]{chung2016hierarchical}
Junyoung Chung, Sungjin Ahn, and Yoshua Bengio.
\newblock Hierarchical multiscale recurrent neural networks.
\newblock \emph{arXiv preprint arXiv:1609.01704}, 2016.

\bibitem[Cisek \& Kalaska(2010)Cisek and Kalaska]{cisek2010neural}
Paul Cisek and John~F Kalaska.
\newblock Neural mechanisms for interacting with a world full of action
  choices.
\newblock \emph{Annual review of neuroscience}, 33:\penalty0 269--298, 2010.

\bibitem[Denton \& Fergus(2018)Denton and Fergus]{denton2018stochastic}
Emily Denton and Rob Fergus.
\newblock Stochastic video generation with a learned prior.
\newblock \emph{arXiv preprint arXiv:1802.07687}, 2018.

\bibitem[Desimone \& Duncan(1995)Desimone and Duncan]{Desimone1995}
Robert Desimone and Jody Duncan.
\newblock Neural mechanisms of selective visual attention.
\newblock \emph{Annual Review of Neuroscience}, 18:\penalty0 193--222, 1995.

\bibitem[Dickinson(1985)]{Dickinson67}
A.~Dickinson.
\newblock Actions and habits: the development of behavioural autonomy.
\newblock \emph{Philosophical Transactions of the Royal Society B: Biological
  Sciences}, 308\penalty0 (1135):\penalty0 67--78, 1985.
\newblock ISSN 0080-4622.
\newblock \doi{10.1098/rstb.1985.0010}.

\bibitem[El~Hihi \& Bengio(1996)El~Hihi and Bengio]{el1996hierarchical}
Salah El~Hihi and Yoshua Bengio.
\newblock Hierarchical recurrent neural networks for long-term dependencies.
\newblock In \emph{Advances in neural information processing systems}, pp.\
  493--499, 1996.

\bibitem[Fernando et~al.(2017)Fernando, Banarse, Blundell, Zwols, Ha, Rusu,
  Pritzel, and Wierstra]{fernando2017pathnet}
Chrisantha Fernando, Dylan Banarse, Charles Blundell, Yori Zwols, David Ha,
  Andrei~A Rusu, Alexander Pritzel, and Daan Wierstra.
\newblock Pathnet: Evolution channels gradient descent in super neural
  networks.
\newblock \emph{arXiv preprint arXiv:1701.08734}, 2017.

\bibitem[Gibson(1977)]{gibson1977theory}
James~J Gibson.
\newblock The theory of affordances.
\newblock \emph{Hilldale, USA}, 1\penalty0 (2), 1977.

\bibitem[Gilmer et~al.(2017)Gilmer, Schoenholz, Riley, Vinyals, and
  Dahl]{gilmer2017neural}
Justin Gilmer, Samuel~S Schoenholz, Patrick~F Riley, Oriol Vinyals, and
  George~E Dahl.
\newblock Neural message passing for quantum chemistry.
\newblock In \emph{Proceedings of the 34th International Conference on Machine
  Learning-Volume 70}, pp.\  1263--1272. JMLR. org, 2017.

\bibitem[Goyal et~al.(2019{\natexlab{a}})Goyal, Islam, Strouse, Ahmed,
  Botvinick, Larochelle, Levine, and Bengio]{goyal2019infobot}
Anirudh Goyal, Riashat Islam, Daniel Strouse, Zafarali Ahmed, Matthew
  Botvinick, Hugo Larochelle, Sergey Levine, and Yoshua Bengio.
\newblock Infobot: Transfer and exploration via the information bottleneck.
\newblock \emph{arXiv preprint arXiv:1901.10902}, 2019{\natexlab{a}}.

\bibitem[Goyal et~al.(2019{\natexlab{b}})Goyal, Sodhani, Binas, Peng, Levine,
  and Bengio]{goyal2019reinforcement}
Anirudh Goyal, Shagun Sodhani, Jonathan Binas, Xue~Bin Peng, Sergey Levine, and
  Yoshua Bengio.
\newblock Reinforcement learning with competitive ensembles of
  information-constrained primitives.
\newblock \emph{arXiv preprint arXiv:1906.10667}, 2019{\natexlab{b}}.

\bibitem[Graves et~al.(2014{\natexlab{a}})Graves, Wayne, and
  Danihelka]{graves2014neural}
Alex Graves, Greg Wayne, and Ivo Danihelka.
\newblock Neural turing machines.
\newblock \emph{arXiv preprint arXiv:1410.5401}, 2014{\natexlab{a}}.

\bibitem[Graves et~al.(2014{\natexlab{b}})Graves, Wayne, and
  Danihelka]{graves2014ntm}
Alex Graves, Greg Wayne, and Ivo Danihelka.
\newblock Neural turing machines.
\newblock \emph{CoRR}, abs/1410.5401, 2014{\natexlab{b}}.
\newblock URL \url{http://arxiv.org/abs/1410.5401}.

\bibitem[Graves et~al.(2016)Graves, Wayne, Reynolds, Harley, Danihelka,
  Grabska-Barwi{\'n}ska, Colmenarejo, Grefenstette, Ramalho, Agapiou,
  et~al.]{graves2016hybrid}
Alex Graves, Greg Wayne, Malcolm Reynolds, Tim Harley, Ivo Danihelka, Agnieszka
  Grabska-Barwi{\'n}ska, Sergio~G{\'o}mez Colmenarejo, Edward Grefenstette,
  Tiago Ramalho, John Agapiou, et~al.
\newblock Hybrid computing using a neural network with dynamic external memory.
\newblock \emph{Nature}, 538\penalty0 (7626):\penalty0 471, 2016.

\bibitem[Ha \& Schmidhuber(2018)Ha and Schmidhuber]{ha2018world}
David Ha and J{\"u}rgen Schmidhuber.
\newblock World models.
\newblock \emph{arXiv preprint arXiv:1803.10122}, 2018.

\bibitem[Hafner et~al.(2018)Hafner, Lillicrap, Fischer, Villegas, Ha, Lee, and
  Davidson]{hafner2018learning}
Danijar Hafner, Timothy Lillicrap, Ian Fischer, Ruben Villegas, David Ha,
  Honglak Lee, and James Davidson.
\newblock Learning latent dynamics for planning from pixels.
\newblock \emph{arXiv preprint arXiv:1811.04551}, 2018.

\bibitem[Henaff et~al.(2016)Henaff, Weston, Szlam, Bordes, and
  LeCun]{henaff2016tracking}
Mikael Henaff, Jason Weston, Arthur Szlam, Antoine Bordes, and Yann LeCun.
\newblock Tracking the world state with recurrent entity networks.
\newblock \emph{arXiv preprint arXiv:1612.03969}, 2016.

\bibitem[Higgins et~al.(2016)Higgins, Matthey, Pal, Burgess, Glorot, Botvinick,
  Mohamed, and Lerchner]{higgins2016beta}
Irina Higgins, Loic Matthey, Arka Pal, Christopher Burgess, Xavier Glorot,
  Matthew Botvinick, Shakir Mohamed, and Alexander Lerchner.
\newblock beta-vae: Learning basic visual concepts with a constrained
  variational framework.
\newblock 2016.

\bibitem[Hinton et~al.(2018)Hinton, Sabour, and Frosst]{hinton2018matrix}
Geoffrey~E Hinton, Sara Sabour, and Nicholas Frosst.
\newblock Matrix capsules with em routing.
\newblock 2018.

\bibitem[Hochreiter \& Schmidhuber(1997)Hochreiter and
  Schmidhuber]{hochreiter1997long}
Sepp Hochreiter and J{\"u}rgen Schmidhuber.
\newblock Long short-term memory.
\newblock \emph{Neural computation}, 9\penalty0 (8):\penalty0 1735--1780, 1997.

\bibitem[Jacobs et~al.(1991)Jacobs, Jordan, Nowlan, Hinton,
  et~al.]{jacobs1991adaptive}
Robert~A Jacobs, Michael~I Jordan, Steven~J Nowlan, Geoffrey~E Hinton, et~al.
\newblock Adaptive mixtures of local experts.
\newblock \emph{Neural computation}, 3\penalty0 (1):\penalty0 79--87, 1991.

\bibitem[Jernite et~al.(2016)Jernite, Grave, Joulin, and
  Mikolov]{jernite2016variable}
Yacine Jernite, Edouard Grave, Armand Joulin, and Tomas Mikolov.
\newblock Variable computation in recurrent neural networks.
\newblock \emph{arXiv preprint arXiv:1611.06188}, 2016.

\bibitem[Ke et~al.(2018)Ke, Goyal, Bilaniuk, Binas, Mozer, Pal, and
  Bengio]{ke2018sparse}
Nan~Rosemary Ke, Anirudh Goyal, Olexa Bilaniuk, Jonathan Binas, Michael~C
  Mozer, Chris Pal, and Yoshua Bengio.
\newblock Sparse attentive backtracking: Temporal credit assignment through
  reminding.
\newblock In \emph{Advances in Neural Information Processing Systems}, pp.\
  7640--7651, 2018.

\bibitem[Kingma \& Ba(2014)Kingma and Ba]{Kingma2014}
Diederik Kingma and Jimmy Ba.
\newblock Adam: A method for stochastic optimization.
\newblock \emph{arXiv preprint arXiv:1412.6980}, 2014.

\bibitem[Kipf et~al.(2018)Kipf, Fetaya, Wang, Welling, and
  Zemel]{kipf2018neural}
Thomas Kipf, Ethan Fetaya, Kuan-Chieh Wang, Max Welling, and Richard Zemel.
\newblock Neural relational inference for interacting systems.
\newblock \emph{arXiv preprint arXiv:1802.04687}, 2018.

\bibitem[Kirsch et~al.(2018)Kirsch, Kunze, and Barber]{kirsch2018modular}
Louis Kirsch, Julius Kunze, and David Barber.
\newblock Modular networks: Learning to decompose neural computation.
\newblock In \emph{Advances in Neural Information Processing Systems}, pp.\
  2408--2418, 2018.

\bibitem[Kool \& Botvinick(2018)Kool and Botvinick]{kool2018mental}
Wouter Kool and Matthew Botvinick.
\newblock Mental labour.
\newblock \emph{Nature human behaviour}, 2\penalty0 (12):\penalty0 899--908,
  2018.

\bibitem[Kostrikov(2018)]{pytorchrl}
Ilya Kostrikov.
\newblock Pytorch implementations of reinforcement learning algorithms.
\newblock \url{https://github.com/ikostrikov/pytorch-a2c-ppo-acktr-gail}, 2018.

\bibitem[Koutnik et~al.(2014)Koutnik, Greff, Gomez, and
  Schmidhuber]{koutnik2014clockwork}
Jan Koutnik, Klaus Greff, Faustino Gomez, and Juergen Schmidhuber.
\newblock A clockwork rnn.
\newblock \emph{arXiv preprint arXiv:1402.3511}, 2014.

\bibitem[Krueger et~al.(2016)Krueger, Maharaj, Kram{\'a}r, Pezeshki, Ballas,
  Ke, Goyal, Bengio, Courville, and Pal]{krueger2016zoneout}
David Krueger, Tegan Maharaj, J{\'a}nos Kram{\'a}r, Mohammad Pezeshki, Nicolas
  Ballas, Nan~Rosemary Ke, Anirudh Goyal, Yoshua Bengio, Aaron Courville, and
  Chris Pal.
\newblock Zoneout: Regularizing rnns by randomly preserving hidden activations.
\newblock \emph{arXiv preprint arXiv:1606.01305}, 2016.

\bibitem[Li et~al.(2018)Li, Li, Cook, Zhu, and Gao]{li2018independently}
Shuai Li, Wanqing Li, Chris Cook, Ce~Zhu, and Yanbo Gao.
\newblock Independently recurrent neural network (indrnn): Building a longer
  and deeper rnn.
\newblock In \emph{Proceedings of the IEEE Conference on Computer Vision and
  Pattern Recognition}, pp.\  5457--5466, 2018.

\bibitem[Neil et~al.(2016)Neil, Pfeiffer, and Liu]{neil2016phased}
Daniel Neil, Michael Pfeiffer, and Shih-Chii Liu.
\newblock Phased lstm: Accelerating recurrent network training for long or
  event-based sequences.
\newblock In \emph{Advances in neural information processing systems}, pp.\
  3882--3890, 2016.

\bibitem[Parascandolo et~al.(2018)Parascandolo, Kilbertus, Rojas-Carulla, and
  Sch{\"o}lkopf]{ParKilRojSch18}
Giambattista Parascandolo, Niki Kilbertus, Mateo Rojas-Carulla, and Bernhard
  Sch{\"o}lkopf.
\newblock Learning independent causal mechanisms.
\newblock In \emph{Proceedings of the 35th International Conference on Machine
  Learning (ICML)}, pp.\  4033--4041, 2018.
\newblock URL \url{http://proceedings.mlr.press/v80/parascandolo18a.html}.

\bibitem[Pearl(2009)]{Pearl2009}
Judea Pearl.
\newblock \emph{Causality: Models, Reasoning, and Inference}.
\newblock Cambridge University Press, New York, NY, 2nd edition, 2009.

\bibitem[Peters et~al.(2017)Peters, Janzing, and Sch{\"o}lkopf]{PetJanSch17}
Jonas Peters, Dominik Janzing, and Bernhard Sch{\"o}lkopf.
\newblock \emph{Elements of Causal Inference - Foundations and Learning
  Algorithms}.
\newblock MIT Press, Cambridge, MA, USA, 2017.
\newblock ISBN 978-0-262-03731-0.

\bibitem[Raposo et~al.(2017)Raposo, Santoro, Barrett, Pascanu, Lillicrap, and
  Battaglia]{raposo2017discovering}
David Raposo, Adam Santoro, David Barrett, Razvan Pascanu, Timothy Lillicrap,
  and Peter Battaglia.
\newblock Discovering objects and their relations from entangled scene
  representations.
\newblock \emph{arXiv preprint arXiv:1702.05068}, 2017.

\bibitem[Reed \& De~Freitas(2015)Reed and De~Freitas]{reed2015neural}
Scott Reed and Nando De~Freitas.
\newblock Neural programmer-interpreters.
\newblock \emph{arXiv preprint arXiv:1511.06279}, 2015.

\bibitem[Ronco et~al.(1997)Ronco, Gollee, and Gawthrop]{ronco1996modular}
Eric Ronco, Henrik Gollee, and Peter~J Gawthrop.
\newblock Modular neural networks and self-decomposition.
\newblock \emph{Technical Report CSC-96012}, 1997.

\bibitem[Rosenbaum et~al.(2017)Rosenbaum, Klinger, and
  Riemer]{rosenbaum2017routing}
Clemens Rosenbaum, Tim Klinger, and Matthew Riemer.
\newblock Routing networks: Adaptive selection of non-linear functions for
  multi-task learning.
\newblock \emph{arXiv preprint arXiv:1711.01239}, 2017.

\bibitem[Rosenbaum et~al.(2019)Rosenbaum, Cases, Riemer, and
  Klinger]{rosenbaum2019routing}
Clemens Rosenbaum, Ignacio Cases, Matthew Riemer, and Tim Klinger.
\newblock Routing networks and the challenges of modular and compositional
  computation.
\newblock \emph{arXiv preprint arXiv:1904.12774}, 2019.

\bibitem[Rusu et~al.(2016)Rusu, Rabinowitz, Desjardins, Soyer, Kirkpatrick,
  Kavukcuoglu, Pascanu, and Hadsell]{rusu2016progressive}
Andrei~A Rusu, Neil~C Rabinowitz, Guillaume Desjardins, Hubert Soyer, James
  Kirkpatrick, Koray Kavukcuoglu, Razvan Pascanu, and Raia Hadsell.
\newblock Progressive neural networks.
\newblock \emph{arXiv preprint arXiv:1606.04671}, 2016.

\bibitem[Sabour et~al.(2017)Sabour, Frosst, and Hinton]{sabour2017dynamic}
Sara Sabour, Nicholas Frosst, and Geoffrey~E Hinton.
\newblock Dynamic routing between capsules.
\newblock In \emph{Advances in neural information processing systems}, pp.\
  3856--3866, 2017.

\bibitem[Santoro et~al.(2017)Santoro, Raposo, Barrett, Malinowski, Pascanu,
  Battaglia, and Lillicrap]{santoro2017simple}
Adam Santoro, David Raposo, David~G Barrett, Mateusz Malinowski, Razvan
  Pascanu, Peter Battaglia, and Timothy Lillicrap.
\newblock A simple neural network module for relational reasoning.
\newblock In \emph{Advances in neural information processing systems}, pp.\
  4967--4976, 2017.

\bibitem[Santoro et~al.(2018)Santoro, Faulkner, Raposo, Rae, Chrzanowski,
  Weber, Wierstra, Vinyals, Pascanu, and Lillicrap]{santoro2018relational}
Adam Santoro, Ryan Faulkner, David Raposo, Jack~W. Rae, Mike Chrzanowski,
  Theophane Weber, Daan Wierstra, Oriol Vinyals, Razvan Pascanu, and Timothy~P.
  Lillicrap.
\newblock Relational recurrent neural networks.
\newblock \emph{CoRR}, abs/1806.01822, 2018.
\newblock URL \url{http://arxiv.org/abs/1806.01822}.

\bibitem[Scarselli et~al.(2008)Scarselli, Gori, Tsoi, Hagenbuchner, and
  Monfardini]{scarselli2008graph}
Franco Scarselli, Marco Gori, Ah~Chung Tsoi, Markus Hagenbuchner, and Gabriele
  Monfardini.
\newblock The graph neural network model.
\newblock \emph{IEEE Transactions on Neural Networks}, 20\penalty0
  (1):\penalty0 61--80, 2008.

\bibitem[Schmidhuber(2018)]{schmidhuber2018one}
J{\"u}rgen Schmidhuber.
\newblock One big net for everything.
\newblock \emph{arXiv preprint arXiv:1802.08864}, 2018.

\bibitem[Sch{\"o}lkopf et~al.(2012)Sch{\"o}lkopf, Janzing, Peters, Sgouritsa,
  Zhang, and Mooij]{SchJanPetSgoetal12}
Bernhard Sch{\"o}lkopf, Dominik Janzing, Jonas Peters, Eleni Sgouritsa, Kun
  Zhang, and Joris Mooij.
\newblock On causal and anticausal learning.
\newblock In J.~Langford and J.~Pineau (eds.), \emph{Proceedings of the 29th
  International Conference on Machine Learning (ICML)}, pp.\  1255--1262, New
  York, NY, USA, 2012. Omnipress.

\bibitem[Schulman et~al.(2017)Schulman, Wolski, Dhariwal, Radford, and
  Klimov]{schulman2017proximal}
John Schulman, Filip Wolski, Prafulla Dhariwal, Alec Radford, and Oleg Klimov.
\newblock Proximal policy optimization algorithms.
\newblock \emph{arXiv preprint arXiv:1707.06347}, 2017.

\bibitem[Shazeer et~al.(2017)Shazeer, Mirhoseini, Maziarz, Davis, Le, Hinton,
  and Dean]{shazeer2017outrageously}
Noam Shazeer, Azalia Mirhoseini, Krzysztof Maziarz, Andy Davis, Quoc Le,
  Geoffrey Hinton, and Jeff Dean.
\newblock Outrageously large neural networks: The sparsely-gated
  mixture-of-experts layer.
\newblock \emph{arXiv preprint arXiv:1701.06538}, 2017.

\bibitem[Simon(1991)]{simon1991architecture}
Herbert~A Simon.
\newblock The architecture of complexity.
\newblock In \emph{Facets of systems science}, pp.\  457--476. Springer, 1991.

\bibitem[Sodhani et~al.(2019)Sodhani, Goyal, Deleu, Bengio, Levine, and
  Tang]{sodhani2019learning}
Shagun Sodhani, Anirudh Goyal, Tristan Deleu, Yoshua Bengio, Sergey Levine, and
  Jian Tang.
\newblock Learning powerful policies by using consistent dynamics model.
\newblock \emph{arXiv preprint arXiv:1906.04355}, 2019.

\bibitem[Tacchetti et~al.(2018)Tacchetti, Song, Mediano, Zambaldi, Rabinowitz,
  Graepel, Botvinick, and Battaglia]{tacchetti2018relational}
Andrea Tacchetti, H~Francis Song, Pedro~AM Mediano, Vinicius Zambaldi, Neil~C
  Rabinowitz, Thore Graepel, Matthew Botvinick, and Peter~W Battaglia.
\newblock Relational forward models for multi-agent learning.
\newblock \emph{arXiv preprint arXiv:1809.11044}, 2018.

\bibitem[Teh et~al.(2017)Teh, Bapst, Czarnecki, Quan, Kirkpatrick, Hadsell,
  Heess, and Pascanu]{teh2017distral}
Yee Teh, Victor Bapst, Wojciech~M Czarnecki, John Quan, James Kirkpatrick, Raia
  Hadsell, Nicolas Heess, and Razvan Pascanu.
\newblock Distral: Robust multitask reinforcement learning.
\newblock In \emph{Advances in Neural Information Processing Systems}, pp.\
  4496--4506, 2017.

\bibitem[Todorov et~al.(2012)Todorov, Erez, and Tassa]{todorov2012mujoco}
Emanuel Todorov, Tom Erez, and Yuval Tassa.
\newblock Mujoco: A physics engine for model-based control.
\newblock In \emph{2012 IEEE/RSJ International Conference on Intelligent Robots
  and Systems}, pp.\  5026--5033. IEEE, 2012.

\bibitem[Van~Steenkiste et~al.(2018)Van~Steenkiste, Chang, Greff, and
  Schmidhuber]{van2018relational}
Sjoerd Van~Steenkiste, Michael Chang, Klaus Greff, and J{\"u}rgen Schmidhuber.
\newblock Relational neural expectation maximization: Unsupervised discovery of
  objects and their interactions.
\newblock \emph{arXiv preprint arXiv:1802.10353}, 2018.

\bibitem[Vaswani et~al.(2017)Vaswani, Shazeer, Parmar, Uszkoreit, Jones, Gomez,
  Kaiser, and Polosukhin]{vaswani2017attention}
Ashish Vaswani, Noam Shazeer, Niki Parmar, Jakob Uszkoreit, Llion Jones,
  Aidan~N Gomez, {\L}ukasz Kaiser, and Illia Polosukhin.
\newblock Attention is all you need.
\newblock In \emph{Advances in neural information processing systems}, pp.\
  5998--6008, 2017.

\bibitem[von Helmholtz(1867)]{von1867handbuch}
H.~L.~F. von Helmholtz.
\newblock \emph{Handbuch der physiologischen Optik}, volume III.
\newblock Voss, 1867.

\bibitem[von Holst \& Mittelstaedt(1950)von Holst and
  Mittelstaedt]{vonHolst1950}
Erich von Holst and Horst Mittelstaedt.
\newblock Das reafferenzprinzip.
\newblock \emph{Naturwissenschaften}, 37\penalty0 (20):\penalty0 464--476, Jan
  1950.
\newblock \doi{10.1007/BF00622503}.

\bibitem[Watters et~al.(2017)Watters, Zoran, Weber, Battaglia, Pascanu, and
  Tacchetti]{watters2017visual}
Nicholas Watters, Daniel Zoran, Theophane Weber, Peter Battaglia, Razvan
  Pascanu, and Andrea Tacchetti.
\newblock Visual interaction networks: Learning a physics simulator from video.
\newblock In \emph{Advances in neural information processing systems}, pp.\
  4539--4547, 2017.

\bibitem[Williams \& Zipser(1989)Williams and Zipser]{williams1989learning}
Ronald~J Williams and David Zipser.
\newblock A learning algorithm for continually running fully recurrent neural
  networks.
\newblock \emph{Neural computation}, 1\penalty0 (2):\penalty0 270--280, 1989.

\end{thebibliography}
\bibliographystyle{iclr2021_conference}

\small

\clearpage
\appendix

\onecolumn

\addcontentsline{toc}{section}{Appendix} 
\part{Appendix} 
\parttoc 

\section{Desiderata for Recurrent Independent Mechanisms}
\label{sec:desiderata}

We have laid out a case for building models composed of modules which by default operate independently and can interact in a limited manner. 
Accordingly, our approach to modelling the dynamics of the world starts by dividing the overall model into small subsystems (or modules), referred to as \textit{Recurrent Independent Mechanisms (RIMs)}, with distinct functions learned automatically from data. 
Our model encourages sparse interaction, i.e.,  we want most RIMs to operate independently and follow their default dynamics most of the time, only rarely sharing information. Below, we lay out desiderata for modules to capture modular dynamics with sparse interactions.

\paragraph{Competitive Mechanisms:}
Inspired by the observations in the main paper, we propose that RIMs utilize competition to allocate representational and computational resources. As argued by \citep{ParKilRojSch18}, this tends to produce independence among learned mechanisms if the training data has been generated by independent physical mechanisms.

\paragraph{Top Down Attention:} 

The points mentioned in Section~\ref{sec:rimsmodel} in principle pertain to synthetic and natural intelligent systems alike. Hence, it is not surprising that they also appear in neuroscience. For instance, suppose we are looking for a particular object in a large scene, using limited processing capacity.
The \emph{biased competition theory} of selective attention conceptualizes basic findings of experimental psychology and neuroscience \citep{Desimone1995}:
our capacity of parallel processing of and reasoning with high-level concepts is limited, and many brain systems representing visual information use competition to allocate resources.  Competitive interactions among multiple objects  occur automatically and operate in parallel across the visual field. Second, the principle of selectivity amounts to the idea that a perceiver has the ability to filter out unwanted information and selectively \emph{process} the rest of the information. Third,  \emph{top-down bias} originating from higher brain areas enables us to selectively devote resources to input information that may be of particular interest or relevance. This may be accomplished by units matching the internal model of an object or process of interest being pre-activated and thus gaining an advantage during the competition of brain mechanisms. 




\paragraph{Sparse Information Flow:} Each RIM's dynamics should only be affected by RIMs which are deemed relevant. The fundamental challenge is centered around establishing sensible communication between modules.  In the presence of noisy or distracting information, a large subset of RIMs should stay dormant, and not be affected by the noise. This way, training an ensemble of these RIMs can be more robust to out-of-distribution or distractor observations than training one big homogeneous neural network \citep{schmidhuber2018one}.

\paragraph{Modular Computation Flow and Modular Parameterization:} Each RIM should have its own dynamics operating by \textit{default}, in the absence of interaction with other RIMs.
The total number of parameters (i.e. weights) can be reduced since the RIMs can specialize on simple sub-problems, similar to \citep{ParKilRojSch18}.
This can speed up computation and improve the generalisation ability of the system \citep{baum1989size}. The individuals RIMs in the ensemble should be simple also to prevent individual RIMs from dominating and modelling complex, composite mechanisms. We refer to a parameterization as modular if most parameters are associated to individuals RIMs only. This has the desirable property that a RIM should maintain its own independent functionality even as other RIMs are changed (due to its behavior being determined by its own self-contained parameters).

\section{Extended Related Work}
\label{sec:extended_related_work}
\begin{table*}[bht]
{\renewcommand{\arraystretch}{1.3}
\caption{A concise comparison of recurrent models with modular memory.  }
\label{tb:related}
\scalebox{0.8}{
\begin{tabular}{|l|l|l|l|l|}
\hline
Method / Property         & \makecell{Modular\\Memory} & \makecell{Sparse \\Information Flow} & \makecell{Modular \\Computation Flow}     & \makecell{Modular\\Parameterization} \\ \hline
LSTM / RNN & \xmark & \xmark & \xmark & \xmark \\ \hline
Relational RNN \citep{santoro2018relational}  & \cmark   & \xmark & \cmark & \xmark \\ \hline
NTM \citep{graves2014ntm} & \cmark & \cmark & \xmark & \xmark \\ \hline
SAB\citep{ke2018sparse}  & \xmark & \cmark  & \xmark & \xmark  \\ \hline
IndRNN\citep{li2018independently} & \cmark & \xmark & \xmark & \cmark  \\ \hline
RIMs    & \cmark & \cmark  & \cmark & \cmark \\ \hline
\end{tabular}
}
}
\end{table*}

The present section provides further details on related work, thus extending Section~\ref{sec:related_work}.

\textbf{Neural Turing Machine (NTM)}. The NTM \citep{graves2014neural} has a Turing machine inspired memory with a sequence of independent memory cells, and uses an attention mechanism to move heads over the cells while performing targeted read and write operations.  This shares a key idea with RIMs: that input information should only impact a sparse subset of the memory by default, while keeping most of the memory unaltered.  The RIM model introduces the idea that each RIM has its own independent dynamics, whereas the mechanism for updating memory cells update is shared.

\textbf{Relational RNN}.  The Relational Models paper \citep{santoro2018relational} is based on the idea of using a multi-head attention mechanism to share information between multiple parts of memory.  It is related to our idea but a key difference is that we encourage the RIMs to remain separate as much as possible, whereas \citep{santoro2018relational} allows information between the parts to flow on each step (in effect making the part distribution only relevant to a particular step).  Additionally, RIMs has the notion of each RIM having its own independent transition dynamics which operate by default, whereas the Relational RNN only does computation and updating of the memory using attention.  

\textbf{Sparse Attentive Backtracking (SAB)}.  The SAB architecture \citep{ke2018sparse} explores RNNs with self-attention across time steps as well as variants where the attention is sparse in the forward pass and where the gradient is sparse in the backward pass.  It shares the motivation of using sparse attention to keep different pieces of information separated, but differs from the RIMs model in that it considers separation between time steps rather than separation between RIMs.

\textbf{Independently Recurrent Neural Network (IndRNN)}.  The IndRNN \citep{li2018independently} replaces the full transition matrix in a vanilla RNN (between time steps) to a diagonal transition weight matrix.  In other words, each recurrent unit has completely independent dynamics.  Intriguingly they show that this gives much finer control over the gating of information, and allows for such an RNN to learn long-term dependencies without vanishing or exploding gradients. Analysis of the gradients shows that having smaller recurrent transition matrices mitigates the vanishing and exploding gradient issue. This may provide further explanation for why RIMs perform well on long sequences.


\textbf{Consciousness Prior} \citep{bengio2017consciousness}: This is based on the inductive bias of a sparse graphical model describing the interactions between high-level variables, each factor involving few variables and capaturing an independent mechanism, and using attention to select only a subset of high-level variables to interact together at any particular time during inference.  This  motivates in RIMs the use of modules (corresponding to factors)
which dynamically select which variables (instances) they apply to, with
modules activated sparsely (only those factors which need to interact)
and communicating sparsely (only a few variables, i.e., heads, involved
for each). Our paper thus helps to validate the inductive bias of the consciousness prior.

\textbf{Recurrent Entity Networks}:   EnTNet \citep{henaff2016tracking} can be viewed as a set of separate recurrent models whose hidden states store the memory
slots. These hidden states are either fixed by the gates, or modified through a simple RNN-style
update. Moreover,  EntNet uses an independent gate for writing to each memory slot. Our work is related in the sense that we also have different recurrent models (i.e.,RIMs, though each RIM has different parameters), but we allow the RIMs to communicate with each other sparingly using an attention mechanism.


\textbf{Capsules and Dynamic Routing:} EM Capsules \citep{hinton2018matrix} and the preceding Dynamic Capsules \citep{sabour2017dynamic} use the poses of parts and learned part $\rightarrow$ object relationships to vote for the poses of objects. When multiple parts cast very similar votes, the object is assumed to be present, which is facilitated by an interactive inference (routing) algorithm.

\textbf{Relational Graph Based Methods:} Recent graph-based architectures have studied combinatorial generalization in the context of modeling dynamical systems like physics
simulation, multi-object scenes,  and motion-capture data, and multiagent systems \citep{scarselli2008graph, bronstein2017geometric, watters2017visual, raposo2017discovering,santoro2017simple, gilmer2017neural, van2018relational, kipf2018neural, battaglia2018relational, tacchetti2018relational}. One can also view our proposed model as a relational graph neural network, where nodes are parameterized as individual RIM and edges are parameterized by the attention mechanism. Though, it is important to emphasize that semantics of the nodes is learned and that the topology of the graph induced in the proposed model is dynamic, while in most graph neural networks the nodes have a meaning directly tied to the inputs and targets and the topology is fixed and given. 

\textbf{Default Behaviour:} Our work is also related to work in behavioural research that deals with two modes of decision making \citep{Dickinson67,  botvinick2015motivation, kool2018mental}: an automatic system that relies on habits and a controlled system that uses some privileged information for decision-making. The proposed model also has two modes of input processing: activated RIMs rely on external sensory information, and hence seem analogous to the controlled system, while inactive RIMs may more loosely correspond to computation taking place outside of conscious processing and corresponding to the automatic or habit system. There is RL work aiming to learn \textit{default policies}, which have shown to improve transfer and generalization in multi-task RL \citep{teh2017distral, goyal2019infobot}. RIMs are different in the sense that we are not trying to learn \textit{default policies} which affect the environment, instead we want to learn mechanisms, which are more about analyzing or understanding the environment. State-dependent activation of different primitive policies was also studied by \citet{goyal2019reinforcement}, and the authors showed that they can learn different primitives, but they also consider that only a single primitive can be active at a particular time-step. Also, note that primitive policies try to \textit{affect} the environment, whereas mechanisms try to \textit{process or understand} the environment. 

\section{Model Setup Details and Hyperparameters}
\label{appendix:detailedexp}

\subsection{RIMs Implementation}

The RIMs model consists of three main components: the input attention, the process for selecting activated RIMs, and the communication between RIMs.  The input attention closely follows the attention mechanism of \citep{santoro2018relational} but with a significant modification: that all of the weights within the attention mechanism are separate per-block.  Thus we remove the normal linear layers and replace them with a batch matrix multiplication over the RIMs (as each block has its own weight matrix). Note that the read-key (or query) is a function of the hidden state of each RIM.  

For selecting activated RIMs, we compute the top-k attention weight over the RIMs (based on the query).  We then select the activated RIMs, using a mask which zeroes out the outputs from inactive RIMs. We compute the independent dynamics over all RIMs by using a separate LSTM for each RIM.  Following this, we compute the communication between RIMs as a multihead attention \citep{santoro2018relational}, with the earlier-discussed modification of having separate weight parameters for each block, and also that we added a skip-connection around the attention mechanism.  This attention mechanism used 4 heads and in general used a key size and value size of 32. We computed the updates for all RIMs but used the activated-block mask to selectively update only the activated subset of the RIMs.  

The use of RIMs introduces two additional hyperparameters over an LSTM/GRU: the number of RIMs and the number of activated RIMs per step.  We also observed that having too few activated RIMs tends to hurt optimization and having too many activated RIMs attenuates the improvements to generalization.  For the future it would be interesting to explore dynamic ways of controlling how many RIMs to activate.  

\textbf{Multiple Heads:} Analogously to \citet{vaswani2017attention, santoro2018relational}, we use multiple heads both for communication between RIMs as well as input attention (as in Sec \ref{sec:selectiveactivationrims}) by producing different sets of queries, keys, and values to compute a linear transformation for each head (different heads have different parameters), and then applying the attention operator for each head separately in order to select conditioning inputs for the RIMs.

\textbf{Selective Activation.} In order to selectively decide which set of RIMs to activate, we also tried appending a vector of zeros to the input representation, and the RIMs, which pay least attention to the null input (in the input attention) are activated.  

\subsection{Detailed Model Hyperparameters}

Table \ref{table::appendix::minigrid::hyperparameters} lists the different hyperparameters.
\begin{table}[tbh]
    \caption{Hyperparameters}
  \begin{center}
    \begin{tabular}{lr}
      \toprule
      Parameter & Value  \\
      \midrule
     
      Optimizer                                   & Adam\citep{Kingma2014}\\
      learning rate                                   & $7\cdot 10^{-4}$      \\
      batch size                                      & 64      \\
     Input keys &  64 \\
     Input Values & Size of individual RIM * 4 \\
     Input Heads & 4 \\ 
     Input Dropout & 0.1 \\
     Communication keys &  32 \\
     Communication Values & 32 \\
     Communication heads & 4 \\ 
     Communication Dropout & 0.1 \\
      \bottomrule
    \end{tabular}
  \end{center}
  \label{table::appendix::minigrid::hyperparameters}
\end{table}

\subsection{Other Architectural Changes that we Explored}

We have not conducted systematic optimizations of the proposed architecture. We believe that even principled hyperparameter tuning may significantly improve performance for many of the tasks we have considered in the paper. We briefly mention a few architectural changes which we have studied:  
\begin{itemize}
    \item On the output side, we concatenate the representations of the different RIMs, and use the concatenated representation for learning a policy (in RL experiments) or for predicting the input at the next time step (for bouncing balls as well as all other experiments). We empirically found that adding another layer of (multi-headed) key-value attention on the output seems to improve the results. We have not included this change in the RIMs implementation of this paper.
    \item In our experiments, we shared the same decoder for all the RIMs, i.e., we concatenate the representations of different RIMS, and feed the concatenated representations to the decoder. In the future it would be interesting to think of ways to allow a more ``structured'' decoder. The reason for this is that even if the RIMs generalize to new environments, the shared decoder can fail to do so. So changing the structure of decoder could be helpful.
    \item For the RL experiments, we also tried providing the previous actions, rewards, language instruction as input to decide the activation of RIMs. This is consistent with the idea of  \emph{efference copies} as proposed by \cite{von1867handbuch, vonHolst1950}, i.e., using copies of motor signals as inputs. Preliminary experiments shows that this improves the performance in Atari games.
\end{itemize}

\section{Experiment Details}

\subsection{Effect of Varying Number and Active RIMs on Copying Task}
\label{table:copying_ablation}
We used a learning rate of 0.001 with the Adam Optimizer and trained each model for 150 epochs (unless the model was stuck, we found that this was enough to bring the training error close to zero).  For the RIMs model we used 600 units split across 6 RIMs (100 units per block).  For the LSTM we used a total of 600 units.  We did not explore this extensively but we qualitatively found that the results on copying were not very sensitive to the exact number of units.  

The sequences to be copied first have 10 random digits (from 0-8), then a span of zeros of some length, followed by a special indicator ``9'' in the input which instructs the model to begin outputting the copied sequence.  


In our experiments, we trained the models with ``zero spans'' of length 50 and evaluated on the model with ``zero spans'' of length 200.    We note that all the ablations were run with the default parameters (i.e number of keys, values as for RIMs model) for 100 epochs.
Tab. \ref{tb:copy_ablation} shows the effect of two baselines as compared to the RIMs model (a) When we allow the input attention for activation of different RIMs but we dont allow different RIMs to communicate. (b) No Input attention, but we allow different RIMs to communicate with each other. Tab. \ref{tb:copy_ablation} shows that the proposed method is better than both of these baselines. For copy task, we used 1 head in input attention, and 4 heads for RIMs communication. We note that even with 1 RIM, its not exactly same as a LSTM, because each RIM can still reference itself.

\begin{table*}[htb!]
\centering
{\renewcommand{\arraystretch}{1.0}
\caption{\textbf{Error (CE for last 10 time steps) on the copying task}.  Note that while all of the methods are able to learn to copy on the length seen during training, the RIMs model generalizes to sequences longer than those seen during training whereas the LSTM fails catastrophically.  }
\label{tb:copy_ablation}
\centering
\scalebox{0.9}{
\begin{tabular}{lrrr} 
\toprule
Approach & Train Length 50 & Test Length 200  \\ 
\midrule

\textbf{RIMs} & \textbf{0.00} & \textbf{0.00} \\ 

\midrule
With input Attention and No Communication & & \\
\midrule
\textbf{RIMs} ($k_T$ = 4, $k_A$ = 2, $h_{dim}$ = 600) & 2.3 & 1.6  \\
\textbf{RIMs} ($k_T$ = 4, $k_A$ = 3, $h_{dim}$ = 600) & 1.7 & 4.3  \\

\textbf{RIMs} ($k_T$ = 5, $k_A$ = 2, $h_{dim}$ = 600) & 2.5 & 4,7  \\ 
\textbf{RIMs} ($k_T$ = 5, $k_A$ = 3, $h_{dim}$ = 600) & 0.4 & 4.0\\
\textbf{RIMs} ($k_T$ = 5, $k_A$ = 4, $h_{dim}$ = 600) & 0.2 & 0.7 \\

\textbf{RIMs} ($k_T$ = 6, $k_A$ = 2, $h_{dim}$ = 600) & 3.3 & 2.4\\ 
\textbf{RIMs} ($k_T$ = 6, $k_A$ = 3, $h_{dim}$ = 600) & 1.2 & 1.0  \\ 
\textbf{RIMs} ($k_T$ = 6, $k_A$ = 4, $h_{dim}$ = 600) & 0.7 & 5.0\\
\textbf{RIMs} 
($k_T$ = 6, $k_A$ = 5, $h_{dim}$ = 600) & 0.22 & 0.56 \\
\midrule
With No input Attention and Full Communication & & \\
\midrule
\textbf{RIMs} ($k_T$ = 6, $k_A$ = 6, $h_{dim}$ = 600) & 0.0 & 0.7  \\
\textbf{RIMs} ($k_T$ = 5, $k_A$ = 5, $h_{dim}$ = 500) & 0.0 & 1.7  \\
\textbf{RIMs} ($k_T$ = 2, $k_A$ = 2, $h_{dim}$ = 256) & 0.0 & 2.9  \\
\textbf{RIMs} ($k_T$ = 2, $k_A$ = 2, $h_{dim}$ = 512) & 0.0 & 1.8  \\
\textbf{RIMs} ($k_T$ = 1, $k_A$ = 1, $h_{dim}$ = 512) & 0.0 & 0.2  \\
\midrule
With input Attention and Full Communication & & \\
\midrule
\textbf{RIMs} ($k_T$ = 9, $k_A$ = 2, $h_{dim}$ = 900) & 0.24 & 0.15  \\
\textbf{RIMs} ($k_T$ = 9, $k_A$ = 3, $h_{dim}$ = 900) & 0.01 & 0.00  \\
\textbf{RIMs} ($k_T$ = 9, $k_A$ = 4, $h_{dim}$ = 900) & 0.01 & 0.00  \\
\textbf{RIMs} ($k_T$ = 9, $k_A$ = 5, $h_{dim}$ = 900) & 0.01 & 0.00  \\
\textbf{RIMs} ($k_T$ = 9, $k_A$ = 6, $h_{dim}$ = 900) & 0.01 & 0.01  \\
\textbf{RIMs} ($k_T$ = 9, $k_A$ = 7, $h_{dim}$ = 900) & 0.04 & 0.07 \\
\textbf{RIMs} ($k_T$ = 9, $k_A$ = 8, $h_{dim}$ = 900) & 0.00 & 0.10 \\
\textbf{RIMs} ($k_T$ = 9, $k_A$ = 9, $h_{dim}$ = 900) & 0.03 & 0.24 \\
\textbf{RIMs} ($k_T$ = 16, $k_A$ = 4, $h_{dim}$ = 1600) & 0.05 & 0.02 \\
\textbf{RIMs} ($k_T$ = 16, $k_A$ = 6, $h_{dim}$ = 1600) & 0.00 & 0.00 \\
\textbf{RIMs} ($k_T$ = 16, $k_A$ = 8, $h_{dim}$ = 1600) & 0.00 & 0.00 \\
\textbf{RIMs} ($k_T$ = 16, $k_A$ = 10, $h_{dim}$ = 1600) & 0.00 & 0.00 \\
\textbf{RIMs} ($k_T$ = 16, $k_A$ = 12, $h_{dim}$ = 1600) & 0.00 & 0.00 \\
\textbf{RIMs} ($k_T$ = 16, $k_A$ = 14, $h_{dim}$ = 1600) & 0.01 & 0.26 \\
\textbf{RIMs} ($k_T$ = 24, $k_A$ = 6, $h_{dim}$ = 2400) & 0.01 & 0.00 \\
\textbf{RIMs} ($k_T$ = 24, $k_A$ = 8, $h_{dim}$ = 2400) & 0.00 & 0.00 \\
\textbf{RIMs} ($k_T$ = 24, $k_A$ = 16, $h_{dim}$ = 2400) & 0.00 & 0.00 \\
\textbf{RIMs} ($k_T$ = 24, $k_A$ = 20, $h_{dim}$ = 2400) & 0.00 & 0.00 \\
\textbf{RIMs} ($k_T$ = 24, $k_A$ = 22, $h_{dim}$ = 2400) & 0.00 & 0.42\\
\bottomrule
\end{tabular}
}
}
\end{table*}

\newpage
\subsection{Effect of Varying Number and Active RIMs on Adding Task}
\label{table:adding_ablation}

In the adding task we consider a stream of numbers as inputs (given as real-values) and then indicate which two numbers should be added together as a set of two input streams which varies randomly between examples.  The length of the input sequence during testing is longer than during training.  This is a simple test of the model's ability to ignore the numbers which it is not tasked with adding together.  We provide the results in Table \ref{tb:adding_ablation} which demonstrates that proposed model generalize better for longer testing sequences as well as adding multiple numbers. We ran the proposed model with different configurations like changing number of RIMs as well as number of active RIMs.

\begin{table*}[htb!]
\centering
{\renewcommand{\arraystretch}{1.0}
\caption{\textbf{Error CE on the adding task}.   }
\label{tb:adding_ablation}
\centering
\scalebox{0.9}{
\begin{tabular}{lrrr} 
\toprule
Approach & Train Length 50 & Test Length 200  \\ 
\midrule
With input Attention and Full Communication & & \\
\midrule
\textbf{RIMs} ($k_T$ = 6, $k_A$ = 3, $h_{dim}$ = 300) & 0.0 & 0.0\\
\textbf{RIMs} ($k_T$ = 6, $k_A$ = 4, $h_{dim}$ = 300) & 0.0 & 0.0\\
\textbf{RIMs} ($k_T$ = 6, $k_A$ = 5, $h_{dim}$ = 300) & 0.0 & 0.0\\
\textbf{RIMs} ($k_T$ = 6, $k_A$ = 3, $h_{dim}$ = 600) & 0.0 & 0.0\\
\textbf{RIMs} ($k_T$ = 6, $k_A$ = 4, $h_{dim}$ = 600) & 0.0 & 0.0\\
\textbf{RIMs} ($k_T$ = 6, $k_A$ = 5, $h_{dim}$ = 600) & 0.0 & 0.0\\
\textbf{RIMs} ($k_T$ = 9, $k_A$ = 3, $h_{dim}$ = 900) & 0.0 & 0.0\\
\textbf{RIMs} ($k_T$ = 9, $k_A$ = 4, $h_{dim}$ = 900) & 0.0 & 0.0\\
\textbf{RIMs} ($k_T$ = 9, $k_A$ = 5, $h_{dim}$ = 900) & 0.0 & 0.0\\
\textbf{RIMs} ($k_T$ = 9, $k_A$ = 6, $h_{dim}$ = 900) & 0.0 & 0.0\\
\textbf{RIMs} ($k_T$ = 9, $k_A$ = 7, $h_{dim}$ = 900) & 0.0 & 0.0\\
\textbf{RIMs} ($k_T$ = 9, $k_A$ = 8, $h_{dim}$ = 900) & 0.0 & 0.0\\
\toprule
Approach & Train Length 500 & Test Length 1000  \\ 
\midrule
With input Attention and Full Communication & & \\
\midrule
\textbf{RIMs} ($k_T$ = 6, $k_A$ = 3, $h_{dim}$ = 600) & 0.0 & 0.0\\
\textbf{RIMs} ($k_T$ = 6, $k_A$ = 4, $h_{dim}$ = 600) & 0.0 & 0.0\\
\textbf{RIMs} ($k_T$ = 6, $k_A$ = 5, $h_{dim}$ = 600) & 0.0 & 0.0\\
\textbf{RIMs} ($k_T$ = 9, $k_A$ = 3, $h_{dim}$ = 900) & 0.0 & 0.0\\
\textbf{RIMs} ($k_T$ = 9, $k_A$ = 4, $h_{dim}$ = 900) & 0.0 & 0.0\\
\textbf{RIMs} ($k_T$ = 9, $k_A$ = 5, $h_{dim}$ = 900) & 0.0 & 0.0\\
\textbf{RIMs} ($k_T$ = 9, $k_A$ = 6, $h_{dim}$ = 900) & 0.0 & 0.0\\
\textbf{RIMs} ($k_T$ = 9, $k_A$ = 7, $h_{dim}$ = 900) & 0.0 & 0.0\\
\bottomrule
\end{tabular}
}
}
\end{table*}

\subsection{Sequential MNIST Resolution Task}
\label{sec:mnist_ablation}

In this task we considered classifying binary MNIST digits by feeding the pixels to an RNN (in a fixed order scanning over the image).  As the focus of this work is on generalization, we introduced a variant on this task where the training digits are at a resolution of 14 x 14 (sequence length of 196).  We then evaluated on MNIST digits of different higher resolutions (16 x 16, 19 x 19, and 24 x 24).  When re-scaling the images, we used the nearest-neighbor based down-scaling and performed binarization after re-scaling.  We trained with a learning rate of 0.0001 and the Adam optimizer.  For RIMs we used a total of 600 hidden units split across 6 RIMs (100 units per RIM).  For the LSTM we used a total of 600 units.  We ran proposed model as well as baselines for 100 epochs. For  sequential MNIST task, we used 1 head in input attention, and 4 heads for RIMs communication.

\subsection{Bouncing Ball Environment}
We use the bouncing-ball dataset from \citep{van2018relational}. The dataset consists of 50,000 training examples and 10,000 test examples showing $\sim$50 frames of either 4 solid balls bouncing in a confined square geometry, 6-8 balls bouncing in a confined geometry, or 3 balls bouncing in a confined geometry with a random occluded region.
In all cases, the balls bounce off the wall as well as off one another.  We train baselines as well as proposed model for about 100 epochs using 0.0007 as learning rate and using Adam as optimizer \citep{Kingma2014}. We use the same architecture for encoder as well as decoder as in \citep{van2018relational}. We train the proposed model as well as the baselines for 100 epochs. Our goal in this section is to give more thorough experimental results omitted from the main paper for the sake of brevity.  Below, we highlight a few different results.

\subsubsection{Different RIMs attend to different balls}
\label{appendix:diffballs}

\begin{figure}[htb!]
    \centering
    \includegraphics[width=0.95\linewidth]{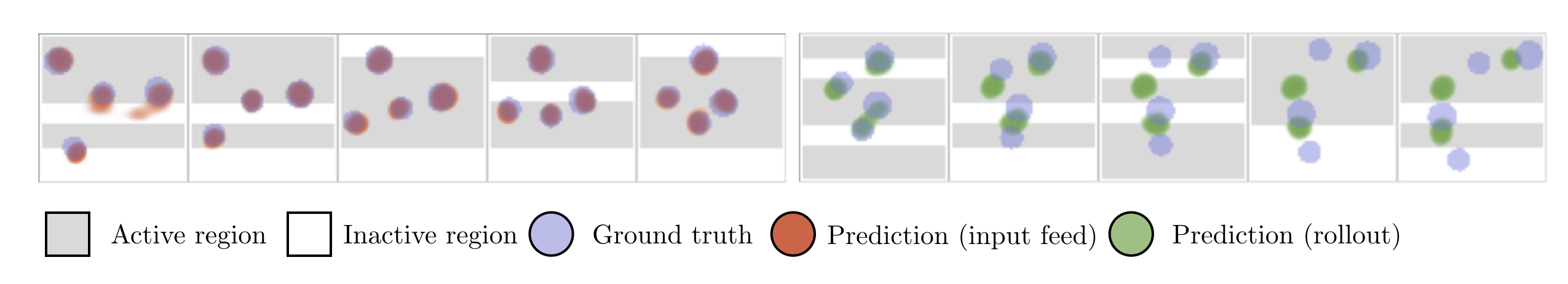}
    \caption{\textbf{Different RIMs attending to Different Balls}. For understanding what each RIM is actually doing, we associate each with a separate encoder, which are spatially masked.  Only 4 encoders can be active at any particular instant and there are four different balls. We did this to check if there would be the expected geometric activation of RIMs. 1.) Early in training, RIM activations correlated more strongly with the locations of the four different balls. Later in training, this correlation decreased and the active strips did not correlate as strongly with the location of balls. As the model got better at predicting the location, it needed to attend less to the actual objects. The top row shows every 5th frame when the truth is fed in and the bottom shows the results during rollout. The gray region shows the active block. In the top row, the orange corresponds to the prediction and in the bottom, green corresponds to the prediction.}
    \label{fig:geometric}
\end{figure}
In order to visualize what each RIM is doing, we associate each RIM with a different encoder. 
By performing spatial masking on the input, we can control the possible spatial input to each RIM.  
We use six non-overlapping horizontal strips and allow only 4 RIMs to be active at a time (shown in Fig.~\ref{fig:geometric}). The mask is fixed mask of zeros with a band of ones that is multiplied by the input to each encoder. Therefore, each of the 6 encoders gets 1/6th of the input. The goal was to see how the RIM activation patterns changed/correlated with the locations of the balls.
We find that early in training, the RIMs' activations are strongly correlated with the location of the 4 balls. However, after training has proceeded for some time this correlation deteriorates. 
This is likely because the predictable dynamics of the system do not necessitate constant attention. 

\subsubsection{Comparison with LSTM Baselines}
\begin{figure}[htb!]
    \centering
    \includegraphics[width=0.95\linewidth]{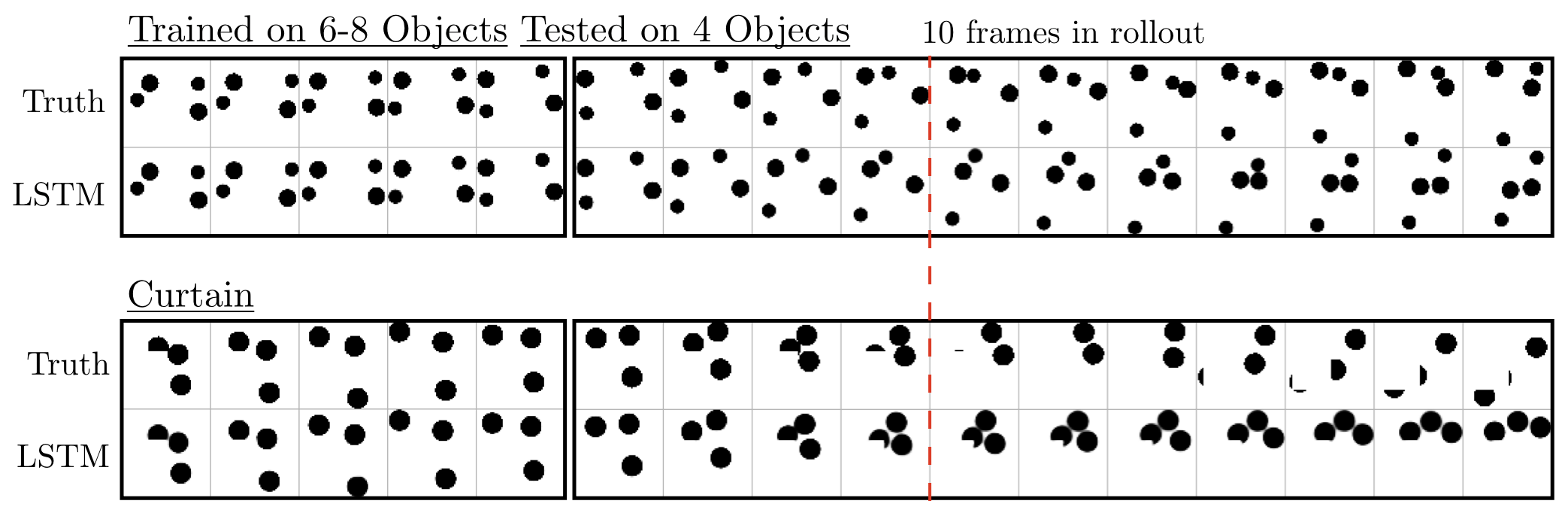}
    \caption{\textbf{Example of the other LSTM baselines}. For the 2 other experiments that we consider, here we show example outputs of our LSTM baselines. In each row, the top panel represents the ground truth and the bottom represents the prediction. All shown examples use an LSTM with 250 hidden units, as shown in Fig.~\ref{fig:err}. Frames are plotted every 3rd time step. The red line marks 10 rollout frames. This is marked because after this we  do not find BCE to be a reliable measure of dissimilarity.}
    \label{fig:lstm2}
\end{figure}

\begin{figure}[htb!]
    \centering
    \includegraphics[width=0.95\linewidth]{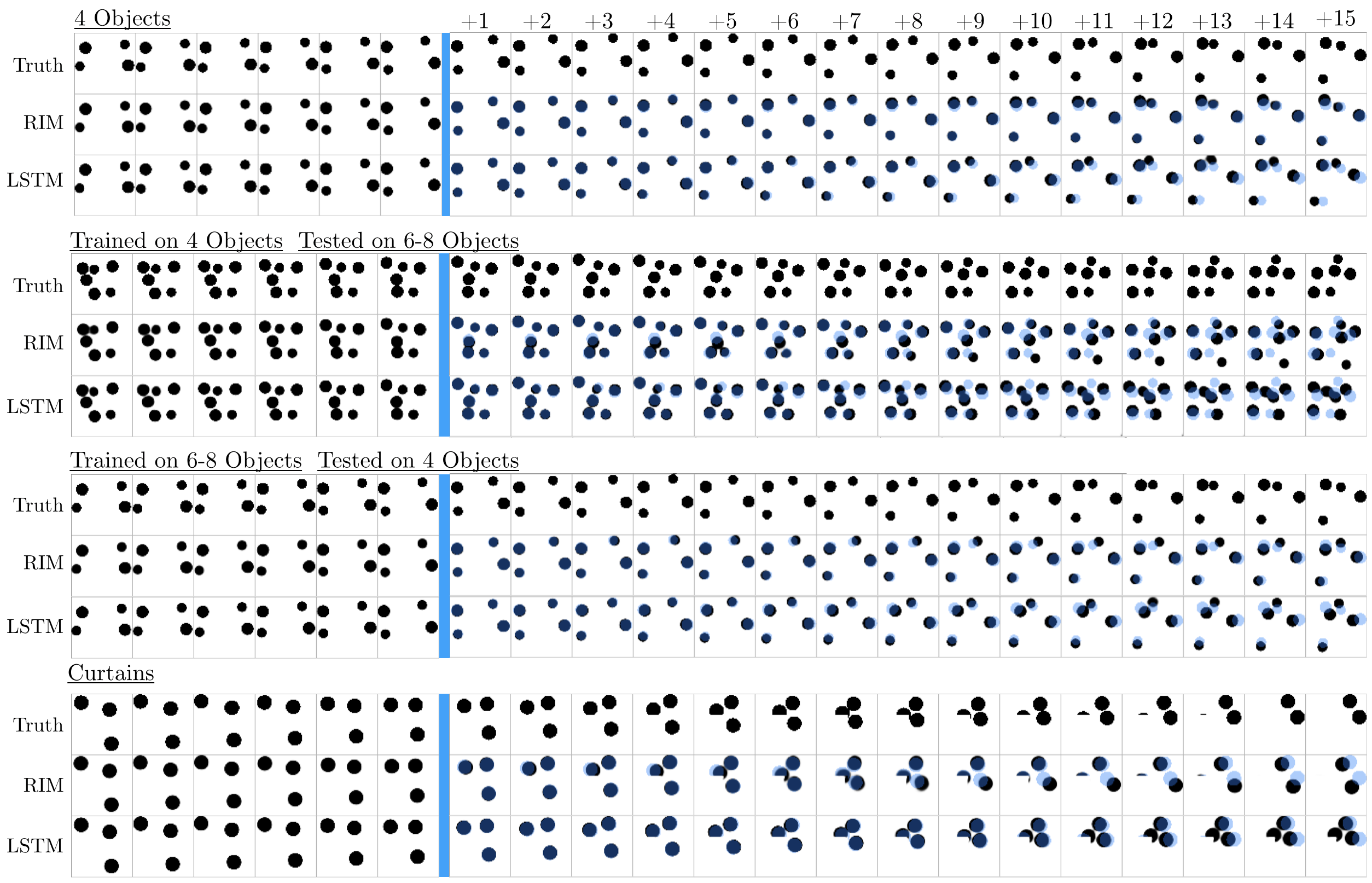}
    \caption{\textbf{Comparison of RIMs to LSTM baseline}. For 4 different experiments in the text, we compare RIMs to two different LSTM baselines. In all cases we find that during rollout, RIMs perform better than the LSTMs at accurately capturing the trajectories of the balls through time. Due to the number of hard collisions, accurate modeling is very difficult. In all cases, the first 15 frames of ground truth are fed in (last 6 shown) and then the system is rolled out for the next 15 time steps, computing the binary cross entropy between the prediction and the true balls at each instant, as in \cite{van2018relational}.  In the predictions, the transparent blue shows the ground  truth, overlaid to help guide the eye.}
    \label{fig:comparison}
    \vspace{-1mm}
\end{figure}

\begin{figure}[htb!]
    \centering
    \includegraphics[width=0.95\linewidth]{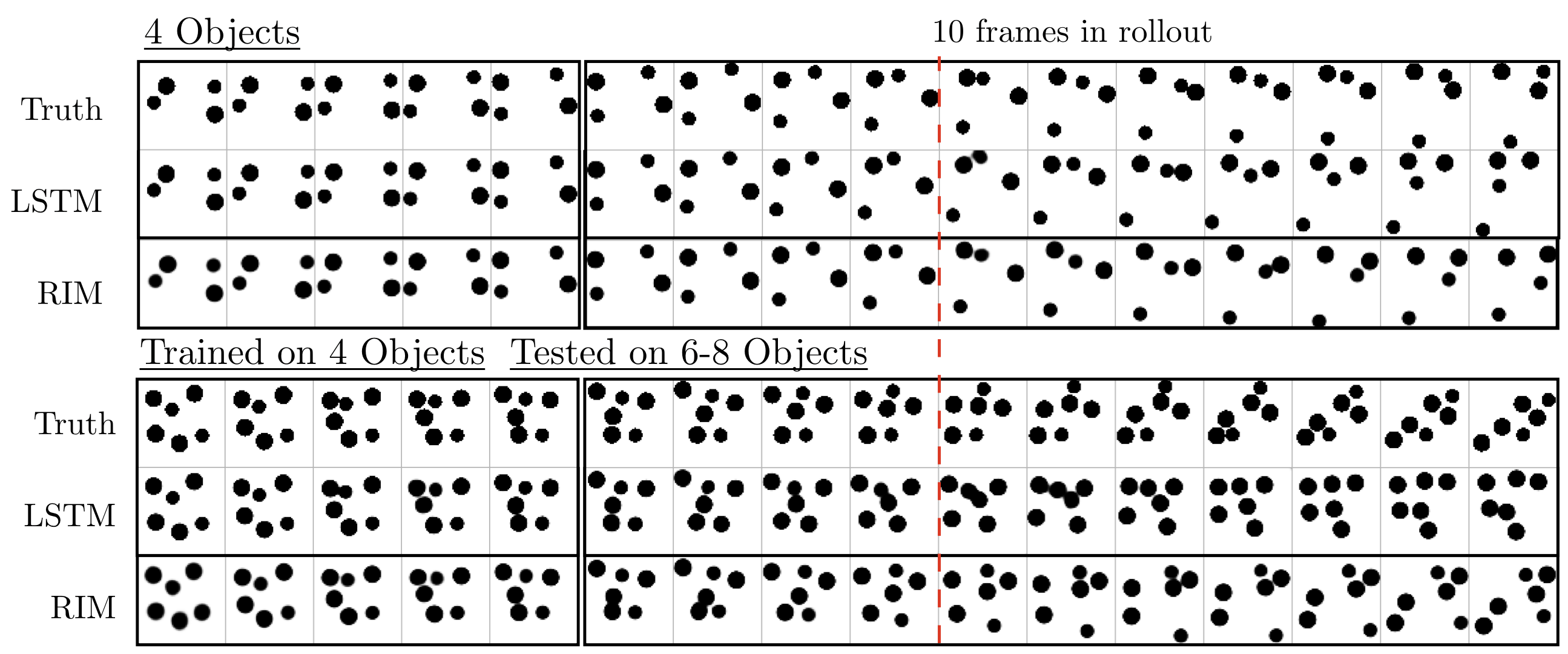}
    \caption{\textbf{Comparison between RIMs and LSTM baseline}. For the 4 ball task and the 6-8 ball extrapolation task, here we show an example output of from our LSTM baseline and from RIMs. All shown examples use an LSTM with 250 hidden units, as shown in Fig.~\ref{fig:err}. Frames are plotted every 3rd time step. The red line marks 10 rollout frames. This is marked because after this we  do not find BCE to be a reliable measure of dissimilarity.}
    \label{fig:lstm}
\end{figure}

\begin{figure}[htb!]
    \centering
    \includegraphics[width=0.95\linewidth]{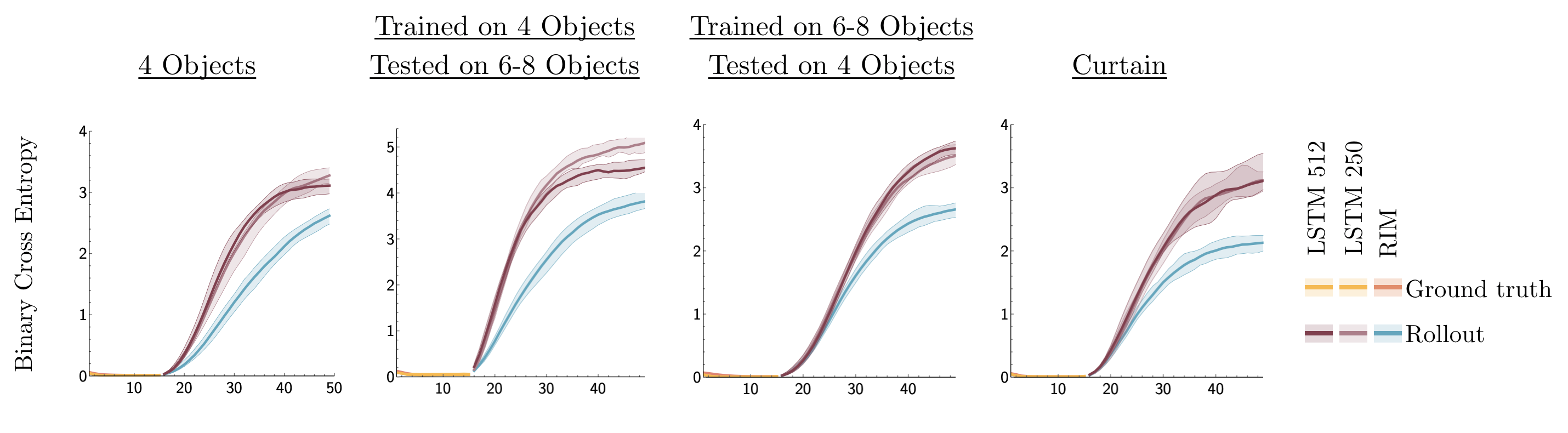}
    \caption{\textbf{Comparison of RIMs to LSTM baseline}. For 4 different experiments in the text, we compare RIMs to two different LSTM baselines. In all cases we find that during rollout, RIMs perform better than the LSTMs at accurately capturing the trajectories of the balls through time. Due to the number of hard collisions, accurate modeling is very difficult.}
    \label{fig:err2}
\end{figure}

In Figures \ref{fig:lstm2}, \ref{fig:comparison}, \ref{fig:lstm}, and \ref{fig:err2} we highlight different baselines and how these compare to the proposed RIMs model.
\vspace{-1mm}


\subsubsection{Occlusion}
\begin{figure}[htb!]
    \centering
    \includegraphics[width=0.95\linewidth]{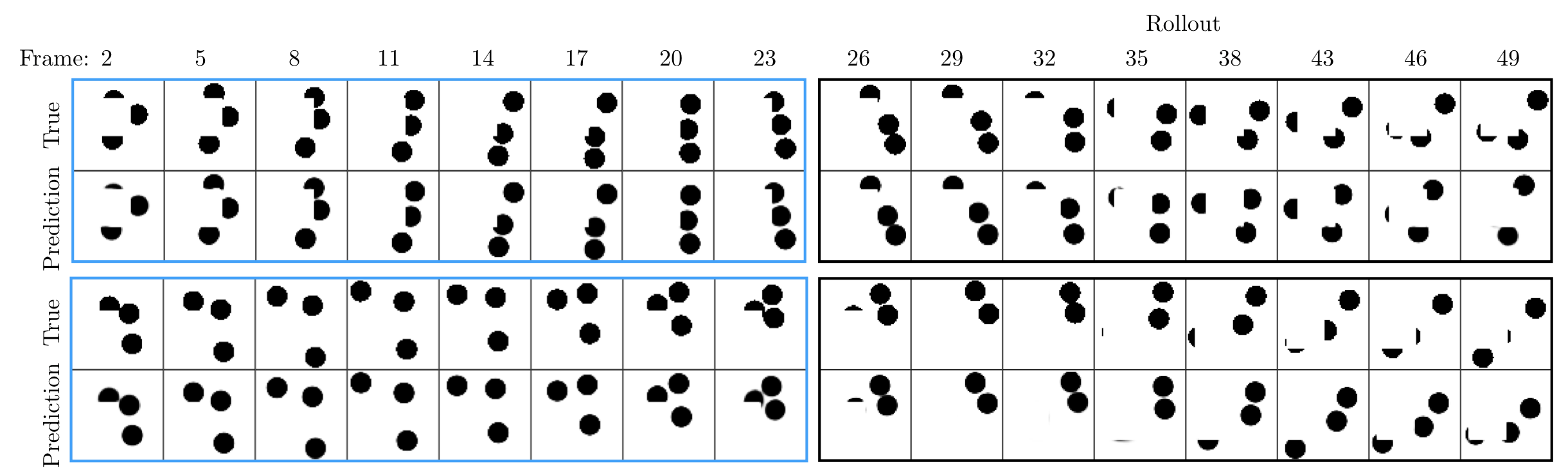}
    \caption{\textbf{RIMs on dataset with an occlusion}. We show two trajectories (top and bottom) of three balls. For the left frames, at each step the true frame is used as input. On the right,  outlined in black, the previous output is used as input.}
    \label{fig:curtain}
\end{figure}

In Fig.~\ref{fig:curtain}, we show the performance of RIMs on the curtain dataset. We find RIMs are able to track balls through the occlusion without difficulty. Note that the LSTM baseline, is also able to track the ball through the ``invisible'' curtain.

\subsubsection{Study of Transfer}
\begin{figure}[htb!]
    \centering
    \includegraphics[width=0.95\linewidth]{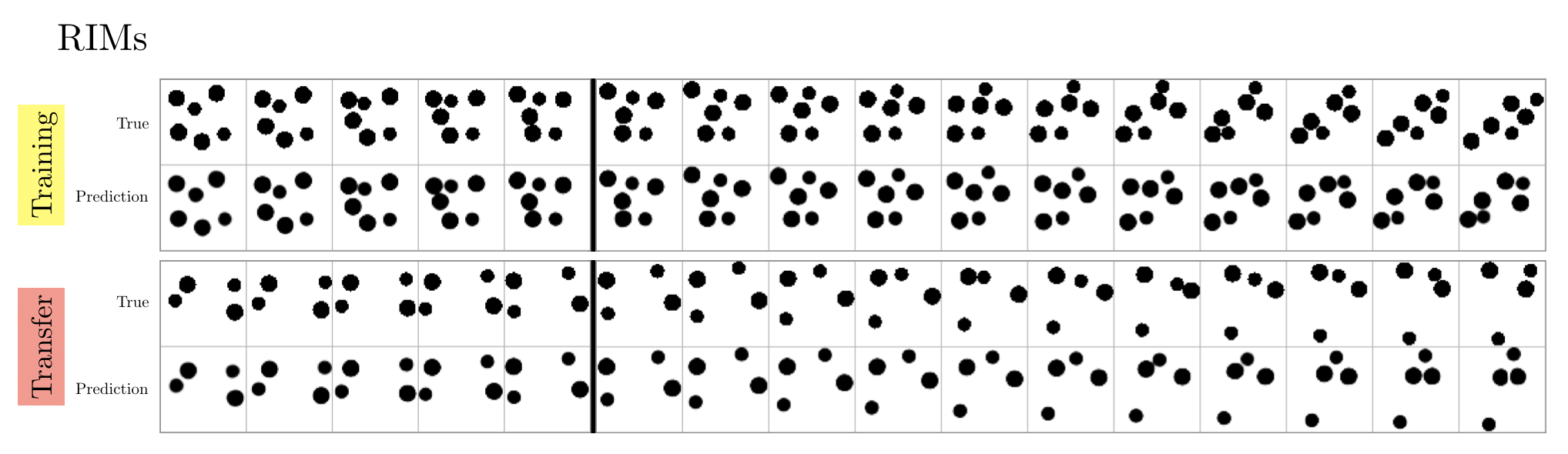}
    \caption{\textbf{RIMs transferred on new data}. We train the RIMs model on the 6-8 ball dataset (as shown in the top row). Then, we apply the model to the 4 ball dataset, as shown in the bottom. 
    }
    \label{fig:interpolation}
\end{figure}

It is interesting to ask how models trained on a dataset with 6-8 balls perform on a dataset with 4 balls. In Fig.~\ref{fig:interpolation} we show predictions during feed-in and rollout phases.

\subsection{Video Prediction from Crops} \label{sec:bballs_appendix}

\begin{figure} 
\centering
\includegraphics[width=0.9\textwidth]{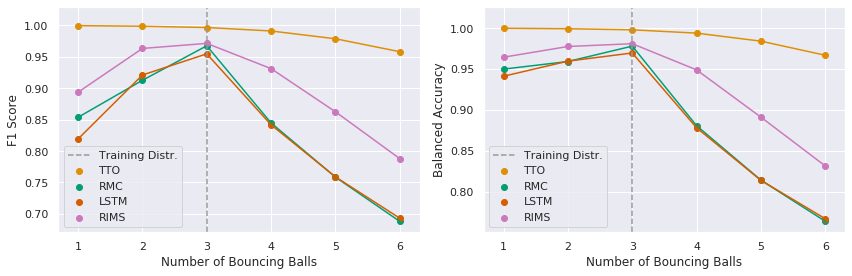}
\caption{Performance metrics on OOD one-step forward prediction task. \textbf{Gist:} RIMs outperforms all RNN baselines OOD.}
\label{fig:bb_ood}
\vspace{-10pt}
\end{figure}


\textbf{Dataset.} We train all models on a training dataset of $20$K video sequences with $100$ frames of $3$ balls bouncing in an arena of size $48 \times 48$. We also include an additional fixed ball in the center to make the task more challenging. We use another $1$K video sequences of the same length and the same number of balls as a held-out validation set. In addition, we also have $5$ out-of-distribution (OOD) test sets with various number of bouncing balls (ranging from $1$ to $6$) and each containing $1$K sequences of length $100$. 

\textbf{Training.} We train all models until the validation loss is saturated, and select the best of three runs. During training, we automatically decay the learning rate by a factor of $2$ if the validation loss does not significantly decrease by at least $0.01\%$ for five consecutive epochs.

\textbf{Evaluation Criteria.} After having trained on the training dataset with 3 bouncing balls, we evaluate the performance on all test datasets with $1$ to $6$ bouncing balls. In Figure~\ref{fig:bb_ood}, we report the balanced accuracy (i.e. arithmetic mean of recall and specificity) and F1-scores (i.e. harmonic mean of precision and recall) to account for class-imbalance 

\textbf{Results.} In Figure~\ref{fig:bb_ood}, we see that proposed method out-perform all non-oracle baselines OOD on the one-step forward prediction task and strike a good balance in regard to in-distribution and OOD performance. 

\textbf{Baselines:} We compared the performance of the proposed method with a baseline LSTM model, as well as state of the art memory model, RMC \citep{santoro2018relational}. As a sanity check we also compare the performance of the proposed method to an oracle baseline that does not model the dynamics (refer to as TTO: Time Travelling Oracle).

\subsection{Environment with Novel Distractors}

We evaluate the proposed framework using Adavantage Actor-Critic (A2C)  to learn a policy $\pi_\theta(a | s, g)$ conditioned on the goal. To evaluate the performance of proposed method, we use a range of maze multi-room tasks from the gym-minigrid framework \citep{gym_minigrid} and the A2C implementation from \citep{gym_minigrid}. For the maze tasks, we used agent's relative distance to the absolute goal position as "goal".

For the maze environments, we use A2C with 48 parallel workers. Our actor network and critic networks consist of two and three fully connected
layers respectively, each of which have 128 hidden units.  The encoder network is also parameterized as a neural network, which consists of 1 fully connected layer. We use RMSProp with an initial learning rate of $0.0007$ to train the models. Due to the partially observable nature of the environment, we further use a LSTM to encode the state and summarize the past observations. 

\subsection{MiniGrid Environments for OpenAI Gym\label{sec:minigrid}}

The  MultiRoom environments used for this research are part of MiniGrid, which is an open source gridworld package\footnote{https://github.com/maximecb/gym-minigrid}. This package includes a family of reinforcement learning environments compatible with the OpenAI Gym framework. Many of these environments are parameterizable so that the difficulty of tasks can be adjusted (e.g., the size of rooms is often adjustable).

\subsubsection{The World}

In MiniGrid, the world is a grid of size NxN. Each tile in the grid contains exactly zero or one object. The possible object types
are wall, door, key, ball, box and goal. Each object has an associated discrete color, which can be one of red, green, blue, purple, yellow and grey. By default, walls are always grey and goal squares are always green.

\begin{figure}
    \centering
    \includegraphics[width=0.6\linewidth]{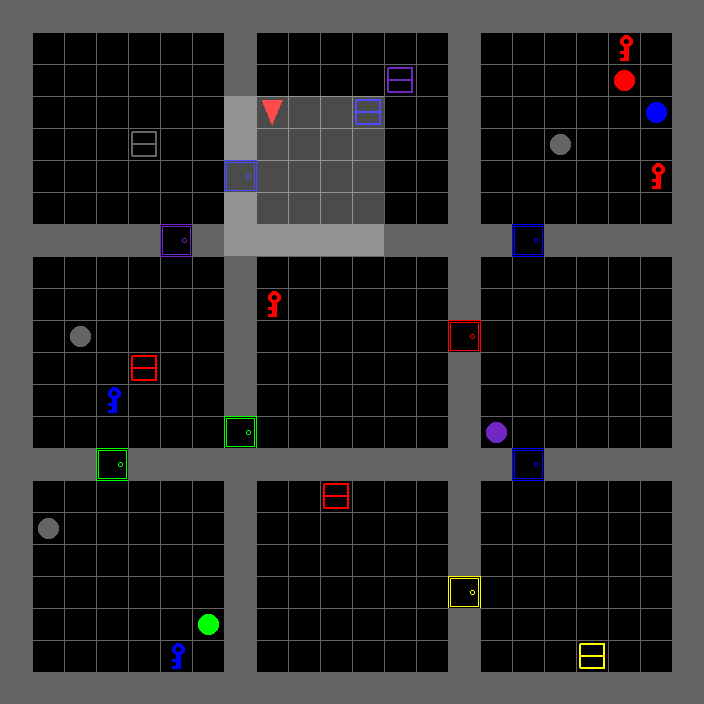}
    \caption{An example of the minigrid task.  }
    \label{fig:minigrid_pick}
\end{figure}

\subsubsection{Reward Function}

Rewards are sparse for all MiniGrid environments. In the MultiRoom environment, episodes are terminated with a positive reward when the agent reaches the green goal square. Otherwise, episodes are terminated with zero reward when a time step limit is reached. In the FindObj environment, the agent receives a positive reward if it reaches the object to be found, otherwise zero reward if the time step limit is reached.

The formula for calculating positive sparse rewards is $1 - 0.9 * (step\_count / max\_steps)$. That is, rewards are always between zero and one, and the quicker the agent can successfully complete an episode, the closer to $1$ the reward will be. The $max\_steps$ parameter is different for each environment, and varies depending on the size of each environment, with larger environments having a higher time step limit.

\subsubsection{Action Space}

There are seven actions in MiniGrid:  turn left, turn right, move forward, pick up an object, drop an object, toggle and done. For the purpose of this paper, the pick up, drop and done actions are irrelevant. The agent can use the turn left and turn right action to rotate and face one of 4 possible directions (north, south, east, west). The move forward action makes the agent move from its current tile onto
the tile in the direction it is currently facing, provided there is nothing on that tile, or that the tile contains an open door.
The agent can open doors if they are right in front of it by using the toggle action.

\subsubsection{Observation Space}

Observations in MiniGrid are partial and egocentric. By default, the agent sees a square of 7x7 tiles in the direction it is facing. These
include the tile the agent is standing on. The agent cannot see through walls or closed doors. The observations are provided as a tensor of shape 7x7x3. However, note that these are  RGB images (which is different from original BabyAI paper). 

\subsubsection{Level Generation}
\label{sec:level_generation}
The level generation in this task works as follows: (1) Generate the layout of the map (X number of rooms with different sizes (at most size Y) and green goal) (2) Add the agent to the map at a random location in the first room. (3) Add the goal at a random location in the last room.
A neural network parameterized as CNN is used to process the visual observation.

We follow the same architecture as 
\citep{gym_minigrid} but we replace the LSTM layer with BlockLSTM.

\subsubsection{Additional Ablation}
\label{sec:bouncing_balls_ablation_1head}
\begin{figure}[htb!]
    \centering
    \includegraphics[width=0.5\linewidth]{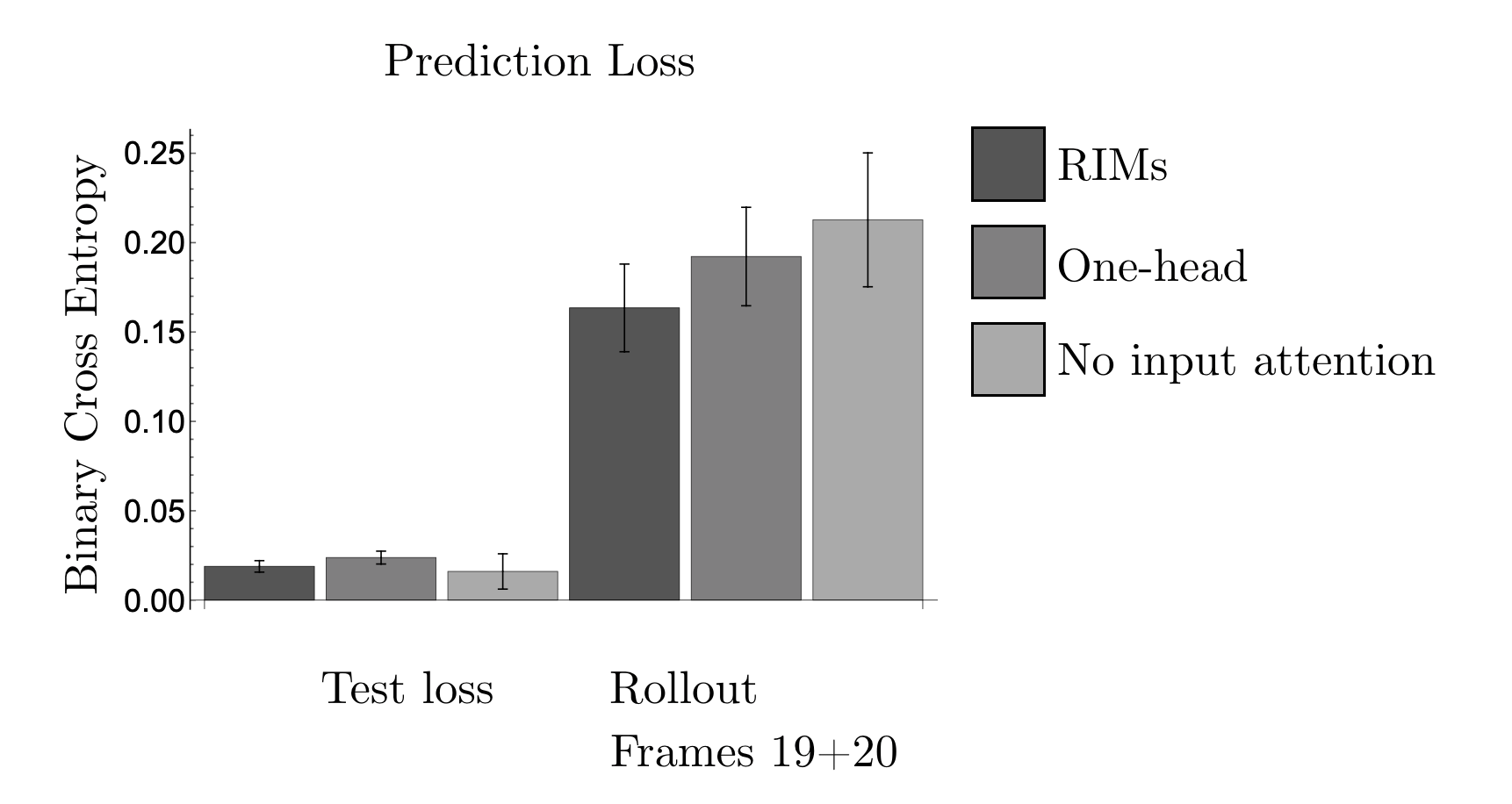}
    \caption{\textbf{Ablation loss} For the normal, a one-head model, and without input attention, we show the loss during training and the loss for the 4th and 5th frame of rollout. We find that the one-head and without input attention models perform worse than the normal RIMs model during the rollout phase.}
    \label{fig:ablate2}
\end{figure}

We present one ablation in addition to the ones in Section~\ref{sec:main_ablation}. In this experiment, we  study the effect on input attention (i.e top down attention) as well as the use of multi-headed head key-value attention. We compare the proposed model (with input attention as well as multi-headed key-value attention) with 2 baselines: (a) In which we remove the input attention (and force all the RIMs to communicate with each other (b) We use 1 head for key-value attention as compared to multi-headed key-value attention. Results comparing the proposed model, with these two baselines is shown in Fig.~\ref{fig:ablate2}. 



\begin{figure}[htb!]
    \centering
    \includegraphics[width=0.95\linewidth]{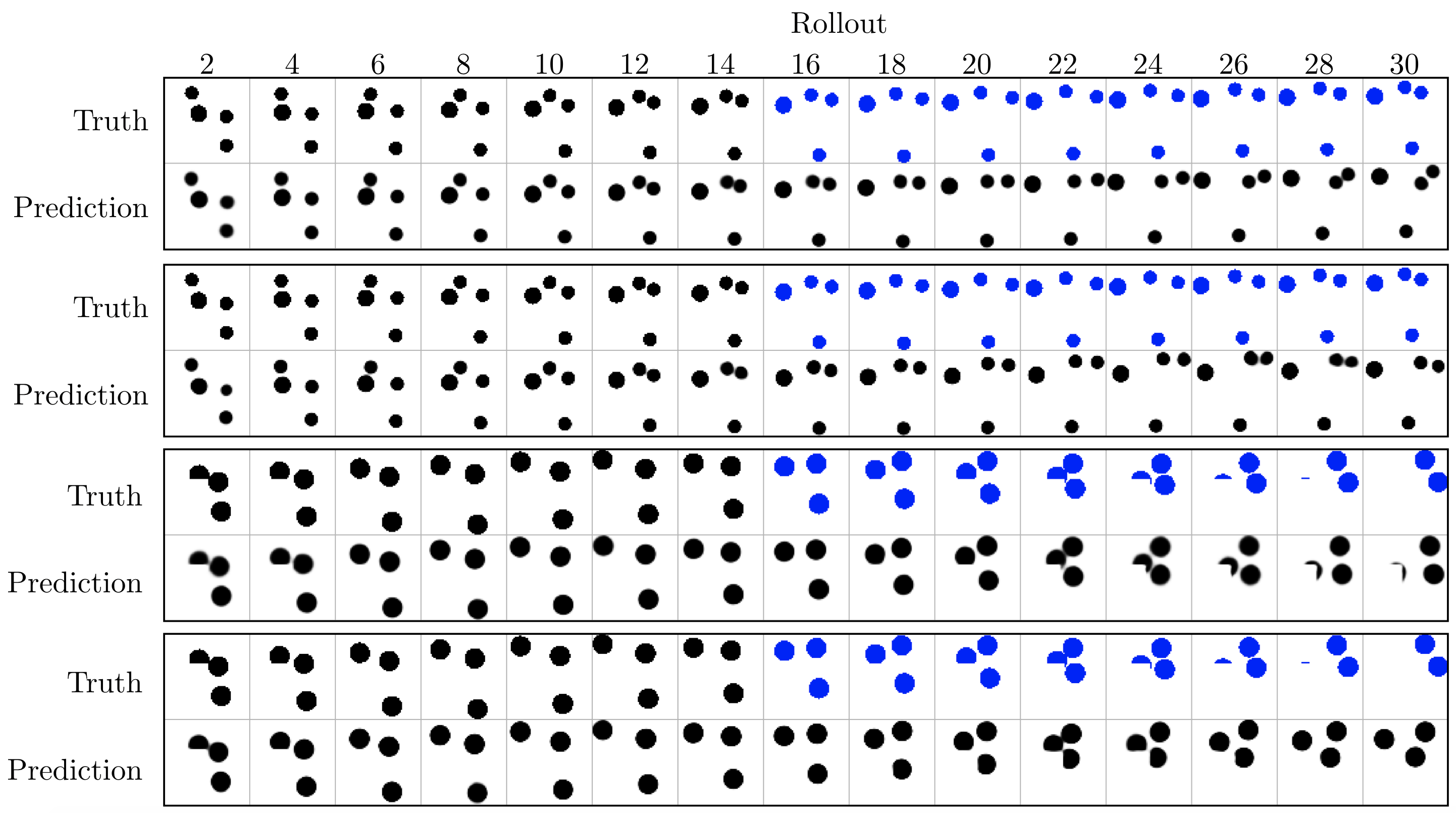}
    \caption{\textbf{One head and no attention} Using one head and no attention models, we show the rollout predictions in blue. On top we show results on the 4 ball dataset and on the bottom we show results on the curtains dataset.}
    \label{fig:onehead}
\end{figure}

In Fig.~\ref{fig:onehead}, we show the predictions that result from the model with only one active head.

\subsection{Atari}
\label{appendix:atariexp}

We used open-source implementation of PPO from \citep{pytorchrl} with default parameters. We ran the proposed algorihtm with 6 RIMs, and kept the number of activated RIMs to 4/5. We have not done any hyper-parameter search for Atari experiments.

\begin{figure}[H]
    \centering
    \includegraphics[width=0.8\linewidth]{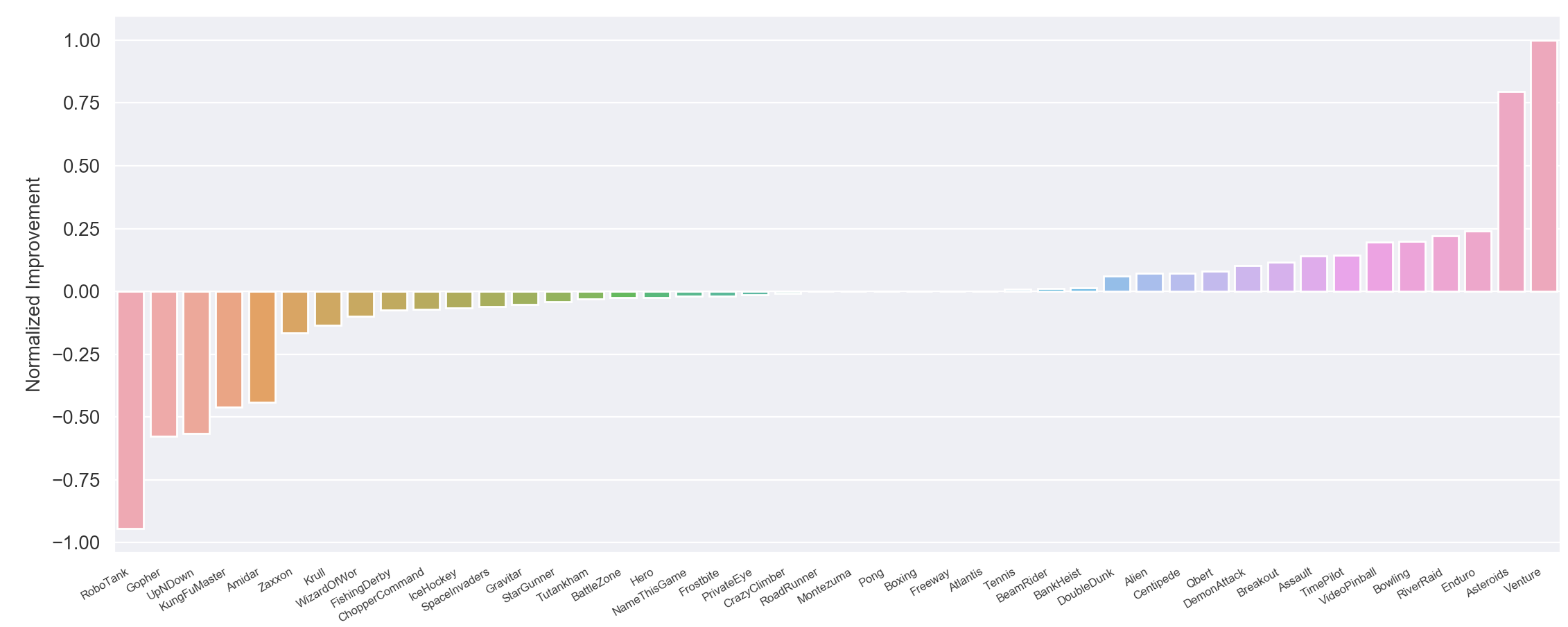}
    \caption{A comparison showing relative improvement of RIMs with $k_A=5$ over a $k_A=4$ baseline.  Using $k_A=5$ performs slightly worse than $k_A=4$ but still outperforms PPO, and has similar results across the majority of games.  }
    \label{fig:atari5vs4}
\end{figure}

\vspace{-1mm}
\begin{figure}[H]
    \centering
    \includegraphics[width=0.8\linewidth]{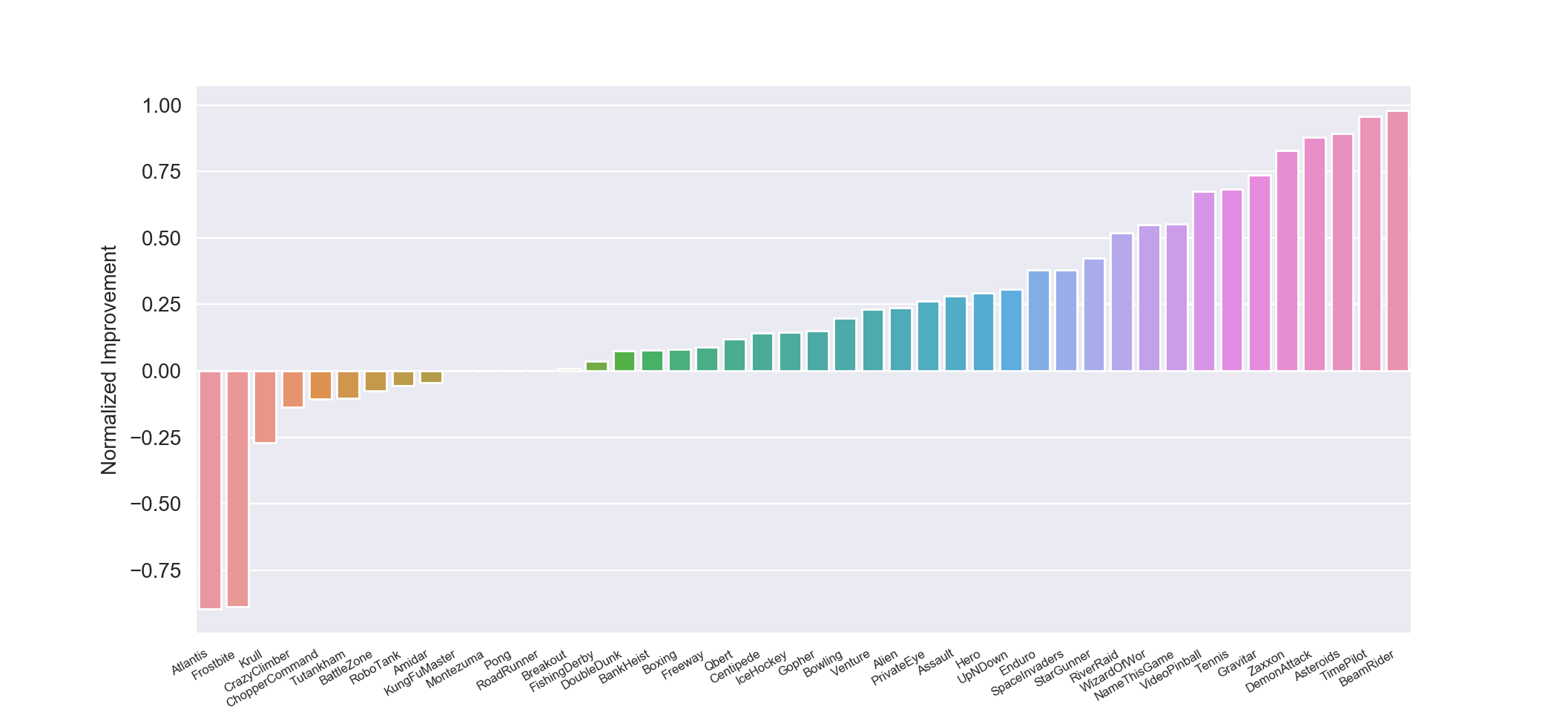}
    \caption{RIMs-PPO relative score improvement over LSTM-PPO baseline \citep{schulman2017proximal} across all Atari games averaged over 3 trials per game.  In both cases PPO was used with the exact same settings with the only change being the choice of the recurrent architecture (RIMs with $k_A=5$).}
    \label{fig:my_label}
\end{figure}

\section{Additional Experiments}

\subsection{Imitation Learning: Robustness to Noise in State Distribution}
\label{sec:imitationmujoco}

Here, we consider  imitation learning where we have training trajectories generated from an expert (Table~\ref{tb:rl}). We evaluate our model on continuous control tasks in Mujoco (in our case, Half-Cheetah) \citep{todorov2012mujoco}.  We take the rendered images as input and compared the proposed model with recurrent policy (i.e., LSTM). Since, using rendered image of the input does not tell anything about the velocity of the Half-Cheetah, it makes the task partially observable.  In order to test how well the proposed model generalizes during test, we add some noise (in the joints of the half-cheetah body). As one can see, after adding noise LSTM baselines performs poorly. On the other hand, for the proposed model, there's also a drop in performance but not as bad as for the LSTM baseline.

\begin{table*}[htb!]
\caption{\textbf{Imitation Learning:} Results on the half-cheetah imitation learning task.  RIMs outperforms a baseline LSTM when we evaluate with perturbations not observed during training (left).  An example of an input image fed to the model (right).  }
\label{tb:rl}
\begin{minipage}{.5\textwidth}
{\renewcommand{\arraystretch}{1.2}
\begin{tabular}{lrr}
\hline
\toprule
\centering
Method / Setting & \makecell{Training \\Observed Reward} & \makecell{Perturbed States\\ Observed Reward} \\
\midrule
LSTM (Recurrent Policy) & 5400 $\pm$ 100                    & 2500 $\pm$ 300                          \\
\textbf{RIMs} ($k_T$ = 6, $k_A$ = 3) & 5300 $\pm$ 200   & 3800 $\pm$ 200 \\
\textbf{RIMs} ($k_T$ = 6, $k_A$ = 6) & 5500 $\pm$ 100  & 2700 $\pm$ 400 \\
\textbf{RIMs} (without Input attention) & 5400 $\pm$ 100  & 3200 $\pm$ 50 \\
\bottomrule
\end{tabular}
}
\end{minipage}%
\begin{minipage}{.4\textwidth}
\centering
\includegraphics[width=0.3\linewidth,right]{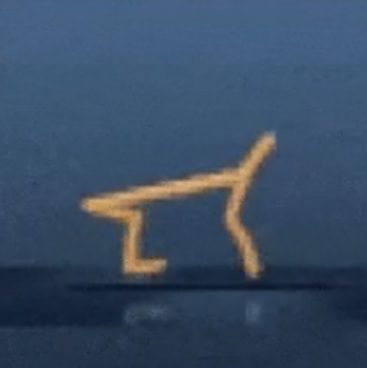}
\end{minipage}
\end{table*}

We use the convolutional network from \citep{ha2018world} as our encoder, a GRU \citep{chung2015recurrent} with 600
units as deterministic path in the dynamics model, and implement all other functions as two fully connected layers of size
256 with ReLU activations. Since, here we are using images as input, which makes the task, partially observable. Hence, we concatenate the past 4 observations, and then feed the concatenated observations  input to GRU (or our model). For our model, we use 6 RIMs, each of size 100, and we set $k_a = 3$. We follow the same setting as in \citep{hafner2018learning, sodhani2019learning}. We also compare the proposed method to the baseline where we dont include input attention (or top-down attention). AS \ref{tb:rl} shows, there's a decline in performance if we dont use input attention, hence justifying the importance




\begin{table*}[t]
\small

\centering
\begin{tabular}{ccc}
    \toprule
    Environment & LSTM-PPO & RIMs-PPO\\
    \midrule
    Alien & 1612  $\pm$ 44 & \textbf{2152} $\pm$ 81 \\
    Amidar & 1000  $\pm$ 58 & \textbf{1800} $\pm$ 43 \\
    Assault &  4000 $\pm$ 213 & \textbf{ 5400} $\pm$ 312 \\
    Asterix & 3090 $\pm$ 420 & \textbf{21040} $\pm$ 548 \\
    Asteroids & 1611.0 $\pm$ 200 & \textbf{ 3801 } $\pm$ 89 \\
    Atlantis & 3280000 $\pm$ 200000 & \textbf{3500000} $\pm$ 120000 \\
    BankHeist & 1153 $\pm$ 23 & \textbf{ 1195 } $\pm$ 4 \\
    BattleZone & 21000 $\pm$ 232.0 & \textbf{22000} $\pm$ 324 \\
    BeamRider & 698 $\pm$ 100 & \textbf{5320} $\pm$ 300 \\
    Bowling & 30 $\pm$ 5 & 42 $\pm$ 13 \\
    Boxing & 80  $\pm$ 3 & \textbf{95} $\pm$ 10 \\
    Breakout & 593 $\pm$ 90 & 590 $\pm$ 10 \\
    Centipede & 4600 $\pm$ 312 & \textbf{5534} $\pm$ 283 \\
    ChopperCommand & 11000 $\pm$ 790 & \textbf{12303} $\pm$ 412 \\
    CrazyClimber & \textbf{138000} $\pm$ 2412 & 132039 $\pm$ 1221 \\
    DemonAttack & 26320 $\pm$ 3234 & \textbf{230324} $\pm$ 4032\\
    DoubleDunk & \textbf{-3.0} $\pm$ 0.5 & -3.8  $\pm$ 0.3 \\
    Enduro & 1600 $\pm$ 200 & \textbf{2800} $\pm$ 232 \\
    FishingDerby & 20 $\pm$ 4 & \textbf{38} $\pm$ 8 \\
    Freeway & 29 $\pm$ 2 & \textbf{ 33 } $\pm$ 2 \\
    Gopher & 7000.0 $\pm$ 402 & \textbf{33000} $\pm$ 2210 \\
    Gravitar & 500 $\pm$ 100 & \textbf{ 1090 } $\pm$ 80 \\
    IceHockey & -5 $\pm$ 0.3 & -4 $\pm$ 1 \\
    Jamesbond & 425 $\pm$ 25 & \textbf{800} $\pm$ 100 \\
    Kangaroo & \textbf{ 13000 } $\pm$ 500 & 1800 $\pm$ 400 \\
    Krull & \textbf{10000} $\pm$ 500 & 7900 $\pm$ 200 \\
    KungFuMaster & 28000 $\pm$ 2000 & \textbf{51000} $\pm$ 800 \\
    NameThisGame & 4200 $\pm$ 400 & \textbf{6800} $\pm$ 300 \\
    Pong & 20 $\pm$ 1 & 20 $\pm$ 1 \\
    PrivateEye & 90 $\pm$ 3 & \textbf{ 100} $\pm$ 0 \\
    Qbert & 22000 $\pm$ 300 & \textbf{22500} $\pm$ 400 \\
    Riverraid & 7500 $\pm$ 300 & \textbf{12000} $\pm$ 100\\
    RoadRunner & 53000 $\pm$ 120 & \textbf{53430} $\pm$ 300 \\
    Robotank & 3 $\pm$ 1 & \textbf{ 11 } $\pm$ 2 \\
    SpaceInvaders & 1600 $\pm$ 40 & \textbf{ 2800} $\pm$ 80 \\
    StarGunner & 35000 $\pm$ 800 & \textbf{ 70000} $\pm$ 1200 \\
    TimePilot & 4000 $\pm$ 100 & \textbf{ 10000} $\pm$ 689 \\
    UpNDown & 70000 $\pm$ 6000 & \textbf{ 390000} $\pm$ 20000 \\
    VideoPinball & 90000 $\pm$ 5000 & \textbf{220000} $\pm$ 9000 \\
    WizardOfWor & 3833 $\pm$ 400 & \textbf{ 10800} $\pm$ 700 \\
    Zaxxon & 200 $\pm$ 100 & \textbf{ 15000} $\pm$ 600 \\
    \bottomrule
\end{tabular}
\caption{\small Scores obtained using PPO with the LSTM architecture and PPO with the RIMs architecture with $k_A = 5$.  }
\label{tbl:atari_large_quantative}
\end{table*}

\subsection{Transfer on Atari}

As a very preliminary result, we investigate feature transfer between randomly selected Atari games. In order to study this question, we follow the  experimental protocol of \cite{rusu2016progressive}.

We start by training RIMs on three source games (Pong, River Raid, and Seaquest) and test if the learned features transfer to a different subset of randomly selected target games (Alien, Asterix, Boxing, Centipede, Gopher, Hero, James Bond, Krull, Robotank, Road Runner, Star Gunner, and Wizard of Wor). We observe, that RIMs result in positive transfer in 9 out of 12 target games, with three cases of negative transfer. On the other hand progressive networks \citep{rusu2016progressive} result in positive transfer in 8 out of 12 target games, and two cases of negative transfer. We also compare to LSTM baseline, which yields positive transfer in  3 of 12 games.

\newpage 

\subsubsection{Atari Results: Comparison with LSTM-PPO}


\begin{figure*}[htb!]
\begin{tabular}{ccccc}
\subfloat{\includegraphics[width = 0.95in]{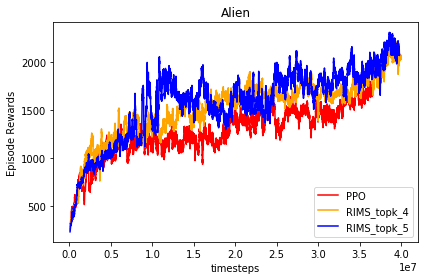}} &
\subfloat{\includegraphics[width = 0.95in]{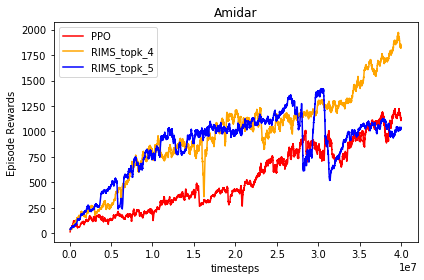}} &
\subfloat{\includegraphics[width = 0.95in]{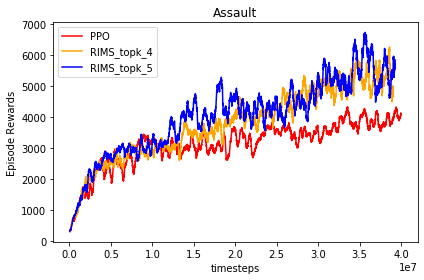}} &
\subfloat{\includegraphics[width = 0.95in]{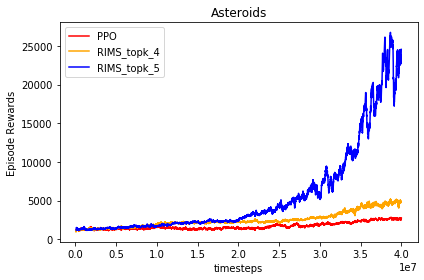}}&
\subfloat{\includegraphics[width = 0.95in]{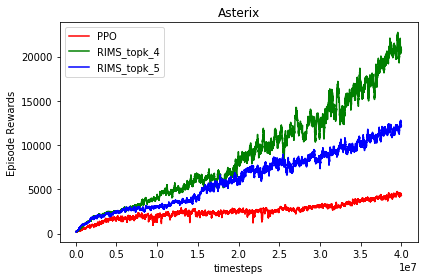}}\\
\subfloat{\includegraphics[width = 0.95in]{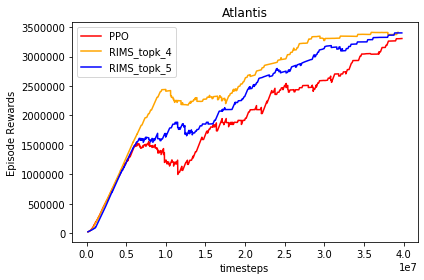}} &
\subfloat{\includegraphics[width = 0.95in]{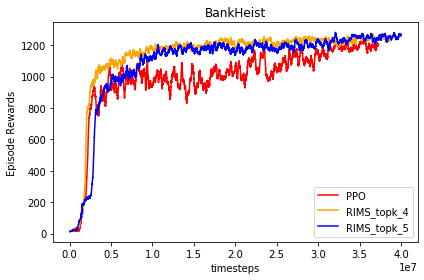}} &
\subfloat{\includegraphics[width = 0.95in]{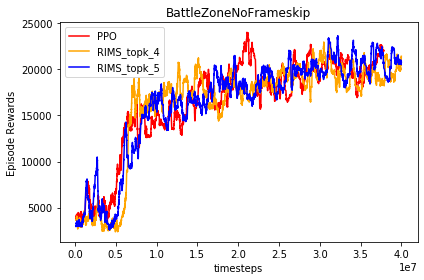}}&
\subfloat{\includegraphics[width = 0.95in]{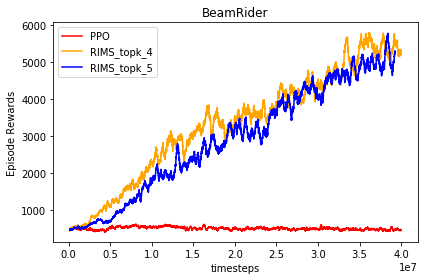}}&
\subfloat{\includegraphics[width = 0.95in]{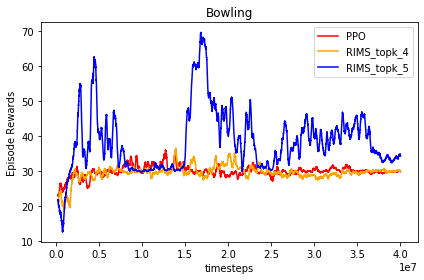}} \\
\subfloat{\includegraphics[width = 0.95in]{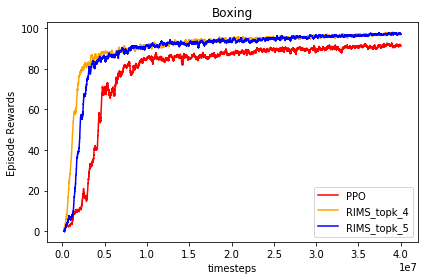}} &
\subfloat{\includegraphics[width = 0.95in]{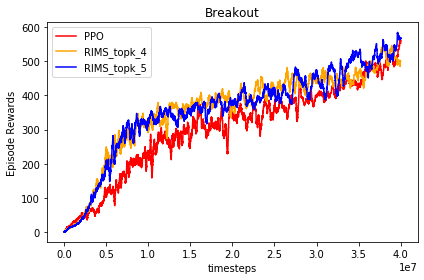}} &
\subfloat{\includegraphics[width = 0.95in]{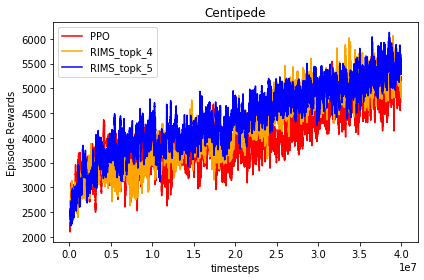}}&
\subfloat{\includegraphics[width = 0.95in]{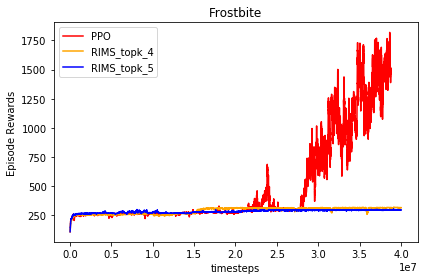}} &
\subfloat{\includegraphics[width = 0.95in]{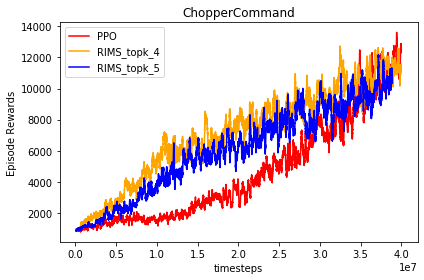}} \\
\subfloat{\includegraphics[width = 0.95in]{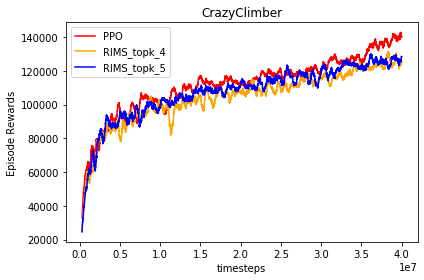}} &
\subfloat{\includegraphics[width = 0.95in]{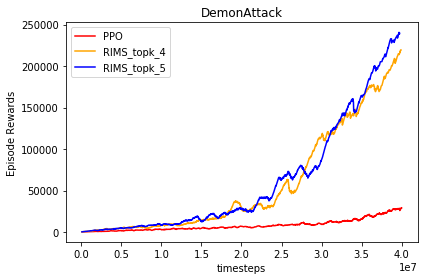}}&
\subfloat{\includegraphics[width = 0.95in]{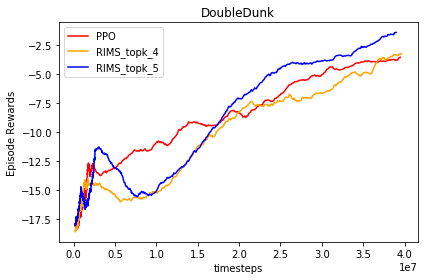}} &
\subfloat{\includegraphics[width = 0.95in]{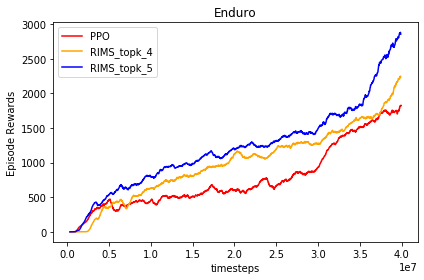}} &
\subfloat{\includegraphics[width = 0.95in]{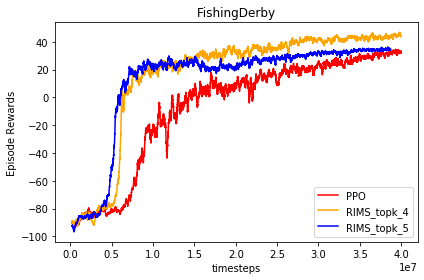}} \\
\subfloat{\includegraphics[width = 0.95in]{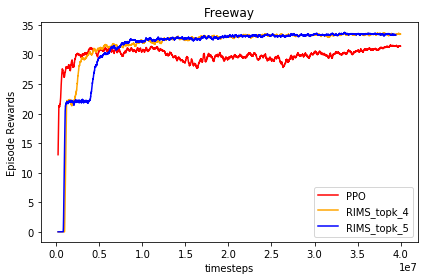}}&
\subfloat{\includegraphics[width = 0.95in]{atari_results/Frostbite.png}}&
\subfloat{\includegraphics[width = 0.95in]{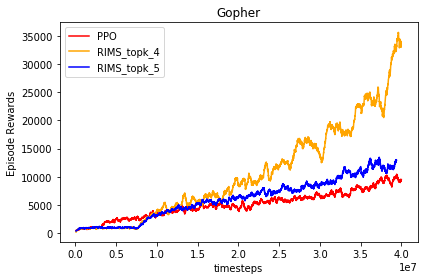}} &
\subfloat{\includegraphics[width = 0.95in]{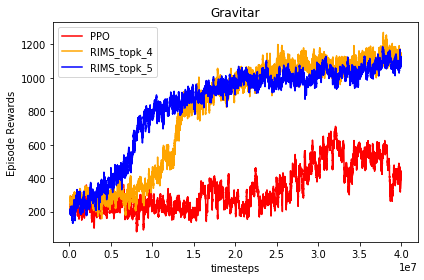}} &
\subfloat{\includegraphics[width = 0.95in]{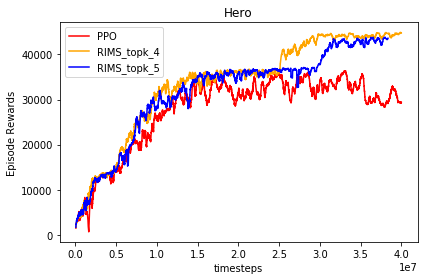}} \\
\subfloat{\includegraphics[width = 0.95in]{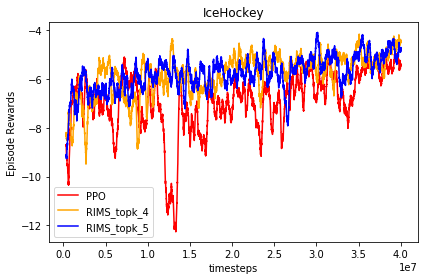}}&
\subfloat{\includegraphics[width = 0.95in]{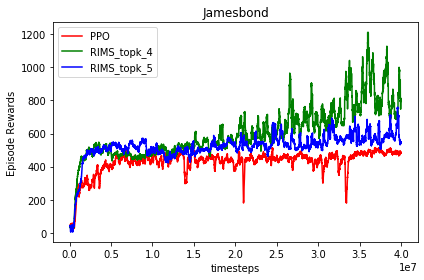}}&
\subfloat{\includegraphics[width = 0.95in]{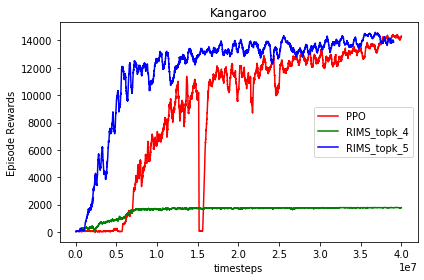}}&
\subfloat{\includegraphics[width = 0.95in]{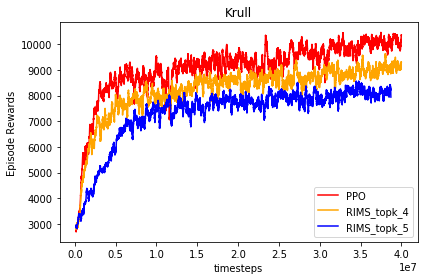}}&
\subfloat{\includegraphics[width = 0.95in]{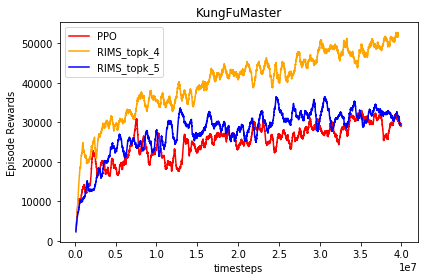}} \\
\subfloat{\includegraphics[width = 0.95in]{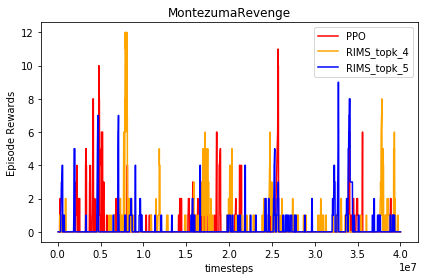}} &
\subfloat{\includegraphics[width = 0.95in]{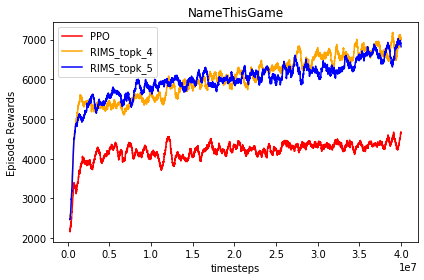}} &
\subfloat{\includegraphics[width = 0.95in]{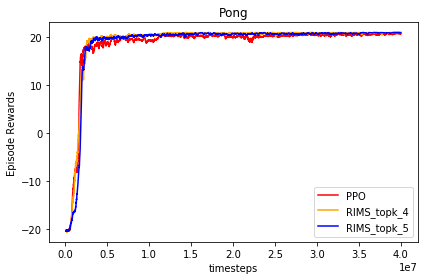}}&
\subfloat{\includegraphics[width = 0.95in]{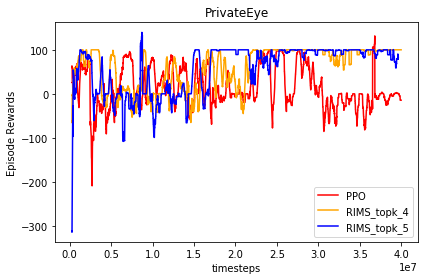}}&
\subfloat{\includegraphics[width = 0.95in]{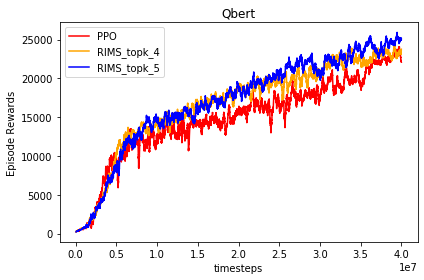}} \\
\subfloat{\includegraphics[width = 0.95in]{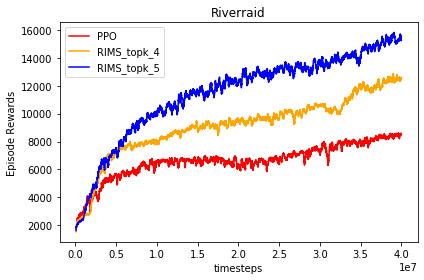}} &
\subfloat{\includegraphics[width = 0.95in]{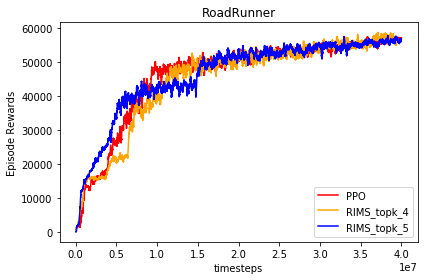}} &
\subfloat{\includegraphics[width = 0.95in]{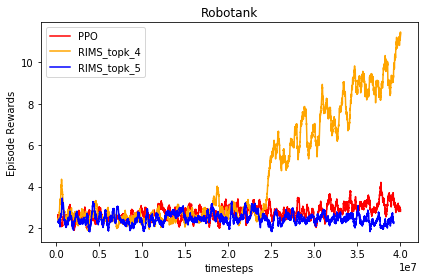}}&
\subfloat{\includegraphics[width = 0.95in]{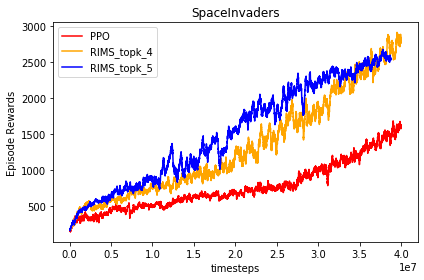}}&
\subfloat{\includegraphics[width = 0.95in]{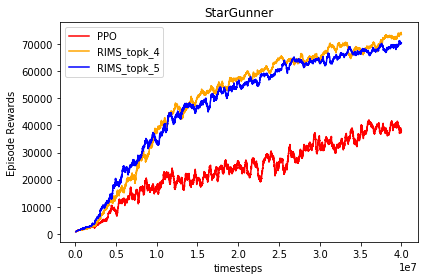}} \\
\subfloat{\includegraphics[width = 0.95in]{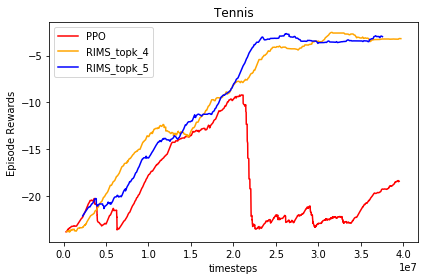}} &
\subfloat{\includegraphics[width = 0.95in]{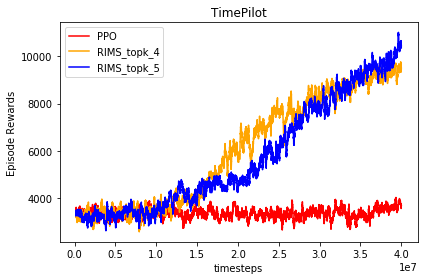}} &
\subfloat{\includegraphics[width = 0.95in]{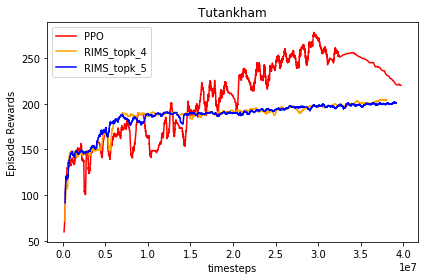}}&
\subfloat{\includegraphics[width = 0.95in]{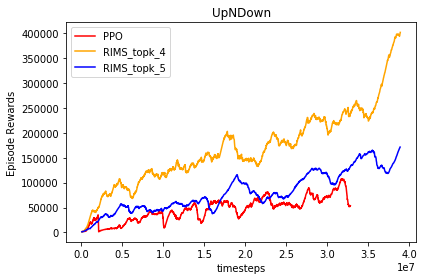}}&
\subfloat{\includegraphics[width = 0.95in]{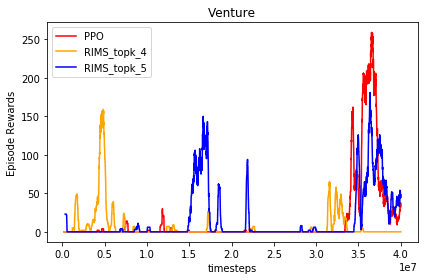}} \\
\subfloat{\includegraphics[width = 0.95in]{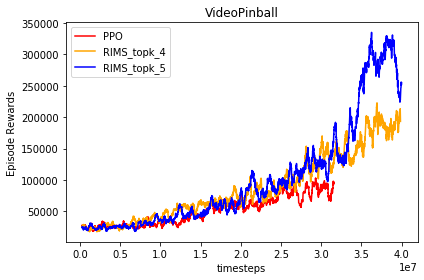}} &
\subfloat{\includegraphics[width = 0.95in]{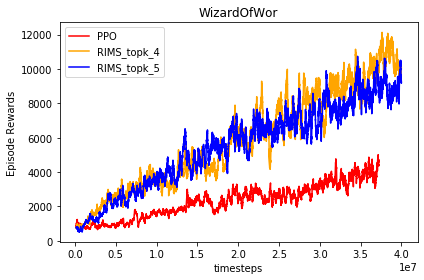}} &
\subfloat{\includegraphics[width = 0.95in]{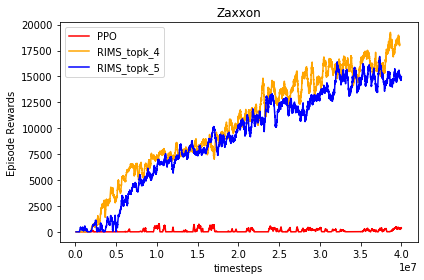}}&
\end{tabular}
\caption{\textbf{Comparing RIMs-PPO with LSTM-PPO:} Learning curves for $k_A=4$, $k_A=5$ RIMs-PPO models and the LSTM-PPO baseline across all Atari games.}
\end{figure*}

\newpage

\subsubsection{Atari Results: No Input attention}

Here we compare the proposed method to the baseline, where we dont use input attention, and we force different RIMs to communicate with each at all the time steps.

\begin{figure*}[htb!]
\begin{tabular}{ccccc}

\subfloat{\includegraphics[width = 1in]{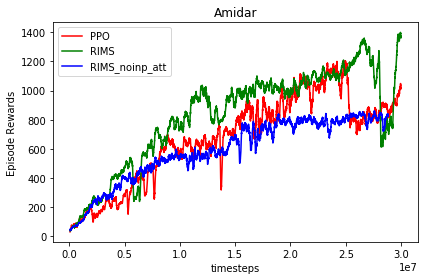}} &
\subfloat{\includegraphics[width = 1in]{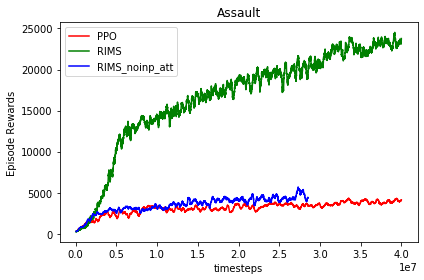}} &
\subfloat{\includegraphics[width = 1in]{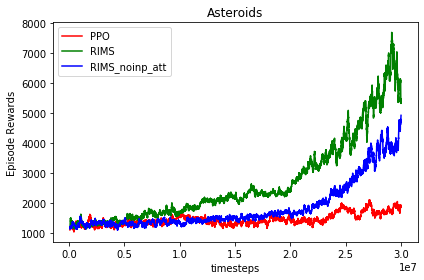}}&
\subfloat{\includegraphics[width = 1in]{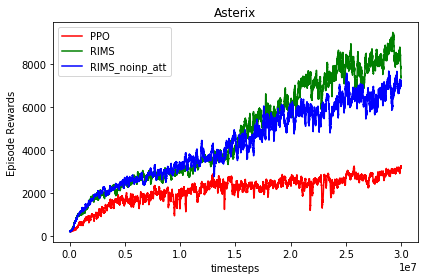}}&
\subfloat{\includegraphics[width = 1in]{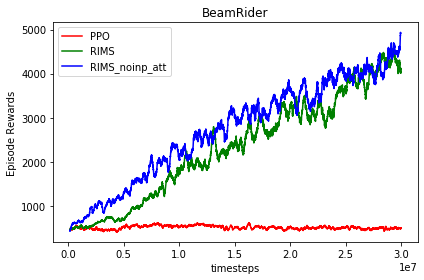}}\\
\subfloat{\includegraphics[width = 1in]{atari_results/BankHeist.png}} &
\subfloat{\includegraphics[width = 1in]{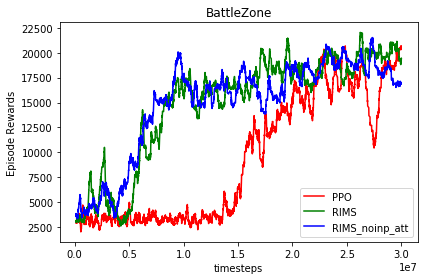}}&
\subfloat{\includegraphics[width = 1in]{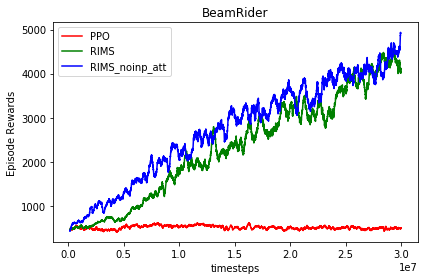}}&
\subfloat{\includegraphics[width = 1in]{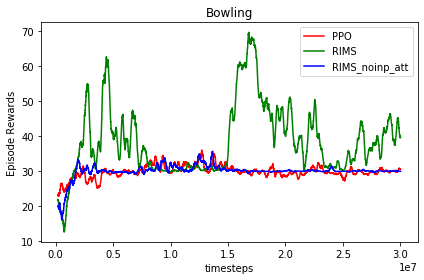}} &
\subfloat{\includegraphics[width = 1in]{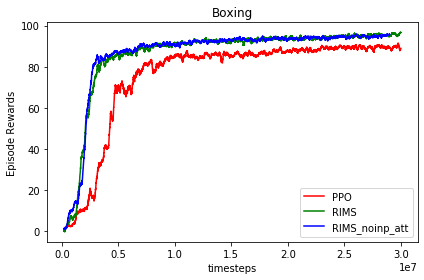}} \\
\subfloat{\includegraphics[width = 1in]{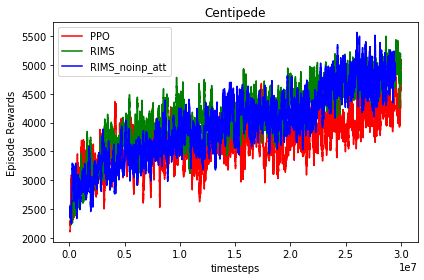}} &
\subfloat{\includegraphics[width = 1in]{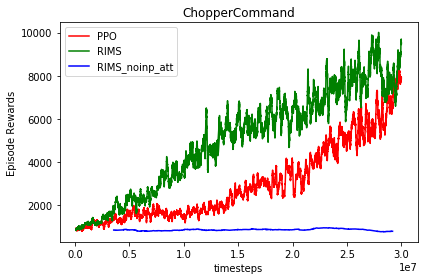}}&
\subfloat{\includegraphics[width = 1in]{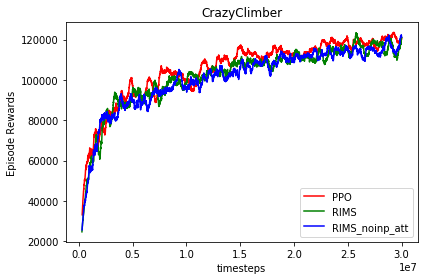}} &
\subfloat{\includegraphics[width = 1in]{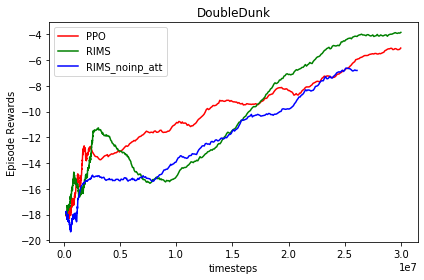}} &
\subfloat{\includegraphics[width = 1in]{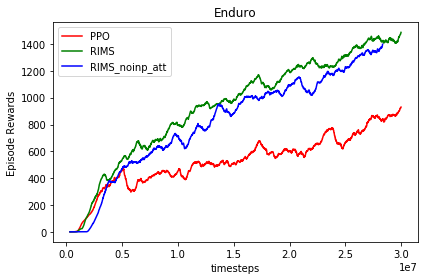}} \\
\subfloat{\includegraphics[width = 1in]{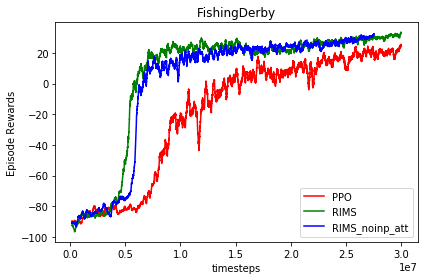}}&
\subfloat{\includegraphics[width = 1in]{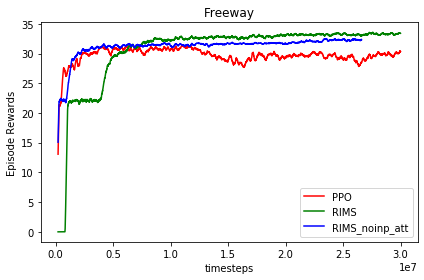}} &
\subfloat{\includegraphics[width = 1in]{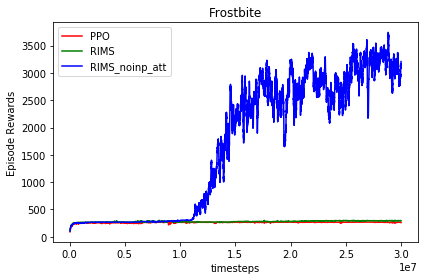}} &
\subfloat{\includegraphics[width = 1in]{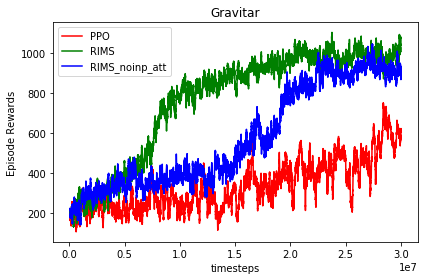}} &
\subfloat{\includegraphics[width = 1in]{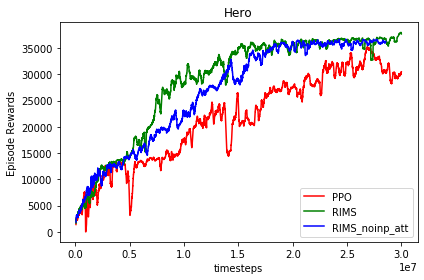}}\\
\subfloat{\includegraphics[width = 1in]{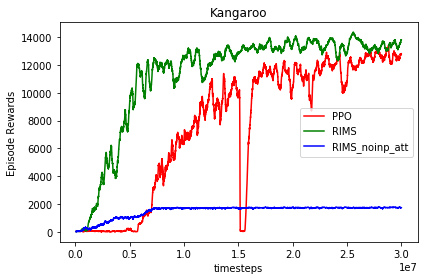}}&
\subfloat{\includegraphics[width = 1in]{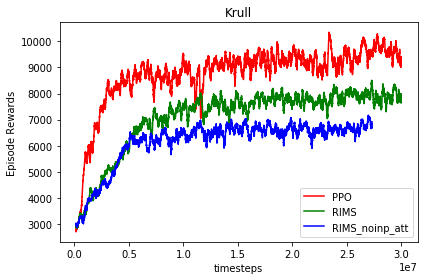}} &
\subfloat{\includegraphics[width = 1in]{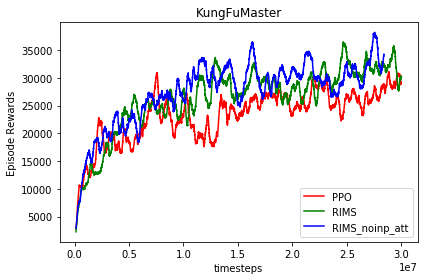}} &
\subfloat{\includegraphics[width = 1in]{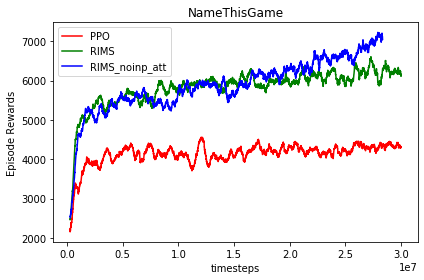}} &
\subfloat{\includegraphics[width = 1in]{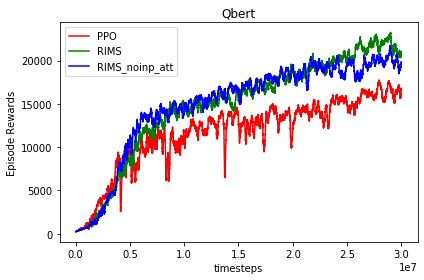}}\\
\subfloat{\includegraphics[width = 1in]{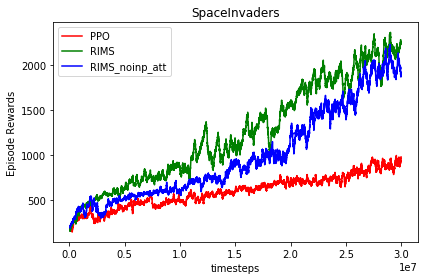}}&
\subfloat{\includegraphics[width = 1in]{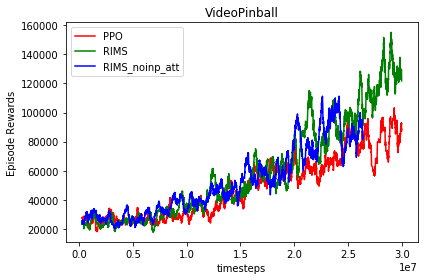}}&
\subfloat{\includegraphics[width = 1in]{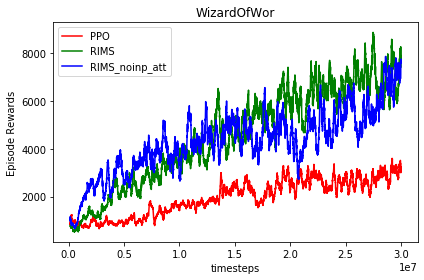}}&
\subfloat{\includegraphics[width = 1in]{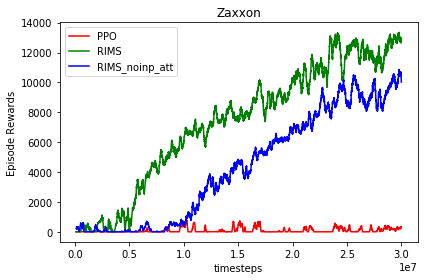}} \\
\end{tabular}
\caption{\textbf{Baseline agent with no input attention mechanism:} Here we compare the RIMs to the baseline, where their is no input attention (i.e., top down attention) as well as all the RIMs communicate with each other at all the time steps. Learning curves for RIMs-PPO models, Baseline Agent, the LSTM-PPO baseline across 30 Atari games.}
\label{fig:no_attention_rims_comm}
\end{figure*}

\newpage

\subsection{Bouncing MNIST: Dropping individual RIMs}
\label{appednix:sec:droprims}

We use the Stochastic Moving MNIST (SM-MNIST) \citep{denton2018stochastic} dataset which consists of sequences of frames of size $64\times 64$, containing one or two MNIST digits moving and bouncing off the walls. Training sequences are generated on the fly by sampling two different MNIST digits from the training set (60k total digits) and two distinct trajectories.

\begin{figure}[ht]
    \centering
    \includegraphics[width=0.6\linewidth, clip=true, trim = 0mm 0cm 0mm 0mm]{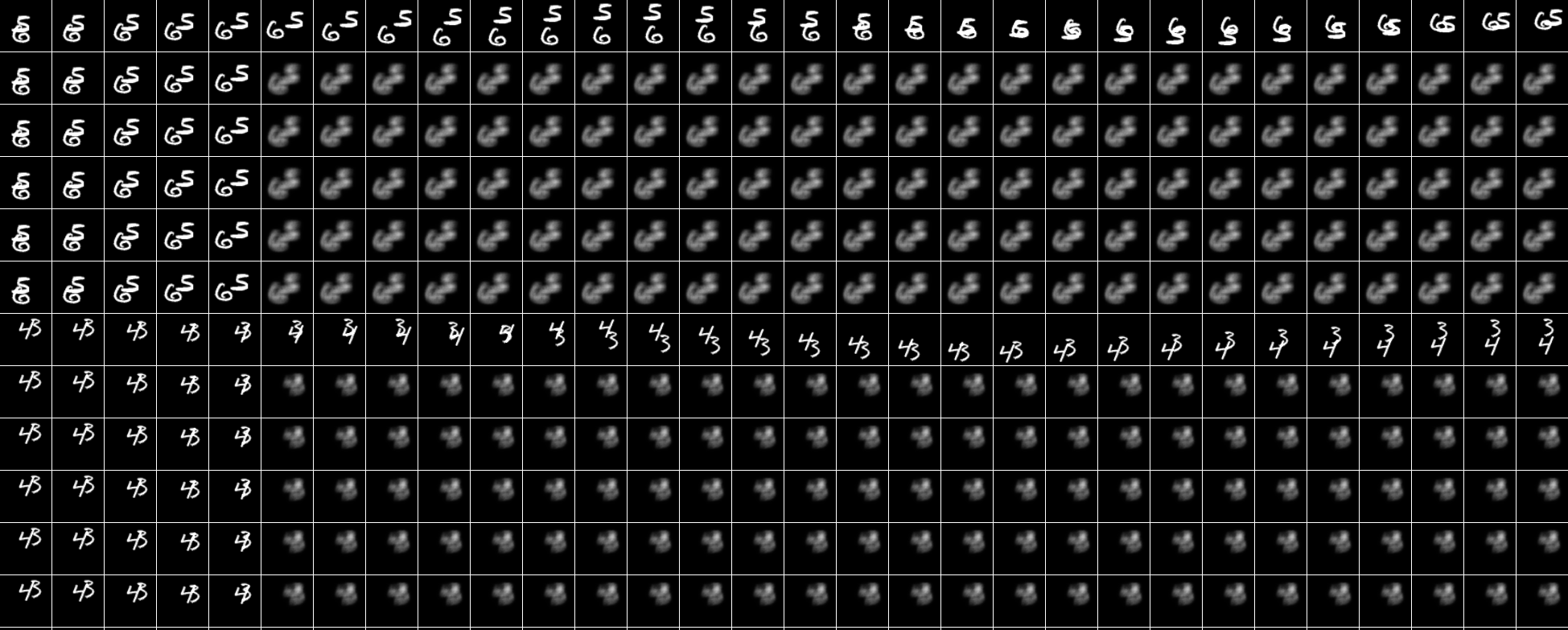}

    \vspace{1cm}
    \includegraphics[width=0.6\linewidth, clip=true, trim = 0mm 0cm 0mm 0mm]{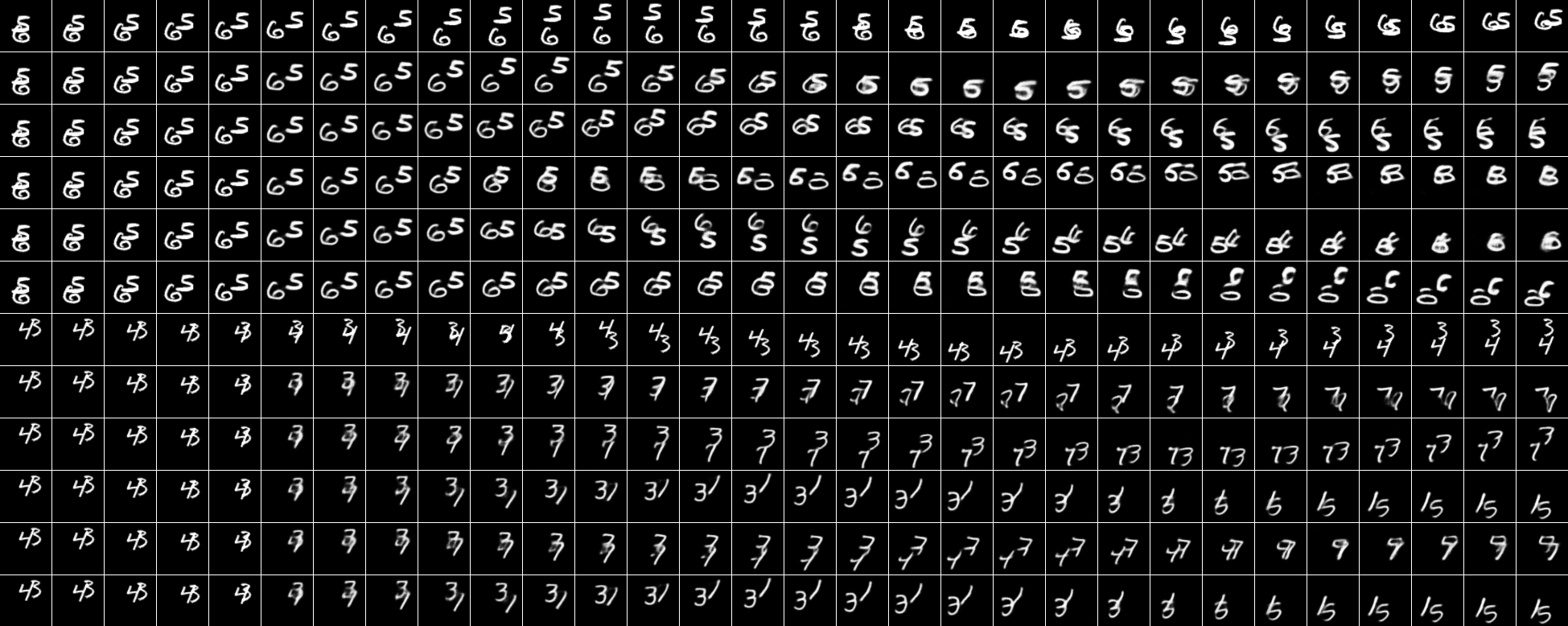}

    \vspace{1cm}

    \includegraphics[width=0.6\linewidth, clip=true, trim = 0mm 0cm 0mm 0mm]{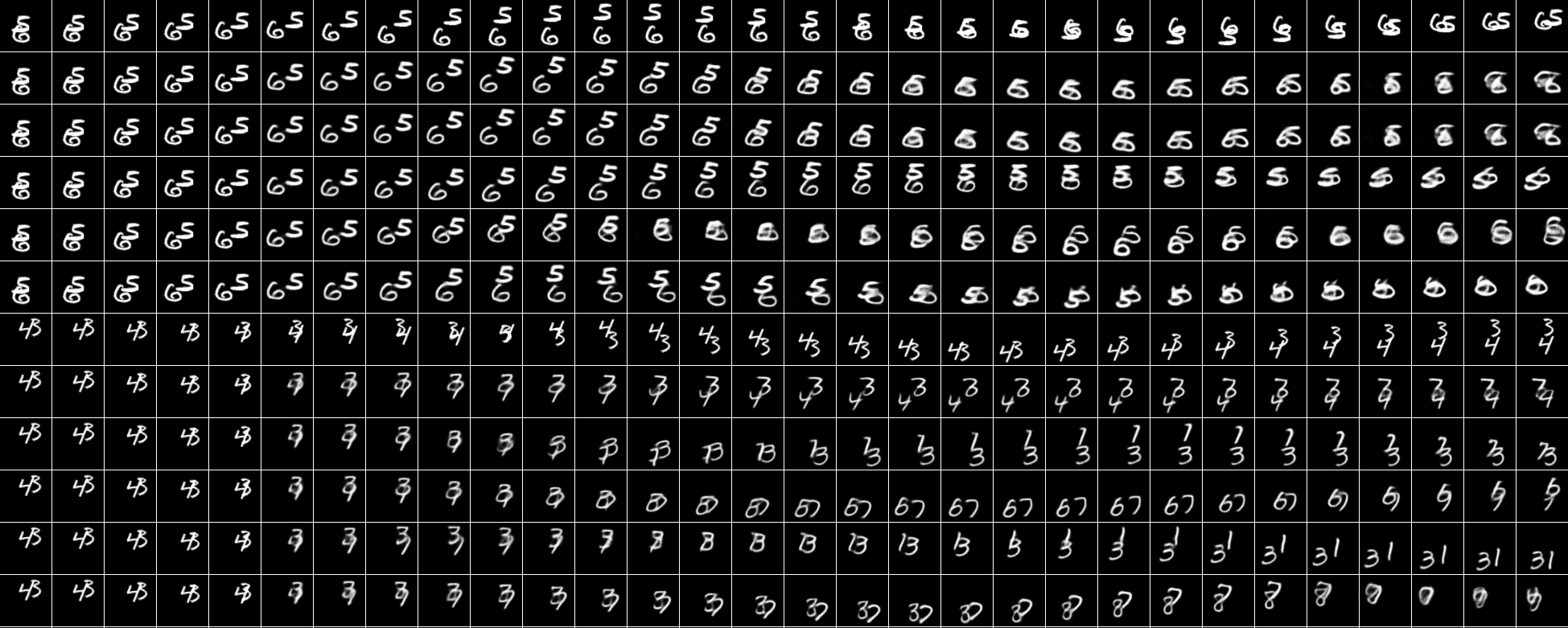}
    \vspace{1cm}
    
    \includegraphics[width=0.6\linewidth, clip=true, trim = 0mm 0cm 0mm 0mm]{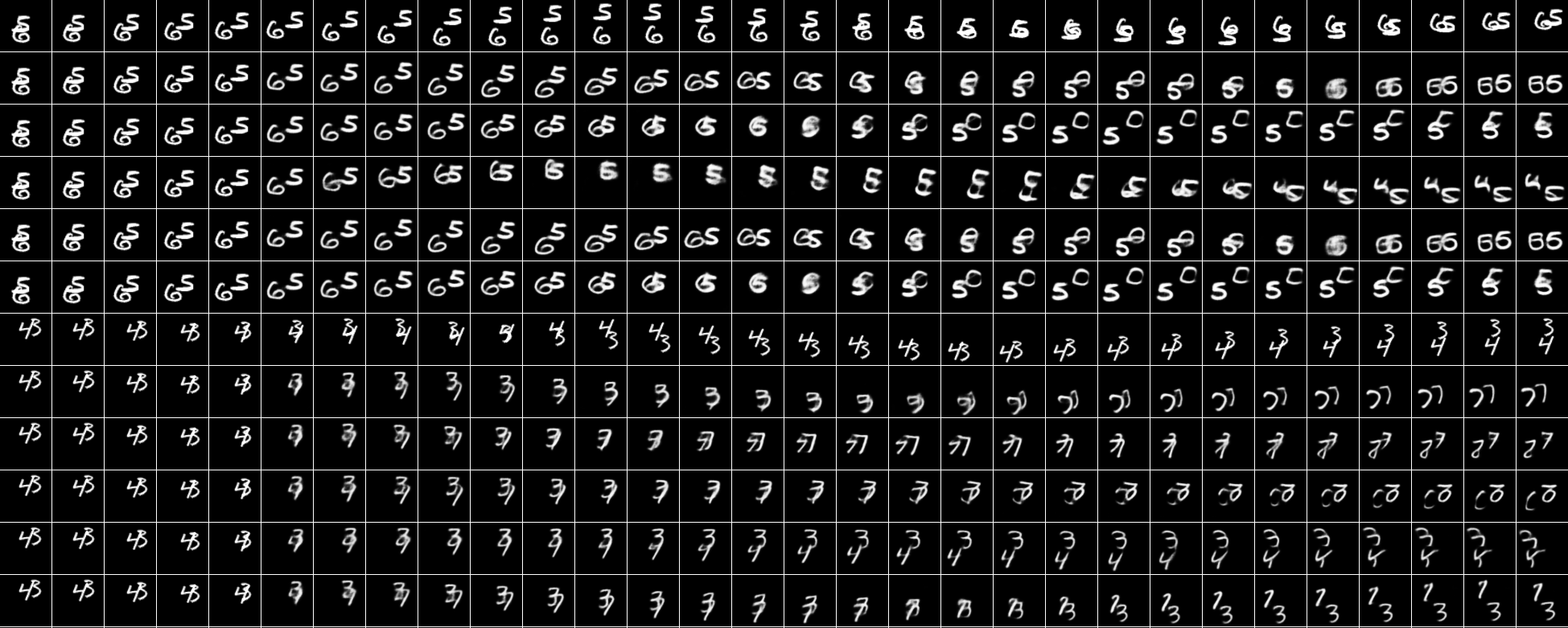}
    \caption{4 RIMs, (top k = 2). Each sub-figure shows the effect of masking a particular RIM and studying the effect of masking on the other RIMs. For example, the top figure shows the effect of masking the first RIM, the second figure shows the effect of masking the second RIM etc.}
    \label{fig:4_blocks_top_k_2}
\end{figure}

\begin{figure}[ht]
    \centering
    \includegraphics[width=0.6\linewidth, clip=true, trim = 0mm 0cm 0mm 0mm]{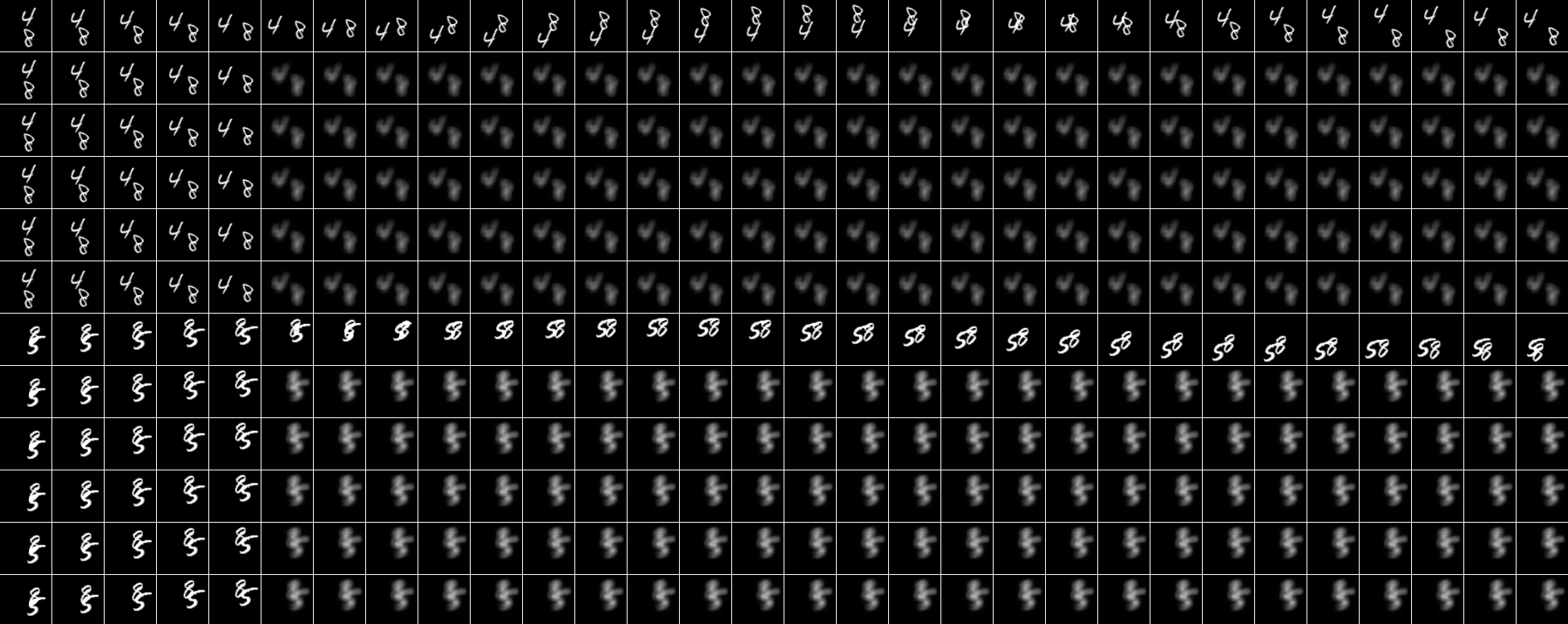}
 
    \vspace{1cm}
    \includegraphics[width=0.6\linewidth, clip=true, trim = 0mm 0cm 0mm 0mm]{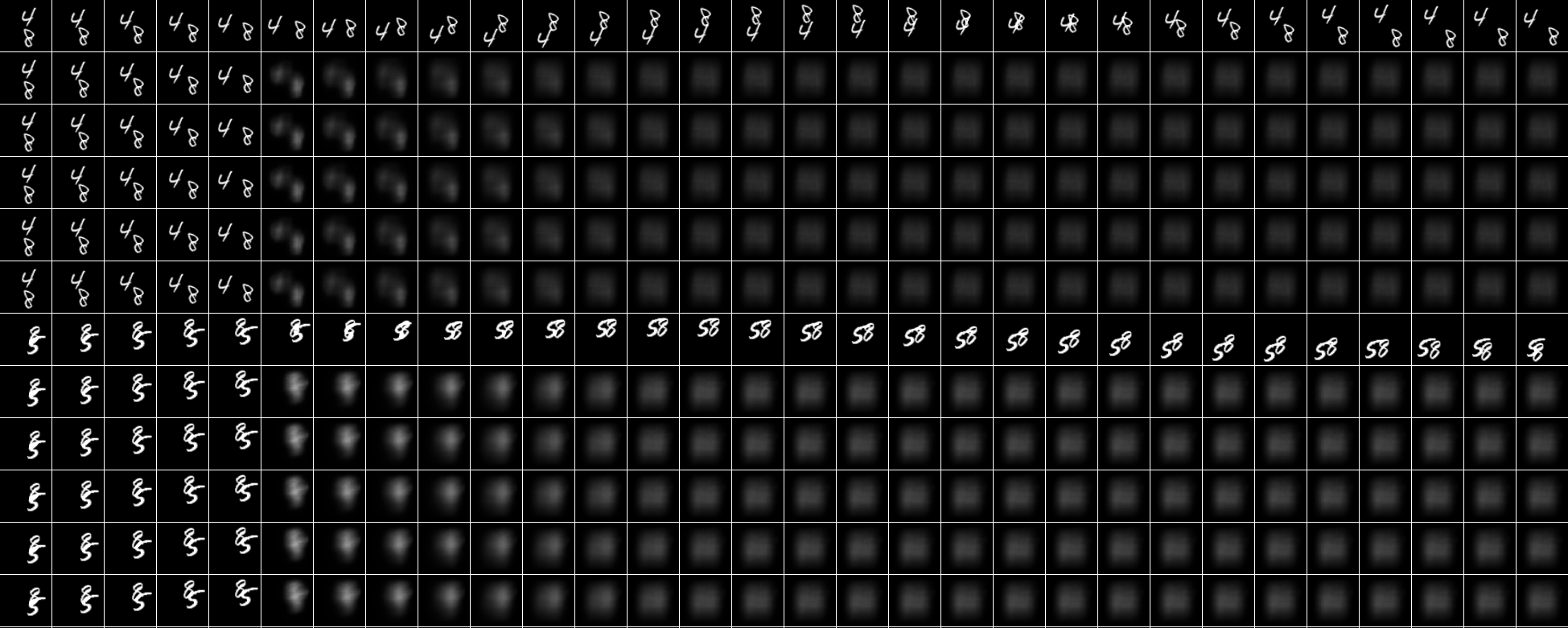}

    \vspace{1cm}
    \includegraphics[width=0.6\linewidth, clip=true, trim = 0mm 0cm 0mm 0mm]{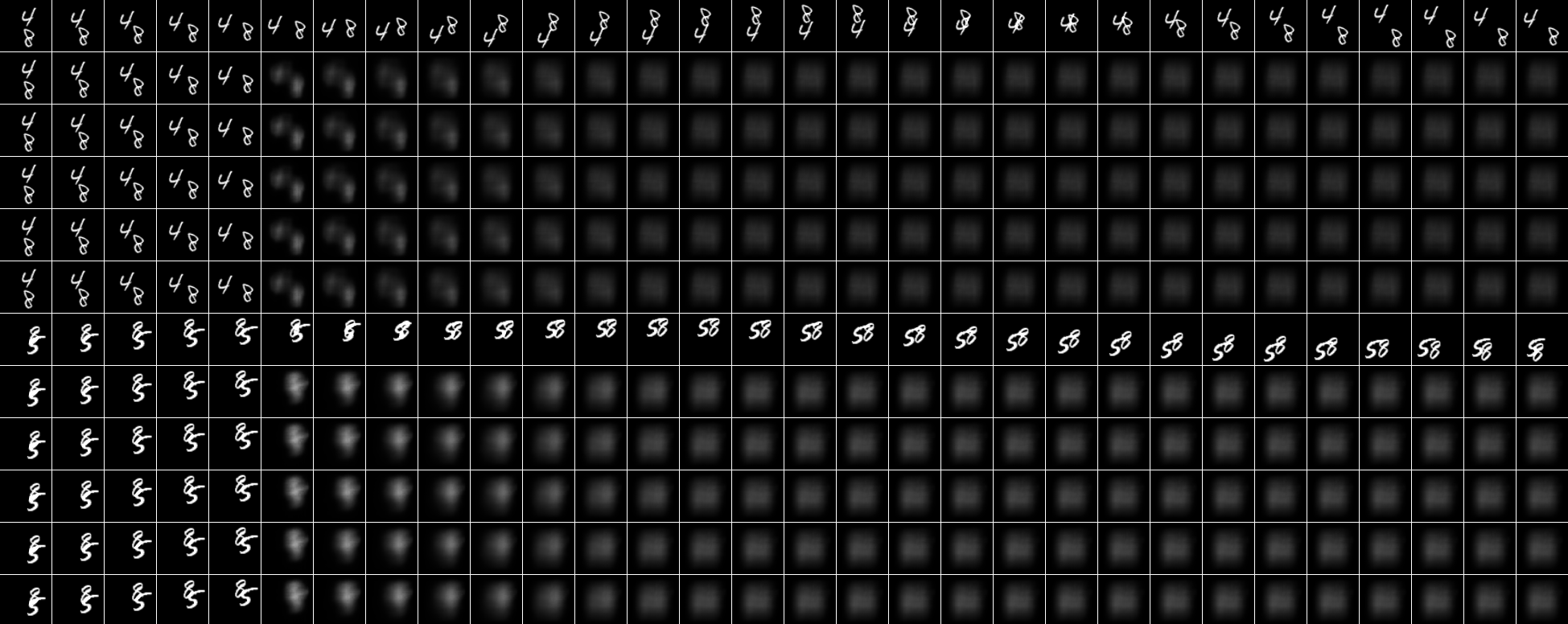}
   
    \vspace{1cm}
    \includegraphics[width=0.6\linewidth, clip=true, trim = 0mm 0cm 0mm 0mm]{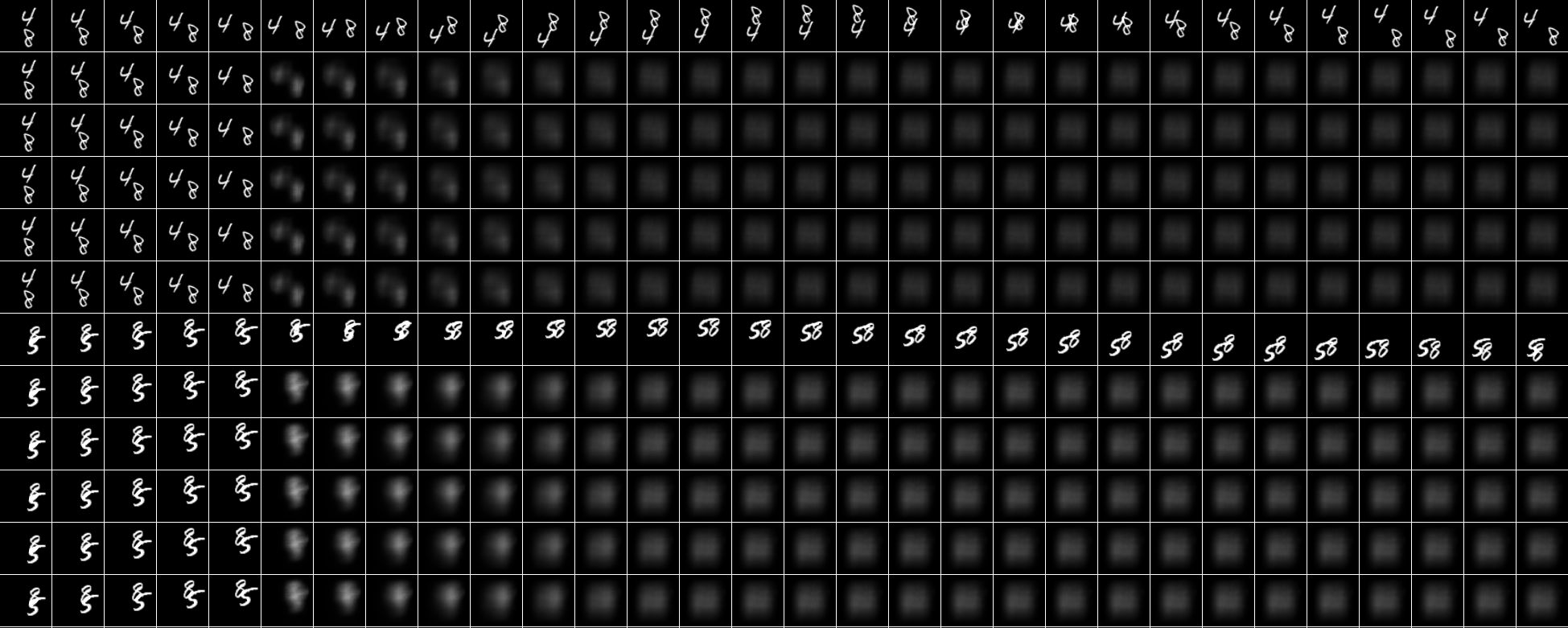}
    \vspace{1cm}
    \caption{4 RIMs, (top k = 3). Each sub-figure shows the effect of masking a particular RIM and studying the effect of masking on the other RIMs. For example, the top figure shows the effect of masking the first RIM, the second figure shows the effect of masking the second RIM etc.}
    \label{fig:4_blocks_top_k_3}
\end{figure}

\begin{figure}[ht]
    \centering
    \includegraphics[width=0.6\linewidth, clip=true, trim = 0mm 0cm 0mm 0mm]{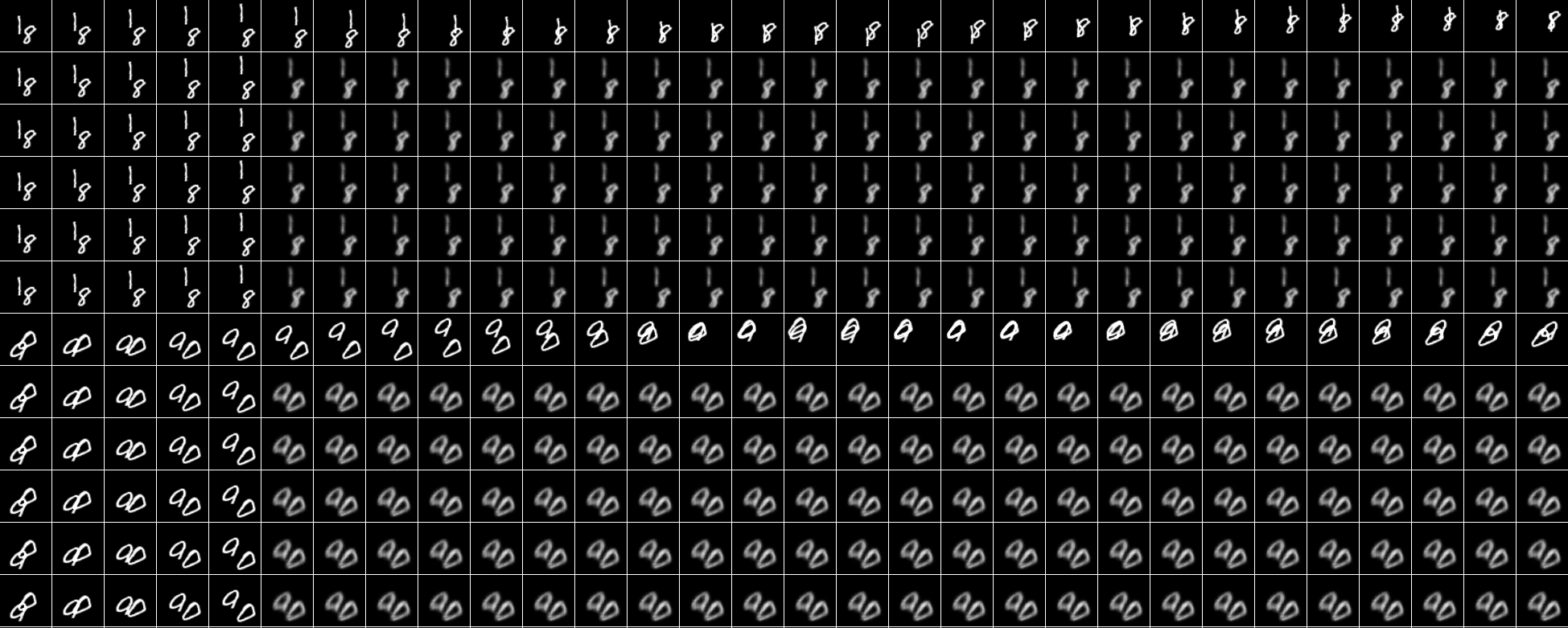}
    \vspace{1cm}
    \includegraphics[width=0.6\linewidth, clip=true, trim = 0mm 0cm 0mm 0mm]{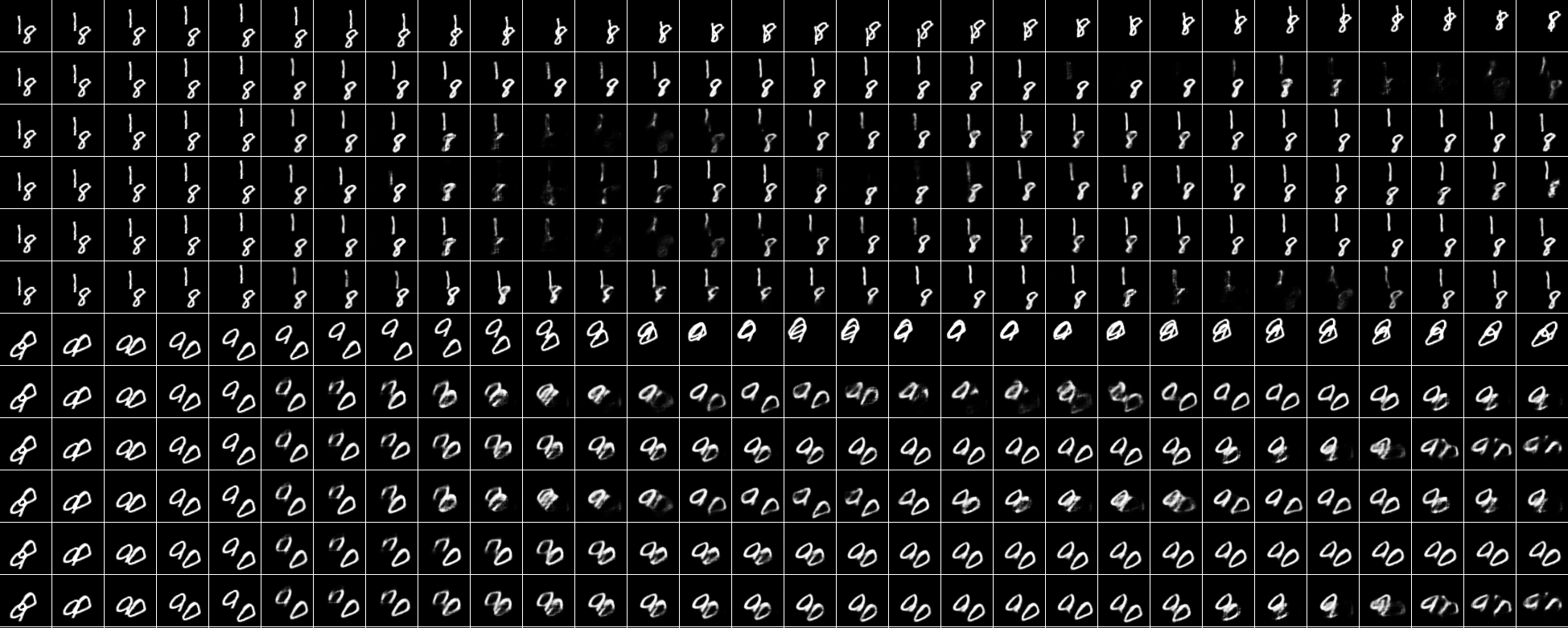}
    \vspace{0.5cm}
    \includegraphics[width=0.6\linewidth, clip=true, trim = 0mm 0cm 0mm 0mm]{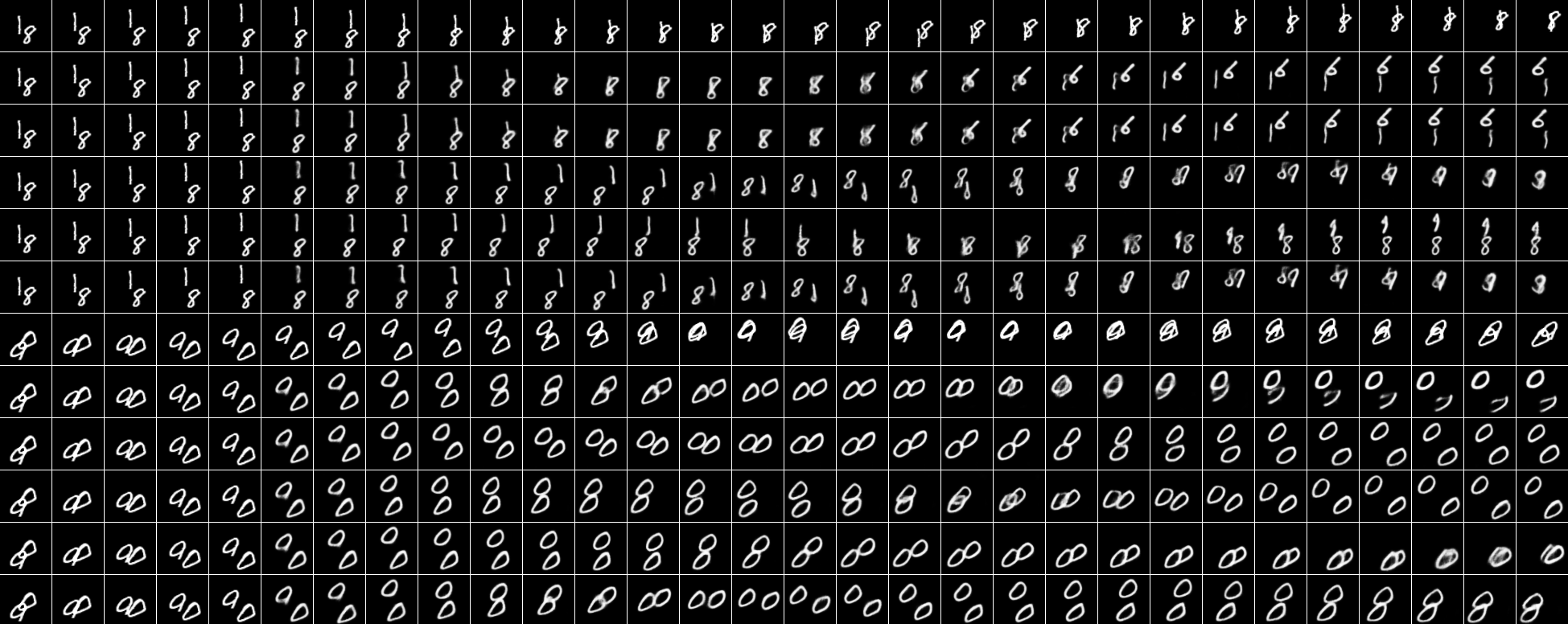}
    \vspace{0.5cm}
    \includegraphics[width=0.6\linewidth, clip=true, trim = 0mm 0cm 0mm 0mm]{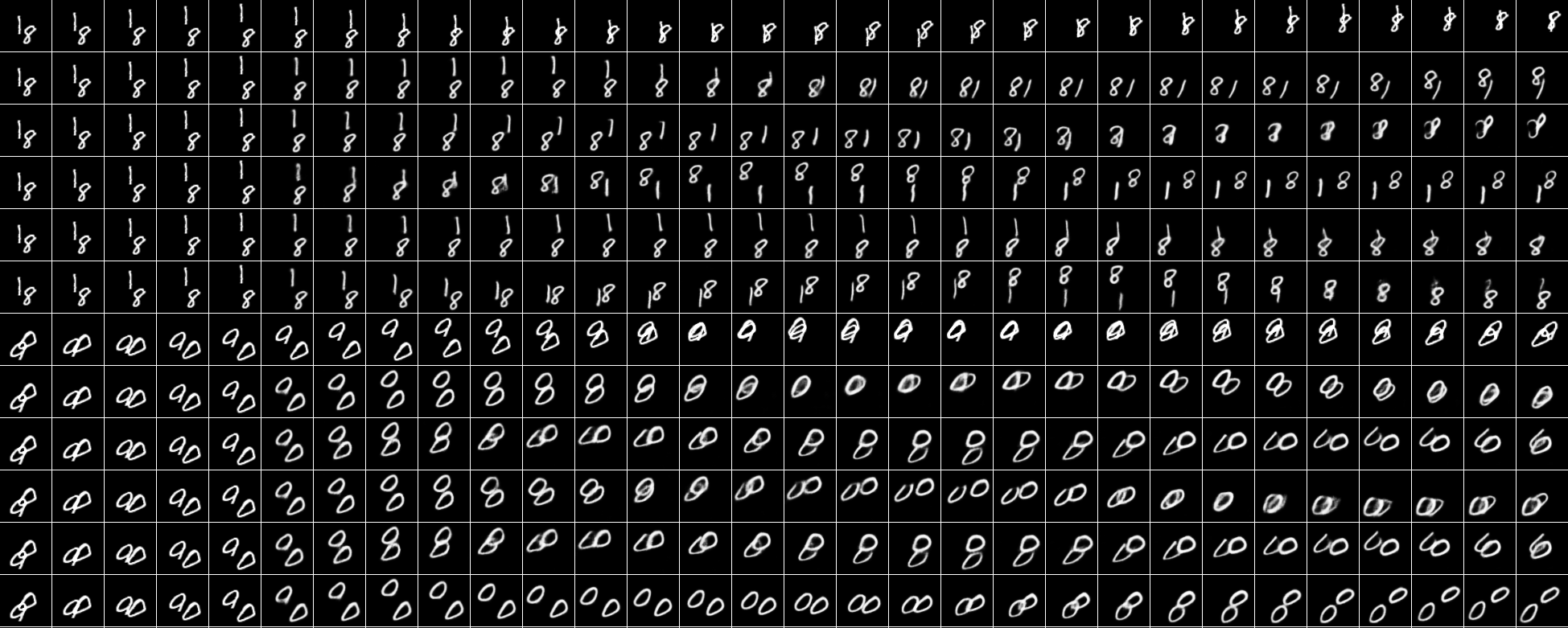}
    \vspace{0.5cm}
    \includegraphics[width=0.6\linewidth, clip=true, trim = 0mm 0cm 0mm 0mm]{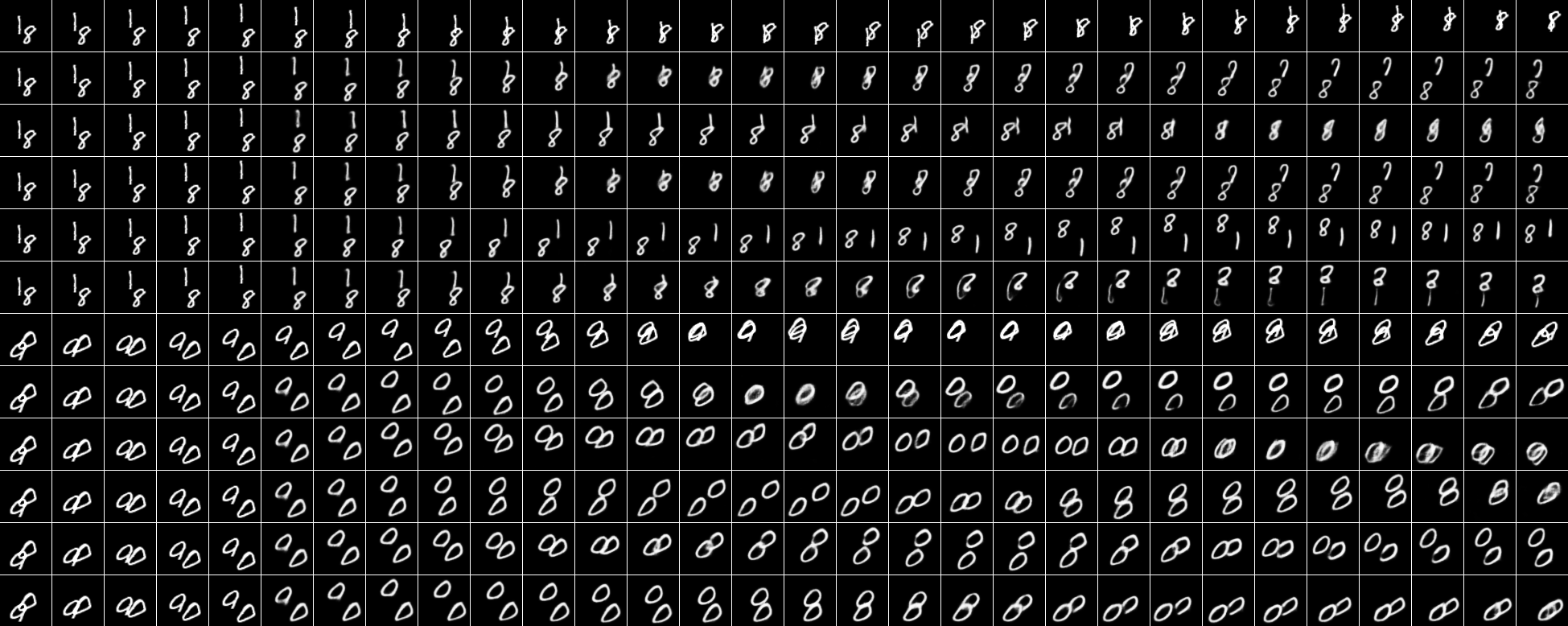}
    \caption{400dim, $k_T = 5$, $k_A = 2$. Each sub-figure shows the effect of masking a particular RIM and studying the effect of masking on the other RIMs. For example, the top figure shows the effect of masking the first RIM, the second figure shows the effect of masking the second RIM etc.}
    \label{fig:5_blocks_top_k_2}
\end{figure}

\begin{figure}[ht]
    \centering
    \includegraphics[width=0.6\linewidth, clip=true, trim = 0mm 0cm 0mm 0mm]{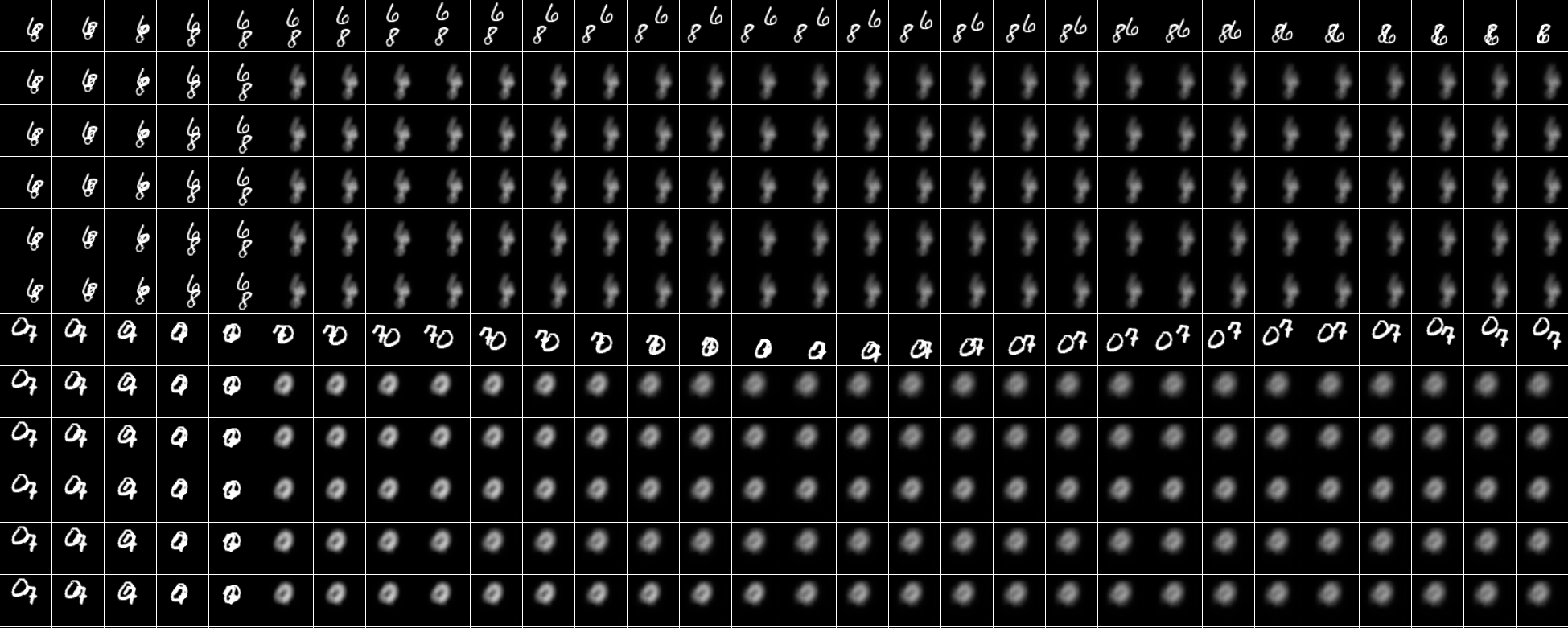}
    \vspace{1cm}
    \includegraphics[width=0.6\linewidth, clip=true, trim = 0mm 0cm 0mm 0mm]{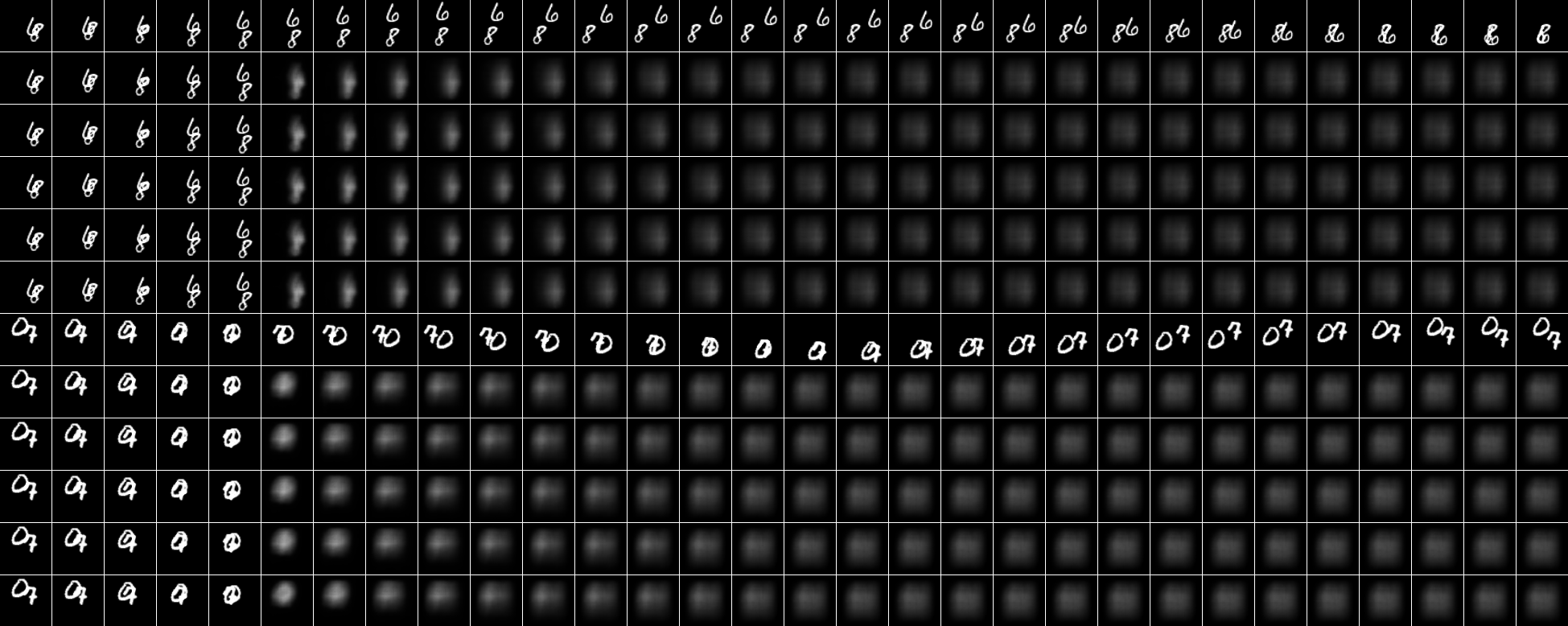}
    \vspace{0.5cm}
    \includegraphics[width=0.6\linewidth, clip=true, trim = 0mm 0cm 0mm 0mm]{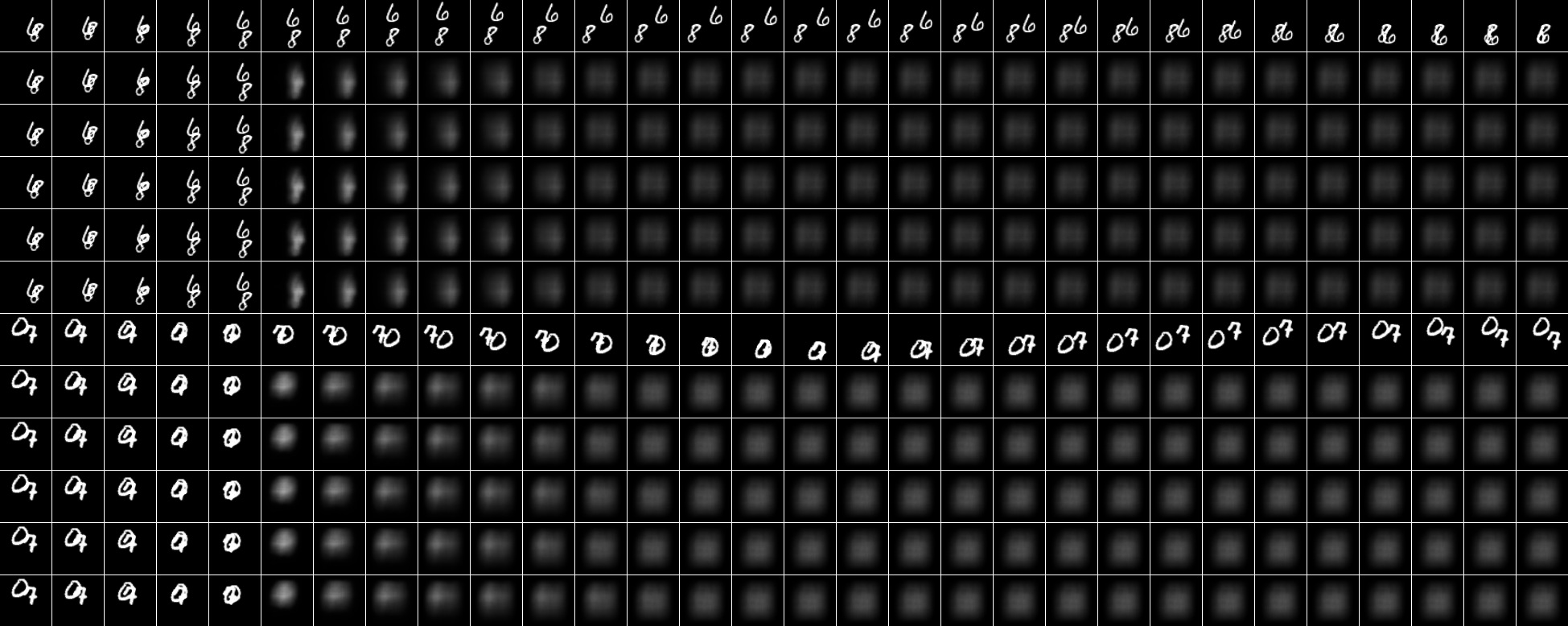}
    \vspace{0.5cm}
    \includegraphics[width=0.6\linewidth, clip=true, trim = 0mm 0cm 0mm 0mm]{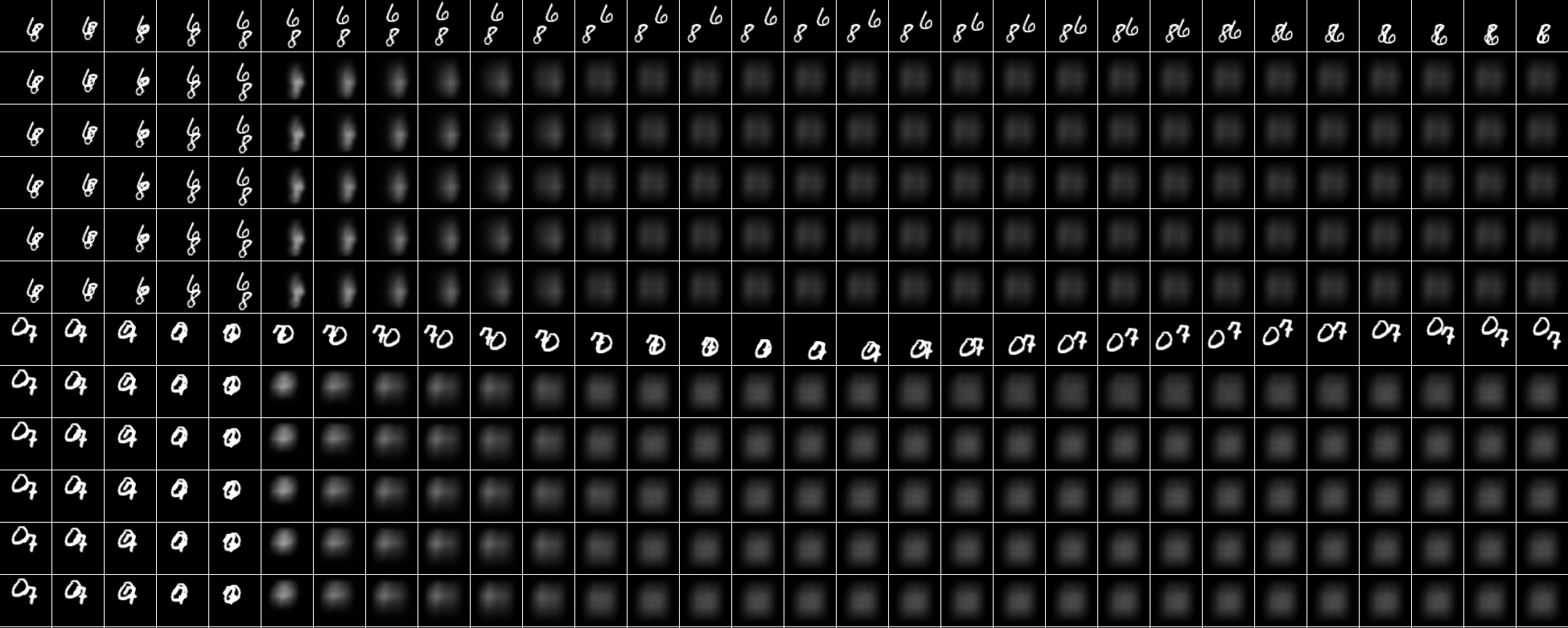}
    \vspace{0.5cm}
    \includegraphics[width=0.6\linewidth, clip=true, trim = 0mm 0cm 0mm 0mm]{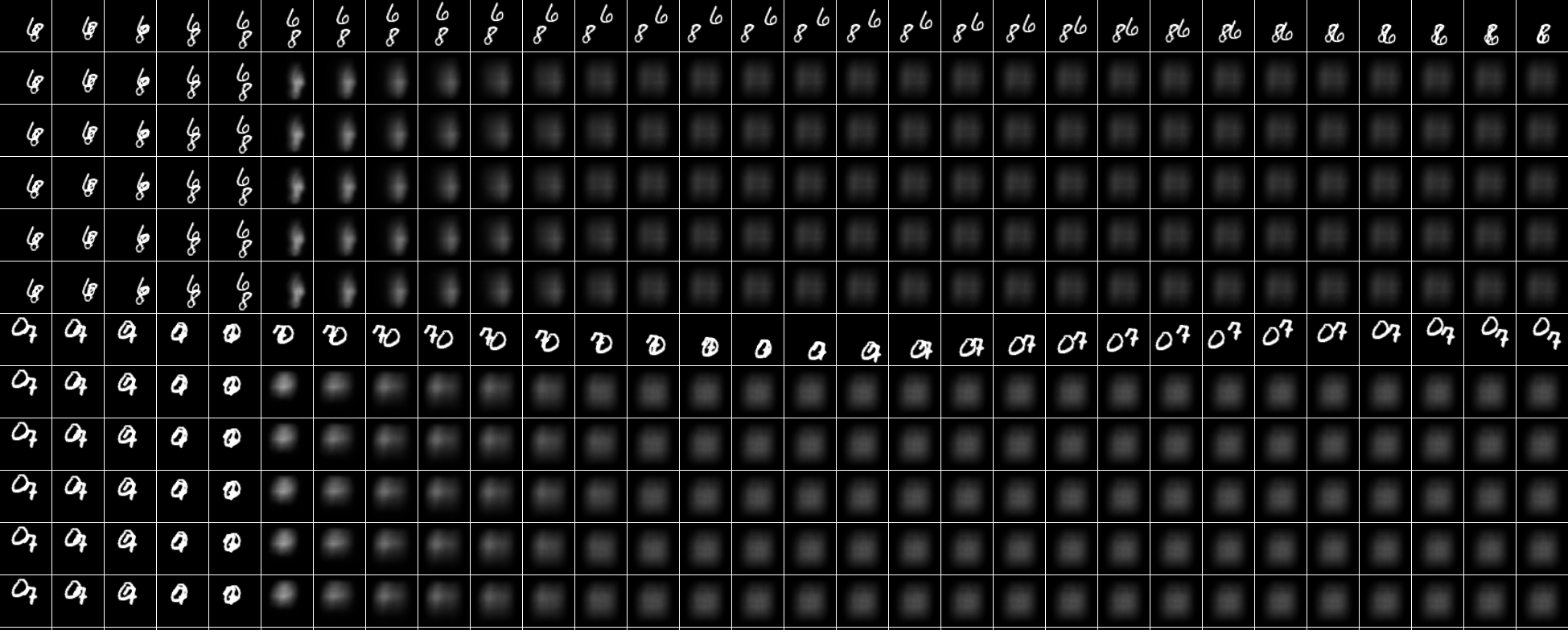}
    \vspace{0.5cm}
    \caption{400dim, $k_T = 5$, $k_A = 3$. Each sub-figure shows the effect of masking a particular RIM and studying the effect of masking on the other RIMs. For examples, the top figure shows the effect of masking the first RIM, the second figure shows the effect of masking the second RIM etc.}
    \label{fig:5_blocks_top_k_3}
\end{figure}

\begin{figure}[ht]
    \centering
    \includegraphics[width=0.6\linewidth, clip=true, trim = 0mm 0cm 0mm 0mm]{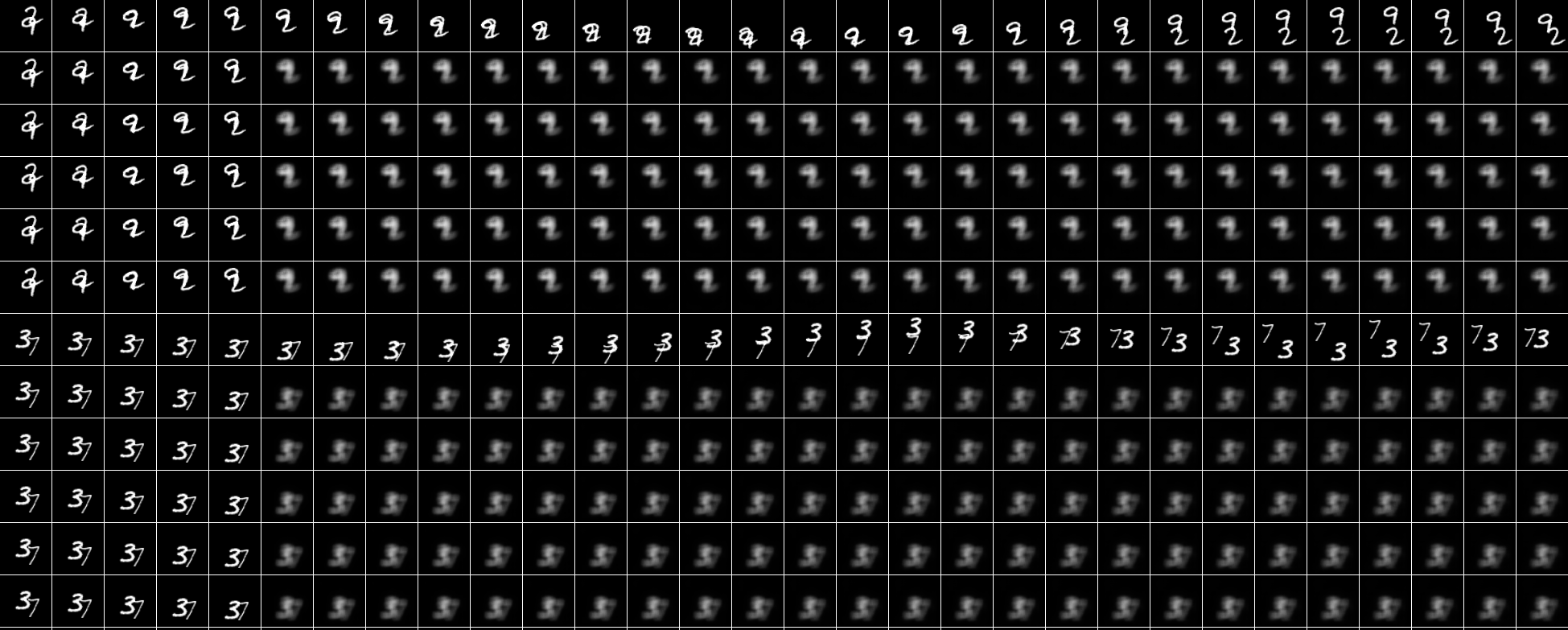}
    \vspace{1cm}
    \includegraphics[width=0.6\linewidth, clip=true, trim = 0mm 0cm 0mm 0mm]{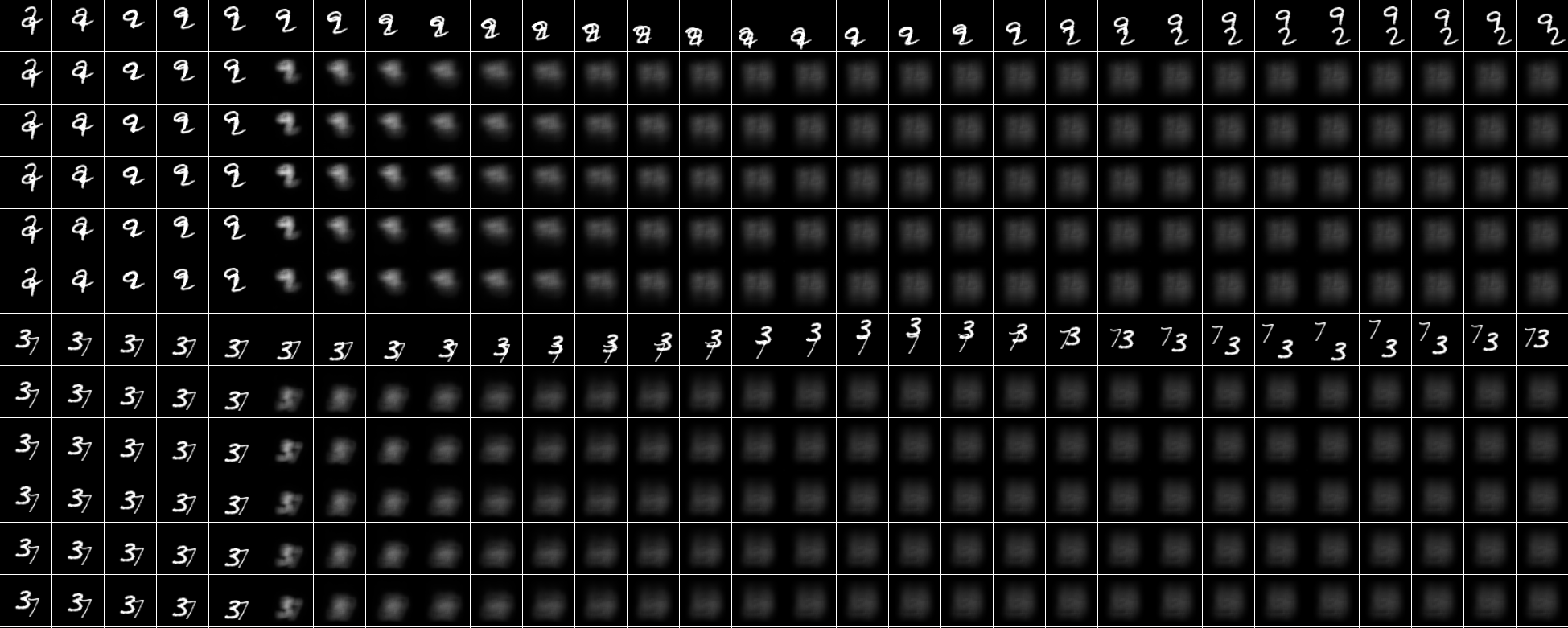}
    \vspace{0.5cm}
    \includegraphics[width=0.6\linewidth, clip=true, trim = 0mm 0cm 0mm 0mm]{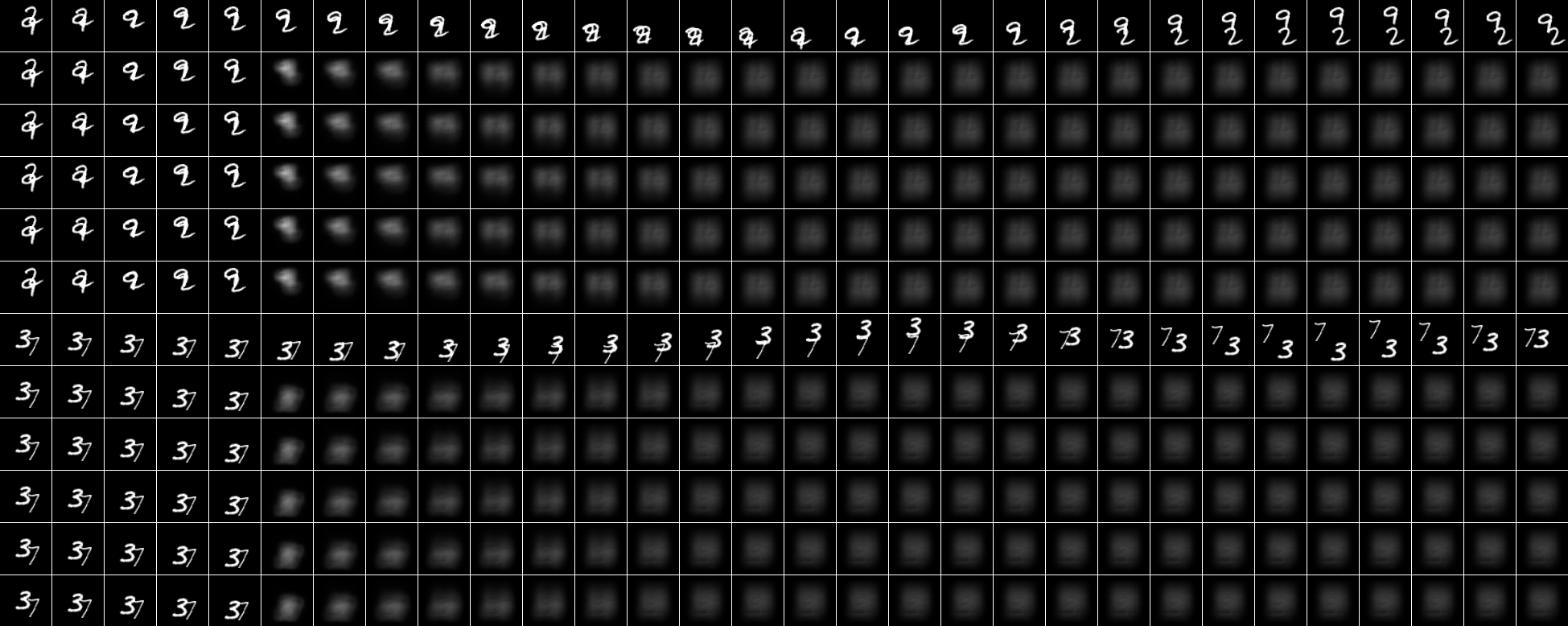}
    \vspace{0.5cm}
    \includegraphics[width=0.6\linewidth, clip=true, trim = 0mm 0cm 0mm 0mm]{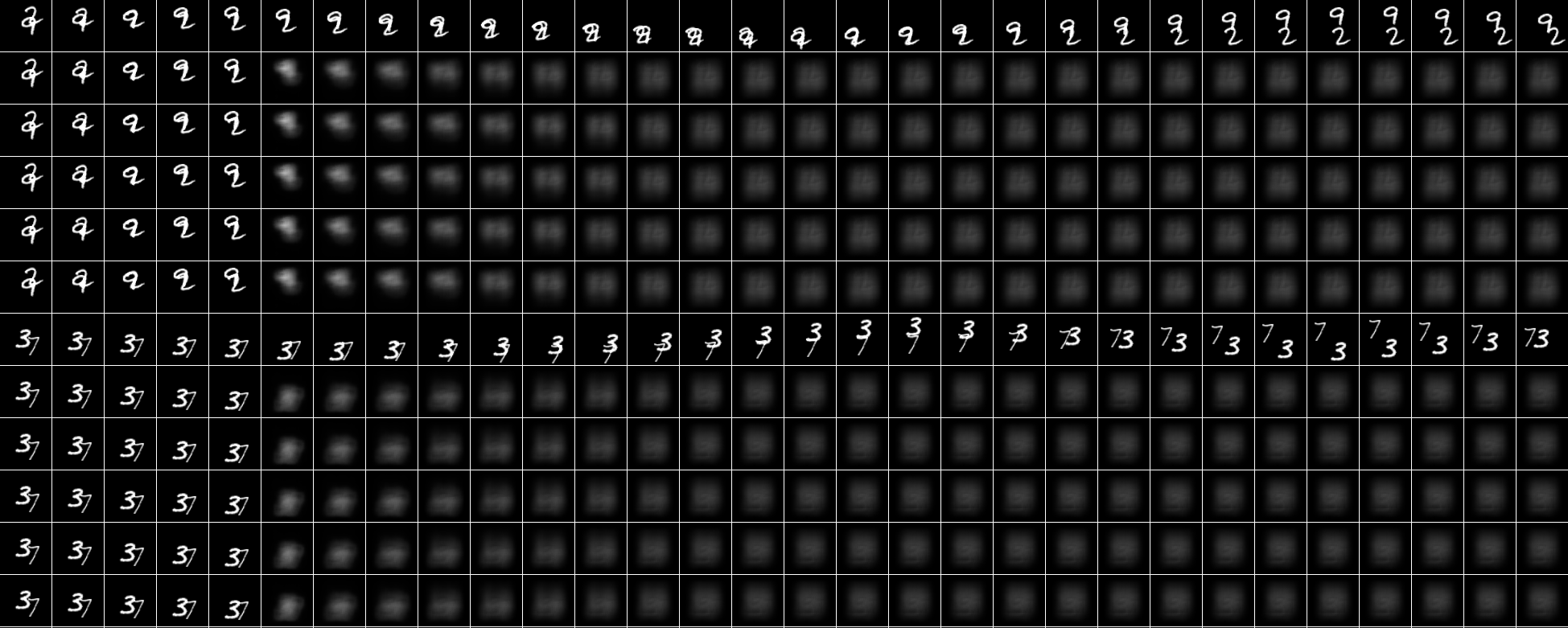}
    \vspace{0.5cm}
    \includegraphics[width=0.6\linewidth, clip=true, trim = 0mm 0cm 0mm 0mm]{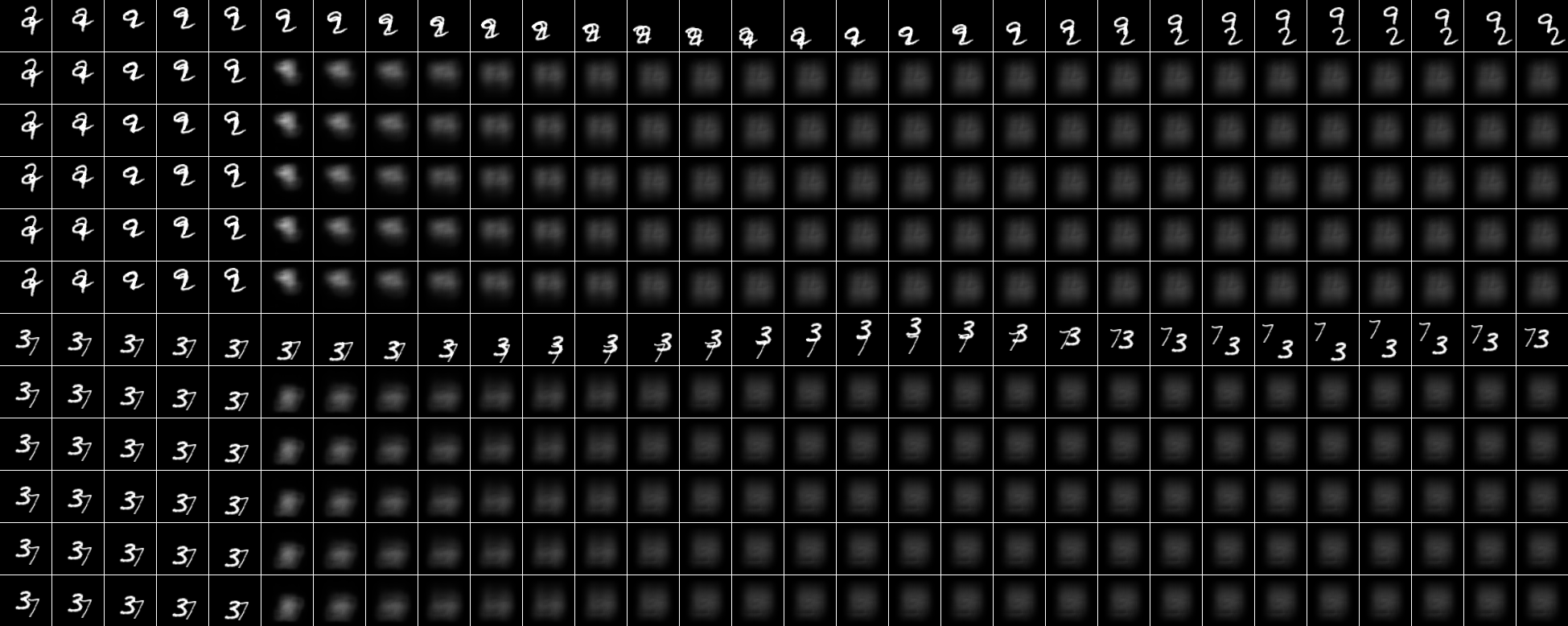}
    \vspace{0.5cm}
    \caption{400dim, $k_T = 5$, $k_A = 4$. Each sub-figure shows the effect of masking a particular RIM and studying the effect of masking on the other RIMs. For example, the top figure shows the effect of masking the first RIM, the second figure shows the effect of masking the second RIM etc.}
    \label{fig:5_blocks_top_k_4}
    
\end{figure}

Here, we show the effect of masking out a particular RIM and study the effect of the masking on the ensemble of RIMs. Ideally, we would want different RIMs not to co-adapt with each other. So, masking out a particular RIM should not really effect the dynamics of the entire model. We show qualitative comparisons in Fig.\ \ref{fig:4_blocks_top_k_2}, \ref{fig:4_blocks_top_k_3}, \ref{fig:5_blocks_top_k_2}, \ref{fig:5_blocks_top_k_3}, \ref{fig:5_blocks_top_k_4}. 
In each of these figures, the model gets the ground truth image as input for first 5 time steps, and then asked to simulate the dynamics for next 25 time-steps. We find that sparsity is needed otherwise different RIMs co-adapt with each other (for ex. see Fig. \ref{fig:4_blocks_top_k_3}, \ref{fig:5_blocks_top_k_3}, \ref{fig:5_blocks_top_k_4}).  We tried similar masking experiments for different models like RMC, Transformers, EntNet (which learns a mixture of experts), LSTMs, but all of them failed to do anything meaningful (after masking). We suspect this is partly due to learning a homogeneous network.


\section{Natural Language Processing: Language Modeling, Machine Translation, and Transfer Learning}

We performed some additional experiments to evaluate how well RIMs improve transfer learning on some widely studied NLP datasets.  


We performed an additional experiment where we trained a seq2seq model on the WMT machine translation dataset and evaluated on the IWSLT14 dataset (English to German).  WMT consists of European Parliament text and news articles, whereas IWSLT14 consists of transcribed spoken text (for example, TED talks).  Thus the two datasets are in the same general domain but have fairly distinct content distributions.  Additionally, we considered either training the WMT model on both English to German and English to French (with shared encoders but separate decoders) or only on English to German.  The results are in Table~\ref{tb:nlp_transfer}.  The LSTM and RIMs models used have a comparable number of parameters in this experiment.  We note that the multi-task training setup substantially hurts the performance of the LSTM baseline, but helps performance when using RIMs (the performance of the transformer is about the same in both settings).  

\begin{table}
\centering
\caption{Transfer from WMT to IWSLT (en $\rightarrow$ de) for different models.  Results in BLEU score (higher is better).  }
\begin{tabular}{lccc}
\hline
Training Data & (Vanilla) Transformer & LSTM & RIMs \\ \hline
en $\rightarrow$ de  & 22.89 & 21.32 & \textbf{23.71} \\
en $\rightarrow$ de, en $\rightarrow$ fr & 22.92 & 20.37 & \textbf{24.23} \\
\end{tabular}
\label{tb:nlp_transfer}
\end{table}






\appendix

\end{document}